\documentclass[final,breaklinks]{colt2020-plain}
\usepackage{booktabs}
\usepackage{multirow}
\usepackage{framed}
\usepackage{times}
\usepackage{amsmath,amssymb,rotating,graphicx,rotating,float}
\usepackage{url}

\usepackage{tikz}
\usetikzlibrary{matrix,arrows,snakes,shapes,trees,positioning}

\usepackage{color}

\usepackage{cleveref}
\newlength{\dhatheight}

\newcommand{\alg}{\mathbb{A}}
\renewcommand{\H}{\mathbb H}

\newcommand{\MB}{{\mathsf{MB}}}
\newcommand{\Regret}{{\mathsf{Reg}}}
\newcommand{\Reg}{\Regret}
\newcommand{\R}{\mathbb{R}}
\newcommand{\vcdim}{\mathsf{VC}}
\newcommand{\vc}{\vcdim}
\newcommand{\td}{{\rm {TD}}}

\newcommand{\LD}{{\mathsf{LD}}}
\newcommand{\Ldim}{{\mathsf{LD}}}

\newcommand{\val}{{\rm val}}

\newcommand{\B}{\mathcal{B}}

\newcommand{\PXY}{P}

\newcommand{\RE}{{\rm RE}}

\newcommand{\h}{h}

\newcommand{\SC}{\mathcal{M}}

\newcommand{\G}{{\mathbb{G}}}

\renewcommand{\epsilon}{\varepsilon}
\newcommand{\eps}{\varepsilon}

\newcommand{\X}{\mathcal X}

\newcommand{\er}{{\rm er}}

\DeclareSymbolFont{bbold}{U}{bbold}{m}{n}
\DeclareSymbolFontAlphabet{\mathbbold}{bbold}
\newcommand{\ind}{\mathbbold{1}}

\renewcommand{\P}{{\rm Pr}}

\newcommand{\nats}{\mathbb{N}}
\newcommand{\reals}{\mathbb{R}}

\newcommand{\E}{\mathop{\mathbb{E}}}

\newcommand{\argmax}{\mathop{\rm argmax}}
\newcommand{\argmin}{\mathop{\rm argmin}}

\newcommand{\supp}{{\rm supp}}

\newcommand{\ignore}[1]{}
\newcommand{\oldstuff}[1]{}

\colorlet{sgreen}{black!45!green}

\newsavebox{\savepar}

\makeatletter
\newcommand{\vast}{\bBigg@{3}}
\newcommand{\Vast}{\bBigg@{4}}
\makeatother

\tikzset{
    >=latex,
    punkt/.style={
           rectangle,
           rounded corners,
           draw=black, very thick,
           text width=6.5em,
           minimum height=2em,
           text centered},
    pil/.style={
           ->,
           double,
           thick,
           shorten <=2pt,
           shorten >=2pt,},
    punkti/.style={
           rectangle,
           rounded corners,
           draw=black, very thick,
           text width=26.5em,
           minimum height=2em,
           text centered},
    punktii/.style={
           rectangle,
           rounded corners,
           draw=black, very thick,
           text width=18.5em,
           minimum height=2em,
           text centered}
}

\newtheorem{question}{Open Question}

\renewenvironment{proof}[1][]{\par\noindent{\bf Proof #1\ }}{\hfill\BlackBox\\[2mm]}

\title[PAC Learnability of Partial Concept Classes]{A Theory of PAC Learnability of Partial Concept Classes}
\coltauthor{%
\Name{Noga Alon} \Email{nalon@math.princeton.edu} \\ \addr Princeton University 
\AND
\Name{Steve Hanneke} \Email{steve.hanneke@gmail.com} \\ \addr Toyota Technological Institute at Chicago 
\AND
\Name{Ron Holzman}
\Email{holzman@technion.ac.il}\\
\addr Technion
 \AND
 \Name{Shay Moran} \Email{smoran@technion.ac.il} \\ \addr Technion and Google Research
}

\begin{document}
\maketitle

\begin{abstract}
We extend the classical theory of PAC learning 
    in a way which allows to model a rich variety of practical learning tasks where the {data} satisfy special properties that ease the learning process. For example, tasks where the distance of the data from the decision boundary is bounded away from zero, or tasks where the data lie on a lower dimensional surface. 
    The basic and simple idea is to consider {\it partial concepts}: these are functions that can be {undefined} on certain parts of the space. When learning a partial concept, we assume that the source distribution is supported only on points where the partial concept is defined. 
    % This seemingly naive modification yields a unified language 
    % that captures a variety of data-dependent assumptions while retaining all the classical notions and parameters such as PAC learnability, VC dimension, Rademacher complexity, Littlestone dimension, {via a natural generalization to partial concepts}. 

This way, one can naturally express assumptions 
    on the data such as lying on a lower dimensional surface, or that it satisfies margin conditions.
    In contrast, it is not at all clear that such assumptions can be expressed by the traditional PAC theory using learnable total concept classes, and in fact we exhibit %incredibly 
    easy-to-learn partial concept classes which provably cannot be captured by the traditional PAC theory.
    This also resolves, in a strong negative sense, a question posed by~\citet*{Attias19robust}.
    
We characterize PAC learnability of partial concept classes
    and reveal an algorithmic landscape which is fundamentally different than the classical one. 
    For example, in the classical  PAC model, learning boils down to {\it Empirical Risk Minimization} (ERM). 
    This basic principle follows from {\it Uniform Convergence} 
    %~\citep*{vapnik:71} 
    and the {\it Fundamental Theorem of PAC Learning}~\citep*{vapnik:71,vapnik:74,blumer:89}.

In stark contrast, we show that the ERM principle    
    fails spectacularly in explaining learnability of partial concept classes. 
    In fact, we demonstrate classes that are incredibly easy to learn,  
    but such that any algorithm that learns them must use  an hypothesis space with unbounded VC dimension.
    We also find that the sample compression conjecture of Littlestone and Warmuth fails in this setting.
    Our impossibility results hinge on the recent breakthroughs 
    in communication complexity and graph theory by~\citet*{Goos15cis,Ben-David17sensitivity,Balodis:21}.

Thus, this theory features problems that cannot be represented in the traditional way and cannot be solved in the 
    traditional way. We view this as evidence that it might provide insights on the nature of learnability in realistic scenarios 
    which the classical theory fails to explain. 
    We include in the paper suggestions for future research and open problems in several contexts,
    including combinatorics, geometry, and learning theory.

\end{abstract}


\newpage

\pagestyle{empty}
\tableofcontents
\clearpage
\pagestyle{headings}

% % \thispagestyle{empty}
% % \tableofcontents

% \clearpage
\setcounter{page}{1}

\section{Introduction}
\label[section]{sec:intro}

% \paragraph{Later sections:}

% Disambiguation (negative and positive)

% Positive results for ERM when supports have finite VCdim too

% Maybe the negative result for reduction to PAC learner (but requires an aggregating reduction)

In many practical learning problems the data satisfy special properties that ease the learning process. 
    For example, imagine a learning task where the distance of the data from the decision boundary is bounded below by some margin $>0$, or learning tasks where the data lie on a low-dimensional surface. 
    (E.g.\ consider the task of classifying photographs of animals by whether the animal is a cat; arguably, the representations of such images lie on a low-dimensional subset of the space of all possible representations, most of which do not even represent a possible photograph.) 
%    (E.g.\ consider the task of classifying images of cats from images of dogs; arguably, the representations of such images lie on a low-dimensional subset of the space of all possible representations, most of which do not even represent a valid image.) 

Common approaches for modelling such tasks often use data-dependent assumptions 
    which are \underline{not} captured by the traditional theory of PAC learning:
    namely, they are not expressed by a PAC learnable concept class. 
    A classical example is the task of learning a high dimensional linear classifier with margin.
    Standard learning algorithms for this task, such as the classical Perceptron algorithm~\citep*{rosenblatt1958perceptron}, use hypothesis classes which are \underline{not} PAC learnable.
	Indeed, the Perceptron uses the hypothesis class of all linear classifiers, whose VC dimension 
	scales linearly with the Euclidean dimension, and is therefore not PAC learnable when the dimension is unbounded.
	To the best of our knowledge, the same applies to all learning algorithms in this context.\footnote{In fact, in \Cref{sec:geomargin} 
	we conjecture that any algorithm that learns this task satisfies that its image, i.e.\ the set of hypotheses it can output, has an unbounded VC dimension.}
    Thus, learnability of large-margin linear classifiers is not expressed as the PAC learnability 
 	of a natural concept class.

Consequently, the general framework for data dependent analysis 
	{deviated} from the traditional PAC setting while relying on additional modeling assumptions \citep*{Shawe-Taylor98data,Herbrich02luckiness}.
	Technically, this is done by introducing a data-dependent ``luckiness'' function
	which induces a (data-dependent) hierarchy of hypotheses (luckier hypotheses precede less lucky  ones, 
	as we discuss in more detail in \Cref{sec:luckiness}).
% 	This enables performing a (data-dependent) {\it Structural Risk Minimization} using that hierarchy.	
	For example, in the case of large margin linear classifiers,
	the luckiness of each linear separator is its margin with respect to the input sample. 
    While this framework has been successfully applied in various contexts, 
    it does not yield a crisp notion of learnability in the spirit of PAC learning.
	Moreover, the general results in this framework assume rather arcane technical conditions
	and, while these conditions suffice for proving bounds on a case-by-case basis in various situations, 	
	it is not clear whether they are necessary in general. %See \Cref{???} for a more technical discussion.

% Consequently, learnability under data-dependent assumptions has deviated from PAC
%     learning theory, while relying on additional modeling assumptions (see e.g., \citet*{Shawe-Taylor98data,Herbrich02luckiness}).
%     Indeed, more refined theoretical results have been obtained on a case-by-case basis in various practical situations, 
%     e.g.~\citep*{Koltchinskii05exponential,audibert:07,Bach18exponential,Nitanda19stochastic}. 
%     \shaymarg{Add more/different citations (these mostly focus on exponential rates which is orthogonal to our focus)}
%     {Such results however rely on additional modeling assumptions and do not provide a comprehensive and systematic theory of the learnable in the spirit of PAC learning.}

To address the above shortcomings, {we aim to develop a mathematical theory that is able to capture some of the above 
    features of practical learning systems, yet admits a complete characterization of learnability in the spirit of the PAC theory.}
    Towards this end, we take a complementary approach for modeling data-dependent assumptions:
	instead of modeling the algorithm's bias using a luckiness function,
	we extend the type of learning tasks and the notion of learnability. % by relaxing the notion of a concept class.
	As will be discussed below, this provides a natural generalization of the traditional learning theory,	
	which allows a unified treatment of data-dependent bounds and model-dependent bounds.

%\smallskip
\paragraph{Partial Concepts.}
The basic idea is simple: rather than learning a class of concepts $\H\subseteq \{0,1\}^{\X}$,
	where each concept $c\in \H$ is a \underline{total} function $c\colon \X\to\{0,1\}$,
	we consider partial concept classes $\H\subseteq \{0,1,\star\}^\X$,
	where each concept $c$ is a {\it \underline{partial}} function;
	specifically, if $x$ is such that $c(x)=\star$ then $c$ is {\it undefined} at $x$.
	The {\it support} of a partial concept $c\colon \X\to\{0,1,\star\}$ is the set 
	$\mathsf{supp}(c):=c^{-1}(\{0,1\}) = \{x\in \X: c(x) \neq\star\}.$

We then note that all the classical parameters such as VC dimension, Littlestone dimension, etc.,
	naturally extend to partial concept classes without modification.  In particular, a key quantity of interest in this work is the \emph{VC dimension} of a partial concept class $\H$,
	denoted by $\vcdim(\H)$, which is defined as the maximum size of a shattered set $U\subseteq\X$, where $U$ is \emph{shattered} if every 
	\underline{binary} pattern $u\in \{0,1\}^U$ is realized by some $h\in \H$ (i.e.\ $h\vert_U= u$).
	{This allows us to express formal connections between these parameters and learnability
	in a unified way that also applies to partial concept classes and covers various data-dependent assumptions.}
	For instance, the learnability of linear separators with margin then 
	reduces to merely noting that the VC dimension of the corresponding 
	partial concept class is bounded by a function of the margin.
	We note, however, that the algorithmic approach to establishing 
	this connection is necessarily quite different from the  
	algorithms typically used in the analysis of total concept classes, 
	as we discuss at length below.

\section{Results}\label[section]{sec:results}
In the next sections we give an overview of the main contributions in this work.
    Some of the formal statements rely on standard terminology (such as VC dimension, PAC learnability, etc)
    which is formally defined in the later technical sections.

\subsection{Expressivity} 
Allowing for partial concepts enables modelling data-dependent assumptions 
    in a natural way: indeed, given any such assumption, consider all legal samples $S$ which satisfy the assumption and define the corresponding partial class of all (partial) concepts such that every sample realizable by them is legal.
    
%    \shmarg{SH: Is the term ``partial class'' appropriate?  Shouldn't ``partial'' refer to the concepts, not the class?}
%    \shaymarg{I agree, for some reason this term comes up naturally for me... Any suggestions for alternatives?}
%    \shmarg{Ron: We can just say "partial concept class", as we do elsewhere.}
For example, consider again the task of learning a $\gamma$-margin linear separator in $\mathbb{R}^N$. 
    A sample $S\in \bigl(\R^N\times\{0,1\}\bigr)^n$ here is legal if the zero- and one-labelled
    examples are linearly separable with margin~$\gamma$, and the corresponding partial class $\H$
    consists of all partial concepts $h\colon \R^N\to\{0,1,\star\}$ such that 
	$h^{-1}(0)$ and $h^{-1}(1)$ are linearly separable with margin at least $\gamma$. 
	(See \Cref{sec:geomargin} for a more elaborate discussion of this partial class.)
	Similarly, one can easily model tasks in which the data lie on a low-dimensional subspace/manifold; 
	such assumptions are naturally captured by partial concepts which are undefined outside some such low-dimensional subset.

In contrast, it is not at all clear that these learning tasks can be expressed 
    in the traditional PAC model using a class of \underline{total} concepts.
    In fact, our first result demonstrates an incredibly easy-to-learn class of partial
	concepts that cannot be represented by {\it any} learnable class of total concepts.
	To state this result we need to formally define when a total concept class $\bar \H$ represents a partial concept class~$\H$:
	intuitively, we want that every learning task definable by $\H$ is also definable by~$\bar\H$.
	Formally, let us say that a total class~$\bar\H\subseteq\{0,1\}^\X$ {\it strongly\footnote{We will later define a weaker notion.} disambiguates} a partial class~$\H\subseteq\{0,1,\star\}^\X$ if every partial concept $ h\in \H$ is extended
	by some total concept $\bar h\in\bar\H$. Namely:
	\[(\forall h\in \H)(\exists \bar h\in \bar \H):\quad \bar h(x) = h(x) \quad\text{for all $x\in\supp(h)$}.\]
\begin{theorem}[Partial Concepts Are More Expressive Than Total Concepts]\label[theorem]{thm:expressive}
There exists a partial concept class $\H\subseteq\{0,1,\star\}^\nats$ whose VC dimension is 1 such that every total class $\bar \H\subseteq\{0,1\}^\nats$ which strongly disambiguates $\H$ must have an infinite VC dimension, i.e.\ $\vcdim(\bar \H)=\infty$.
\end{theorem}
% \begin{theorem}[Partial Concepts Are More Expressive Than Total Concepts]\label{thm:expressive}
% There exists a partial concept class $\H\subseteq\{0,1,\star\}^\nats$ whose VC dimension is 1 such that every class of total concepts $\bar \H\subseteq\{0,1\}^\nats$ that \underline{strongly\footnote{We will later define a weaker notion.} disambiguates} $\H$,
% namely:
% \[(\forall h\in \H)(\exists \bar h\in \bar \H):\quad \bar h(x) = h(x) \quad\text{for all $x\in\supp(h)$},\]
% must have an infinite VC dimension, i.e.\ $\vcdim(\bar \H)=\infty$.
% \end{theorem}
As we will discuss below, the above partial class $\H$ is easy to learn:
    since its VC dimension is one, we show there is an algorithm that PAC learns whenever the examples have distribution  
    with support contained in the support of a partial concept from $\H$,
    and the sample complexity is $O(\log(1/\delta)/\epsilon)$,
    where $\epsilon,\delta$ are the standard accuracy and confidence parameters.
    However, the above theorem implies that if one tries to 
    extend each partial concept in $\H$ by a total concept 
    by disambiguating the~$\star$'s in it, 
    then one must end up with a class whose VC dimension is unbounded, and hence is \underline{not} PAC learnable.

\Cref{thm:expressive} answers an open question posed by \citet*{Attias19robust}.
    It might be interesting to note that our proof of it
    exploits a surprising connection with the theory of communication complexity.
    In particular, it hinges on the recent breakthroughs concerning
    the {clique vs.\ independent set problem} by~\citet*{Goos15cis,Ben-David17sensitivity,Balodis:21}.
    We discuss this further in \Cref{sec:disambiguation-intro} below.

% 1. Definitions, informal examples, not clear how to express them as total classes

% Define disambiguation in the intro, 
% but just using the stronger notion of 
% filling in all the stars, so we don't have to 
% motivate the finite sets issue.

% And maybe just be vague about the notion of realizability.

% 2. Theorem 1: There are learnable partial concept classes such that no disambiguation is learnable.

\subsection{PAC Learnability}
\label[section]{sec:pac-intro}

We next present a characterization of the PAC learnable partial concept classes.
    But first, we should clarify the definition of PAC learning in this context. 
    Let us begin with the noiseless and realizable setting: 
    intuitively, we want realizability to express the premise that the data
    drawn from the source distribution satisfy the data-dependent assumptions captured by the partial concept class $\H$. This gives rise to the following definition:
    a distribution $\PXY$ on $\X \times \{0,1\}$ is {\it realizable by $\H$} if almost surely (i.e., with probability $1$), a sample $S=((x_i,y_i))_{i=1}^{n}\sim \PXY^n$ (for any $n$) is realizable by some partial concept $h\in \H$: that is, $\{x_i\}_{i=1}^{n}\subseteq\supp(h)$, and $h(x_i)=y_i$ for all $i\leq n$.
    The connection between this definition of a realizable distribution and the one used in the classical PAC model is clarified in \Cref{lem:def-realizable}.
    For a partial concept $h$ and a distribution $\PXY$ on $\X \times \{0,1\}$, 
    we define the \emph{prediction error}: 
    $\er_{\PXY}(h) := \PXY( \{(x,y) : h(x) \neq y \})$.
    To be clear, this means we \emph{always} count 
    the case $h(x) = \star$ as a prediction mistake.
    
\begin{definition}[PAC Learnability]
\label[definition]{defn:pac-learnable}
A partial concept class $\H$ is \emph{PAC learnable} 
if, for every $\eps,\delta \in (0,1)$, 
there exists a finite $\SC(\eps,\delta) \in \nats$ 
and a learning algorithm $\alg$ such that, 
for every distribution $\PXY$ on $\X \times \{0,1\}$ 
realizable w.r.t.\ $\H$, for $S \sim \PXY^{\SC(\eps,\delta)}$, 
with probability at least $1-\delta$, 
%\[\er_{\PXY}(\alg(S)) = \PXY( (x,y) : (\alg(S))(x) \neq y ) \leq \eps.\]
\[\er_{\PXY}(\alg(S)) \leq \eps.\]
The value $\SC(\eps,\delta)$ is called the 
\emph{sample complexity} of $\alg$, and the 
\emph{optimal} sample complexity is the 
minimum achievable value of $\SC(\eps,\delta)$ 
for every given $\eps,\delta$.
\end{definition}
In Section~\ref{sec:agnostic} we define learnability in the agnostic case in a similar manner, 
    using the convention that $\star$'s are always treated as errors.
    
\smallskip  
    
We begin by addressing the following fundamental question:
\begin{center}
Which Partial Concept Classes Are PAC Learnable and How?    
\end{center}
    The {\it Fundamental Theorem of PAC Learning} asserts that
    a \underline{total} concept class $\H$ is PAC learnable if and only if its VC dimension is finite
    \citep*{vapnik:74,blumer:89,shalev-shwartz:14}.
	This theorem also yields the celebrated {\it Empirical Risk Minimization} principle: \underline{any} algorithm which outputs an hypothesis $h\in \H$ which minimizes the empirical error learns $\H$. Such algorithms are called Empirical Risk Minimizers (ERMs).
%	Moreover, the learning rate of any ERM is near-optimal (up to log factors).
    In the following theorem we show that the characterization of PAC learnability
    in terms of the VC dimension extends to partial concept classes:

\begin{theorem}[A Characterization of PAC Learnability.]
\label[theorem]{thm:pac-learnability}
The following statements are equivalent for any partial concept class $\H\subseteq\{0,1,\star\}^\X$. 
\begin{itemize}
    \item $\vcdim(\H) < \infty$.
    \item $\H$ is PAC learnable.
    \item $\H$ is agnostically PAC learnable.
\end{itemize}
%in the realizable setting, and is $\tilde{\Theta}(\sqrt{\vcdim(\H)/n})$ in the 
%agnostic setting.
\end{theorem}

It is important to note that our proof of \Cref{thm:pac-learnability}  
    is fundamentally different from the classical uniform-convergence-based argument, 
    and it does not yield any version of the ERM principle. (We discuss this in more detail below.)
    Instead, our proof hinges on a combination of sample compression 
    and a variant of the {\it 1-Inclusion-Graph Algorithm} due to  \citet*{haussler:94}.
	The obtained algorithm is transductive, in the sense that its output hypothesis is not computed explicitly:
	rather, given any test point, it uses the entire training set to compute its label 
	(as is the case, e.g., for the $k$-Nearest Neighbor Algorithm).
	An interesting property of our algorithm (as well as other transductive algorithms) is that the complexity of the model (hypothesis) 
	it outputs can increase with the size of the input sample.
	Below we show that in general, this property is inevitable:
	there exist partial classes $\H$ with $\vcdim(\H)=1$ such that any algorithm which PAC learns them	must satisfy that its range (i.e.\ the set of hypotheses it can output)
	has an unbounded VC dimension.

\subsection{Failure of Traditional Learning Principles}

One of the conceptual contributions of the traditional PAC learning theory 
    is the ERM principle: any learnable class $\mathcal{H}$ is learned by any algorithm which outputs 
	a concept $h\in \mathcal{H}$ that minimizes the empirical error on the training set. 
    Moreover, any ERM algorithm achieves the optimal %uniform learning rate,
    sample complexity, up to lower order factors~\citep*{vapnik:74,blumer:89}.
    This simple principle is attractive from an algorithmic perspective as
	it reduces a learning problem (in which the goal is to minimize an unknown function $\er_{\PXY}$), to an optimization problem (in which the goal is to minimize the known empirical loss).
    % This is summarized in the literature by the Fundamental Theorem of PAC Learning (see e.g.\ the book by \citet*{shalev-shwartz:14}) which asserts that a (total) concept class has a finite VC dimension
    % if and only if it is PAC learnable and if and only if it satisfies the ERM principle.

However, recent machine learning breakthroughs demonstrate important phenomena that 
	lack explanations, and sometimes even contradict 
	conventional wisdom (see e.g.~\citep{Zhang17rethinking,Nagarajan19uniform,Maennel20,Ilya20weights,Feldman20,Brown20memorization}).
	For example, consider the modern approach of training very rich models 
	to (and often beyond) the point of complete interpolation of the training-set.  
	In the lens of traditional learning theory, this would constitute a clear example of overfitting;
	however, this approach achieves excellent results in practice when implemented in deep neural networks,
	as well as in other hypothesis spaces such as ensembles of decision trees, kernel machines, and minimum norm linear regressors~\citep{Belkin19double,Barak20double}.

One reason for the incapacity of traditional generalization theory 
    to model modern machine learning is because the traditional theory reduces learning to {an empirical risk minimization task over \underline{not-too-large} hypothesis spaces}. 
    In contrast, modern algorithms typically train hypotheses with a huge number of parameters.

Thus, it is interesting to seek extensions of the classical PAC theory which necessitate alternative
    principles beyond ERM. \Cref{thm:pac-learnability} implies that the equivalence between finite VC dimension
    and PAC learnability extends to partial concept classes.
    However, we next demonstrate that the ERM principle has no useful analogue here.
	In order to address this, we first need to specify what empirical risk minimization even means in this context.

\paragraph{Naive ERM Fails.}
One natural option is to define an empirical risk minimizer to be any algorithm 
    which outputs a \underline{partial} concept $h\in \H$ that minimizes the empirical loss (i.e., that interpolates the input data in the realizable case).
	However, it is easy to see that such algorithms fail to learn even very simple classes: 
\begin{proposition} 
\label[proposition]{prop:partial-ERM-failure}
There exists a partial concept class $\H$ with $\vcdim(\H)=0$ such that 
any proper algorithm (i.e., which outputs a partial concept from $\H$) 
fails to PAC learn $\H$.
\end{proposition}
\begin{proof}[Sketch]
	Let $n\in\mathbb{N}$ be even and consider the class $\H\subseteq\{0,1,\star\}^{[n]}$ defined by	
	\[
	\H=\{h_A : A\subseteq[n], \lvert A\rvert = n/2\},\, \text{where}\,
	h_A(x) = \begin{cases}
	0 &x\in A,\\
	\star &x\in [n]\setminus A.
	\end{cases}\]
	Note that $\H$ has VC dimension $0$ and that it is trivially PAC learnable by the algorithm which always outputs the all-zero function $h_0\equiv 0$ (which is \underline{not} in $\H$).
	However, any algorithm which is restricted to outputting partial concepts from $\H$ 
	(and in particular any such ERM)
	will fail to learn this class unless it gets at least $\Omega(n)$ examples;
	indeed, this follows by a similar argument as in the standard no-free-lunch argument for VC classes:
	let the target concept $c\in \H$ be drawn uniformly at random and let the marginal distribution be uniform over $\mathsf{supp}(c)$; {if the learner observes fewer than $n/4$ examples, and must output a hypothesis $\hat{h}_n \in \H$, it must \emph{guess} the locations of at least $n/4$ elements of $\mathsf{supp}(c)$ not observed in the data, and very likely will guess incorrectly for a constant fraction of them.}
	An infinite variant of this construction yields a $0$-dimensional class that cannot be PAC learned by \underline{any} ERM: namely, on 
	$\X = \nats$, let $\H$ be all $\{0,\star\}$-valued functions $h$ with, $\forall t \geq 2$, exactly $2^{t-2}$ points $x \in [2^t] \setminus [2^{t-1}]$ with $h(x)=0$; then the above argument can be applied 
	in any region $[2^t] \setminus [2^{t-1}]$ to show $2^{t-3}$ examples 
	do not suffice for proper learners, for any $t$.
\end{proof}

\paragraph{General ERM fails.}
A stronger (and natural) family of empirical risk minimization algorithms in this context
	are algorithms which learn $\H$ by performing empirical risk minimization 
	over an appropriate class $\H'\subseteq\{0,1\}^\X$.
	For example, for the class $\H$ discussed above, 
	we can pick $\H' = \{h_0\}$ to be the class	consisting only of the all-zero function. 
	Observe that indeed any ERM for $\H'$ successfully learns $\H$.
	The existence of such an $\H'$ yields a reduction from PAC learning $\H$ to PAC learning $\H'$. Does the ERM principle apply in this sense? 
	That is:
	\begin{center}
	    Given a partial concept class $\H$,	
	    does there always exist a class $\H'$ such that\\ \underline{any} ERM w.r.t $\H'$ learns $\H$?
	\end{center}
	Can the task of learning a given partial class $\H$ be reduced to the task of empirical risk minimization over some total class $\H'$? 
	The following theorem provides a negative answer (Proof in Section~\ref{sec:erm-failure}):
\begin{theorem}[Failure of Empirical Risk Minimization]
\label[theorem]{thm:total-ERM-failure-intro}
There exists a partial concept class $\H$ with $\vcdim(\H) = 1$ 
such that, for any total concept class $\bar{\H}$, there exists an ERM algorithm for $\bar{\H}$ that 
is not a PAC learning algorithm for $\H$.  
%Indeed, its prediction error can be made arbitrarily close to $1$ 
%for any sample size $m$ by choosing 
%a distribution $\PXY$ (realizable w.r.t.\ $\H$) depending on $m$.
%Further, this applies even if we allow the learner to pick $\bar{\H}$ based on the size of the input sample.
\end{theorem}

The next theorem (also proved in Section~\ref{sec:erm-failure}) shows that regardless of ERM, a partial concept class may require that any learning algorithm that outputs total concepts must have a large image (in the sense of VC dimension).

\begin{theorem}
\label[theorem]{thm:vcdim-image}
There exists a partial concept class 
$\H$ with $\vcdim(\H)=1$ 
such that any learning algorithm 
$\alg$ that only outputs 
total concepts must have 
image with infinite VC dimension.
\end{theorem}

\paragraph{Algorithmic Principles That Complement ERM?}

Let us conclude this section with a suggestion for future work:
    {\it Explore for general algorithmic principles
     that apply in this more general setting and complement the traditional ERM Principle.} As noted above, while \Cref{thm:pac-learnability} 
     asserts that indeed every partial VC class is PAC learnable,
     our proof of it does not seem to give rise to a general
     principle in the spirit of ERM.

\subsection{The Landscape of Partial VC Classes}

In this section we investigate basic properties of partial VC classes and
   their relationship with total classes.
   We begin by exhibiting two learning-theoretical differences 
   between partial and total classes: in the contexts of sample compression (\Cref{sec:compression}) and differentially private learning (\Cref{sec:DP}).
   Then, in \Cref{sec:disambiguation-intro} we investigate the following
   question which is central to this work: 
   given a partial class $\H$ with VC dimension $d$,
   can one find a ``small'' class $\bar \H$ which {\it disambiguates} $\H$?
   We provide negative as well as positive results in this context.
   %Finally, in \Cref{sec:closure} we discuss closure properties 
   %of partial VC classes.

\subsubsection{Sample Compression Schemes}\label[section]{sec:compression}

Sample compression is a fundamental technique for proving generalization bounds.
    \citet*{littlestone1986relating} proposed it as an intuitive, algorithm-dependent, technique 
    for establishing PAC learnability of concept classes of interest. 
    Later works have demonstrated its usefulness in various statistical learning settings, 
    including semi-supervised and even unsupervised learning~\citep*{graepel:05,hanneke:15a,david:16,Kontorovich17nearest,Gottlieb18compression,hanneke:19f,Ashtiani20density}.
    %\shaymarg{Add Citations}
    In fact, \citet*{david:16} established that this technique
    is in a sense universal by proving that learnability is equivalent to
    compressibility in a general and abstract learning setting.

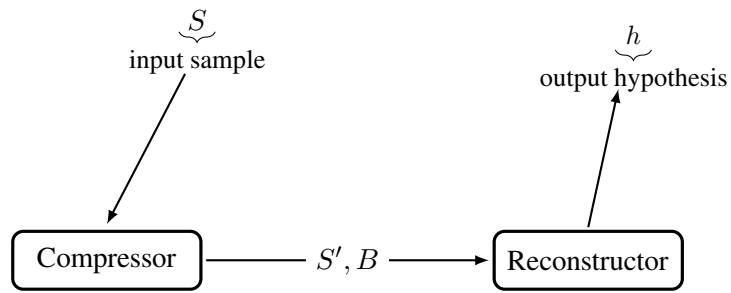
\begin{figure}
\centering
    \textbf{A pictorial definition of a sample compression scheme}\par\bigskip   
\begin{tikzpicture}[scale=0.8]

\node[inner sep=0pt,punkt] (bob) at (7,0) {Reconstructor};

\node[inner sep=0pt,punkt] (alice) at (-1,0){Compressor};

\draw[->,thick] (0.65,0) -- (5.35,0)
    node[midway,fill=white] {$S',B$};

%\draw [decorate,decoration={brace,amplitude=4pt,mirror},yshift=0pt]
%(2.5,-0.2) -- (3.4,-0.2) node [black,midway,yshift=-0.5cm]
%{\small{ a subsample and an additional binary string}};

\node[inner sep=0pt] (sample) at (0.5,4){$S$};
\draw[->,thick] (0.3,3.1) -- (-1,0.6);

\draw [decorate,decoration={brace,amplitude=4pt,mirror},yshift=0pt]
(0.2,3.8) -- (0.8,3.8) node [black,midway,yshift=-0.4cm]
{\small input sample};

\node[inner sep=0pt] (reconstruction) at (7.75,3.75){\small $h$};

\draw [decorate,decoration={brace,amplitude=4pt,mirror},yshift=0pt]
(7.45,3.5) -- (8.05,3.5) node [black,midway,yshift=-0.4cm]
{\small output hypothesis};

\draw[->,thick] (7,0.55) -- (7.5,2.85);
\end{tikzpicture}
\caption{
$S'$ is a subsample of $S$ and $B$ is a binary string of additional information.}
\label[figure]{fig:comp}
\end{figure}

A sample compression scheme can be seen as a protocol between a {\it compressor} $\kappa$ and a {\it reconstructor} $\rho$ (see \Cref{fig:comp}):
the compressor gets the input sample $S$, 
from which she picks a small subsample $S'$. 
The compressor sends to the reconstructor the subsample $S'$, 
along with a short binary string $B$ 
of additional information: i.e., $(S',B) = \kappa(S)$. 
The reconstructor then, 
based on $S'$ and $B$, outputs a concept~$h=\rho(S',B)$.
For a given partial concept class $\H$, 
we say $(\kappa,\rho)$ is a compression scheme \emph{for} $\H$ 
if, for all finite data sequences $S$ realizable w.r.t.\ $\H$, 
the above $h=\rho(\kappa(S))$ returned by the reconstructor 
is correct on the entire sample $S$
%The ``correctness'' criterion is that the error of $h$ on the entire sample $S$ 
(including the examples in $S$ that were not sent to the reconstructor). 
%should be small (or competitive with the loss of the best hypothesis in $\H$ in agnostic settings). 
The size of the compression scheme on $S$ is defined to be $|B|+|S'|$;
the size of the compression scheme for a given sample size $m$ is 
the maximum size $|B|+|S'|$ over all $S \in (\X \times \{0,1\})^m$.
The formal definition is given in \Cref{sec:complexities} 
in Definition~\ref{defn:compression-scheme}.

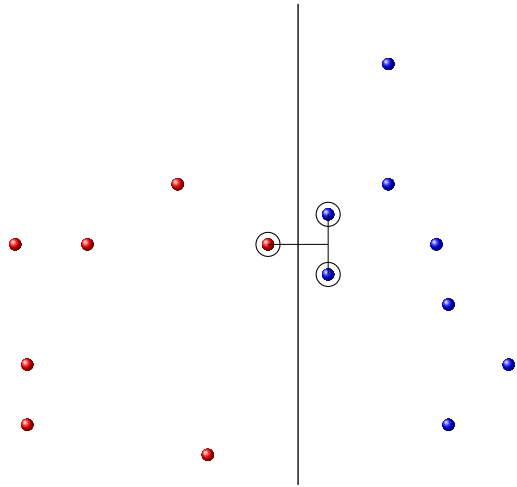
\begin{figure}
\centering
\textbf{\small Support Vector Machine as a sample compression}\par\bigskip   
\begin{tikzpicture}[scale=0.8] 
\shade[shading=ball, ball color=red]  (8,0.5) circle (.1);
\shade[shading=ball, ball color=red]  (4.8,4) circle (.1);
\shade[shading=ball, ball color=red]  (5,2) circle (.1);
\shade[shading=ball, ball color=red]  (5,1) circle (.1);
\shade[shading=ball, ball color=red]  (6,4) circle (.1);
\shade[shading=ball, ball color=red]  (9,4) circle (.1);%
\shade[shading=ball, ball color=red]  (7.5,5) circle (.1);
\shade[shading=ball, ball color=blue]  (11.8,4) circle (.1);
\shade[shading=ball, ball color=blue]  (13,2) circle (.1);
\shade[shading=ball, ball color=blue]  (12,1) circle (.1);
\shade[shading=ball, ball color=blue]  (10,3.5) circle (.1);%
\shade[shading=ball, ball color=blue]  (10,4.5) circle (.1);%
\shade[shading=ball, ball color=blue]  (12,3) circle (.1);
\shade[shading=ball, ball color=blue]  (11,5) circle (.1);
\shade[shading=ball, ball color=blue]  (11,7) circle (.1);

\v<2->{\draw  (9,4) circle (.2);}
%\v<2->{\draw  (7.5,5) circle (.2);}
%\v<2->{\draw  (8,0) circle (.2);}

\v<2->{\draw  (10,3.5) circle (.2);}
\v<2->{\draw  (10,4.5) circle (.2);}
%\v<2->{\draw  (11,7) circle (.2);}
%\v<2->{\draw  (12,1) circle (.2);}

\v<3->{\path[draw, line width=0.5pt] (9.5,0) -- (9.5,8);}
\v<3->{\path[draw, line width=0.1pt] (9,4) -- (10,4);}
\v<3->{\path[draw, line width=0.1pt] (10,3.5) -- (10,4.5);}
\end{tikzpicture}
\caption{
\small The input sample $S$ consists of all red and blue points;
the separating hyperplane with maximum margin 
is determined by the subsample of the $d+1$ {\it support points}. 
Thus, the compression scheme uses only these points.}
\label[figure]{fig:svm}
\end{figure}

A classical example of an algorithm that can be presented as a compression scheme is the {\it Support Vector Machine} algorithm in $\mathbb{R}^d$. 
Here, the compressor sends to the reconstructor the~$d+1$ support vectors which determine the maximum margin separating hyperplane (see Figure~\ref{fig:svm}).

\paragraph{Warmuth's \$600 Sample Compression Question.}
    
Sample compression is the topic of one of the longest-standing and most well-studied open problems in learning theory:
    \begin{center}
    {\it Does every concept class $\H$ have a compression scheme of size $O(\vcdim(\H))$?}
    \end{center}
This question has been studied since the pioneering work by \citet*{littlestone1986relating},
    and later~\citet*{Warmuth03open} even announced a \$600 reward for solving it!
    For a discussion of this question in a broader context, 
    we refer the reader to the book by~\citet*{Wigderson2019}.

It is therefore interesting to explore sample compression schemes in the setting of partial concept classes.
    Perhaps surprisingly, it turns out that in this context the
    answer to the sample compression question is negative (in a strong sense).
    On the positive side, we show that every partial VC class has a compression scheme 
    whose size scales logarithmically with the input sample size:
    
\begin{theorem}[Sample Compression for Partial Concept Classes]\label[theorem]{thm:compression}\ \\
\vspace{-5mm}
\begin{enumerate}
    \item Let $\H$ be a partial concept class. Then, there exists a sample compression scheme for~$\H$
of size $\tilde O(\vcdim(\H)\log (m))$, where $m$ is the size of the input sample.
    \item There exists a partial concept class $\H$ with $\vcdim(\H)=1$ such that any sample compression scheme for $\H$
must have size $\Omega((\log(m))^{1-o(1)})$, where $m$ is the size of the input sample,
    {and the $o(1)$ term vanishes as $m\to\infty$.}
%    \item For every $d \in \nats$, there exists a partial concept class $\H$ with $\vcdim(\H)=d$ such that any sample compression scheme for $\H$
%must have size $\Omega(d(\log(m/d))^{1-o(1)})$, where $m$ is the size of the input sample.
In particular, the bounded-size sample compression conjecture is \textbf{false} for partial concept classes.
\end{enumerate}
\end{theorem}

The proof of this 
result is in Section~\ref{sec:erm-failure}.
\Cref{thm:compression} demonstrates a stark difference between total and partial VC classes:
    \citet*{Moran16compression} proved that every \underline{total} VC class has a sample compression scheme whose size is bounded by a function of the VC dimension. 
    By Item 2 above, this result does not extend to partial VC classes,
    even with VC dimension one.

% \begin{theorem}
% \label{cor:compression}
% The bounded-size sample compression conjecture is \textbf{false} for partial concept classes:
% there exists a partial concept class $\bar H$ of $\vc(\bar H)=1$ such that any sample compression scheme for $\H$
% must have size $\Omega(\log m)$, where $m$ is the size of the input sample.
% \end{theorem}

% \begin{theorem}
% Let $\H$ be a partial concept class. Then, there exists a sample compression scheme for~$\H$
% of size $O(\vc(\H)\log m)$.
% \end{theorem}

\subsubsection{Littlestone Dimension vs Private Learning}\label[section]{sec:DP}

Differentially private PAC learning is an additional setting which 
    demonstrates a curious difference between partial classes and total classes.
    
Differential privacy (DP) \citep*{DMNS06} is a sound theoretical approach to reason about privacy in a precise and quantifiable fashion. 
    It  has become the gold standard of statistical data privacy~\citep*{DR14} 
    and been implemented in practice, notably by Google~\citep*{Erlingsson14google}, Apple \citep*{apple,apple1}, and in the 2020 US census~\citep*{Census20}.

A recent line of work revealed a qualitative characterization of DP-learnability in the PAC model:
    A total concept class $\H$ can be PAC learned by a DP-algorithm
    if and only if its {\it Littlestone dimension} $\Ldim(\H)$ is finite~\citep*{AlonLMM19,Gonen19privateonline,Bun20equivalence,Ghazi20efficient}.
    (The Littlestone dimension is a combinatorial parameter which arises
    in the context of online learning, see \Cref{sec:complexities} for a formal definition.)
    It is therefore natural to ask whether this characterization 
    extends to partial concept classes:
\begin{question}[Private PAC Learnability]
Does the characterization of differentially private PAC learning extend to partial classes?:
Let $\H$ be a partial class. 
Is it the case that $\H$ is PAC learnable by a differentially private algorithm if and only if it has a finite Littlestone dimension?
\end{question}
Despite the fact that natural partial classes with finite Littlestone dimension
    are known to be DP learnable (e.g.\ halfspaces with margin~\citep*{Nguyen:20}),
    it is not clear how to generally prove either of the implications 
    ``$\Ldim(\H)<\infty\implies$ $\H$ is DP-learnable''
    or ``$\H$ is DP-learnable $\implies \Ldim(\H)<\infty$''.

The known proofs of the direction ``$\Ldim(\H)<\infty\implies$ $\H$ is DP-learnable''
    for total concept classes \citep*{Bun20equivalence,Ghazi20efficient}
    utilize (among other things) the ERM principle and uniform convergence which, 
    as discussed earlier, is not satisfied by partial concept classes.
    
As for the direction ``$\Ldim(\H)=\infty \implies \H$ is \underline{not} DP-learnable '',
    the very first step of the proof by~\citet*{AlonLMM19}
    fails for partial classes:
    the proof proceeds by first reducing an arbitrary class with an unbounded Littlestone dimension
    to the class of one-dimensional thresholds, 
    and then proving that one-dimensional thresholds 
    are not privately PAC learnable.

The reduction to one-dimensional thresholds boils down to a combinatorial
    parameter called the {\it threshold dimension}:
    the threshold dimension of a class $\H$, denoted by $\td(\H)$, 
    is the maximum integer $d$ for which there exist $x_1,\ldots, x_d\in \X$ and $h_1,\ldots h_d\in\H$ such that $h_i(x_j)=\ind[i \leq j]$.
    For total concept classes $\H$ it is known\footnote{It is also known that $\Ldim(\H) \geq \lfloor \log(\td(\H))\rfloor$, but this inequality extends also to partial classes with the same proof (see \citet*{AlonLMM19}). } $\td(\H) \geq \lfloor \log(\Ldim(\H))\rfloor$; this essentially implies that any class with a large Littlestone dimension
    contains a large subclass of thresholds.
    Interestingly, this relation fails to extend to partial classes, as shown in the next theorem (proved in Section~\ref{sec:erm-failure}):
\begin{theorem}\label[theorem]{thm:threshold}
There exists a partial concept class $\H$ with $\td(\H)\leq 2$ but $\LD(\H)=\infty$.
\end{theorem}

\subsubsection{Disambiguations}\label[section]{sec:disambiguation-intro}

We next present one of the main focuses of this work
    which concerns the following questions:
    Can partial VC classes be represented by total VC classes?
    Relatedly, can one reduce the task of learning a given partial VC class
    to the task of learning a total VC class?
    We begin with the following central definition of disambiguation:
    
\begin{definition}[Disambiguation]\label[definition]{defn:disambiguation-intro}
A total concept class $\bar{\H}$ is a special type of 
partial concept class such that every $h \in \bar{\H}$ 
has range $\{0,1\}$: i.e., is a total concept.
A total concept class $\bar{\H} \subseteq \{0,1\}^{\X}$ 
is said to \emph{disambiguate} a partial concept class $\H$ 
if every finite data sequence 
$S \in (\X \times \{0,1\})^*$
realizable w.r.t.\ $\H$ 
is also realizable w.r.t.\ $\bar{\H}$.  In this case, 
$\bar{\H}$ is called a \emph{disambiguation} of $\H$.
%\shaymarg{Has the term data-sequence been defined? Shall we define it explicitly?}
\end{definition}

Note the difference between \Cref{defn:disambiguation-intro} and the definition
    used in \Cref{thm:expressive}: the latter poses a stricter requirement,
    namely that each partial concept in $\H$ is extended by some total concept in~$\bar\H$. We note that the two definitions are equivalent when $\X$ is finite (more generally, when $\supp(h)$ is finite for every $h \in \H$), and are essentially equivalent when $\X$ is countable.\footnote{In the sense that whenever $\H$ can be disambiguated according to the weaker definition by $\bar\H$ then it can also be disambiguated according to the stronger definition 
    by $\bar\H'$ such that $\vcdim(\bar\H)=\vcdim(\bar\H')$. }

\Cref{defn:disambiguation-intro} is more suitable in the context of learning 
    because it suffices to guarantee that every PAC learner for $\bar\H$ is a PAC learner for $\H$, and hence reduces the task of PAC learning the partial class $\H$ to PAC learning the total class $\bar\H$. One could further relax \Cref{defn:disambiguation-intro}
    by allowing errors and by only requiring to disambiguate short samples,
    in a way that implies that a learner for $\bar\H$ is a weak learner for $\H$.
    However, the next proposition implies that such relaxations are essentially equivalent to \Cref{defn:disambiguation-intro}.
        %Towards this end we introduce the following notation:
    % % For $n\in\nats$ and $\eps>0$, we say that a partial class $\H$ is $(n,\eps)$-disambiguated by a total class $\bar\H$ if for every sample $\bigl((x_i,y_i)\bigr)_{i=1}^n\in(\X\times\{0,1\})^n$ which is realizable by $\H$, there exists $\bar h\in \bar\H$ such that 
    % % \[\frac{1}{n}\sum_{i=1}^n1[\bar h(x_i)\neq y_i] \leq \eps.\]
    % % The following proposition shows asserts that if $\H$ is $(n,\eps)$-disambiguated by $\bar\H$ such that $\vcdim(\H) = \tilde O(\eps\cdot n)$ then $\H$ is VC-disambiguable.
    % % That is, to show that $\bar H$ is VC-disambiguable, it is enough
    % to demonstrate a total class $\bar H$ 
\begin{proposition}[Approximate Disambiguation $\implies$ Disambiguation]\label[proposition]{prop:compact}
Let $\H$ be a partial class and let $\gamma>0$. 
Assume that there exists a total class $\bar\H$
with $\vcdim(\bar\H) = d$ that ``weakly disambiguates'' $\H$ in the following sense: for every sample $S = \{(x_i,y_i)\}_{i=1}^n$ realizable by $\H$ of size $\lvert S\rvert = n = O(\frac{d}{\gamma^2})$ there exists $\bar h\in \bar\H$ such that 
    \[\hat{\er}_{S}(\bar h):=\frac{1}{n}\sum_{i=1}^n \ind [\bar h(x_i)\neq y_i] \leq \frac{1-\gamma}{2}.\]
Then, $\H$ can be disambiguated (in the sense of \Cref{defn:disambiguation-intro}) by a total class whose VC-dimension is at most $\tilde O(\frac{d\cdot d^\star}{\gamma^2})$, where $d^\star \leq 2^{d+1}$ is the dual VC dimension of $\bar \H$.
\end{proposition}
\Cref{prop:compact} might be viewed as a 
    kind of compactness theorem; for example, 
    it implies that in order to disambiguate $\H$ it suffices to 
    represent only the samples realizable by $\H$ of size at most $100d$
    by a total class~$\bar \H$ with $\vcdim(\bar \H) = d$. The proofs of all the results in this subsection appear in Section~\ref{sec:proofs-disambiguation}.
%    \shay{It might be fun to find the optimal constant here.}

The following result demonstrates a one-dimensional class which cannot be disambiguated
    while retaining a bounded VC dimension.

\begin{theorem}[A VC Class Which Cannot be Disambiguated]
\label[theorem]{thm:disambiguation-negative1}
For any $n\in\mathbb{N}$ there exists a partial concept class $\H_n\subseteq\{0,1,\star\}^{[n]}$ 
with $\vcdim(\H_n) = 1$ and $\td(\H_n)\leq 2$ such that 
any disambiguation $\bar{\H}$ of $\H_n$ has size 
at least $n^{(\log(n))^{1 - o(1)}}$, 
where the $o(1)$ term tends to $0$ as $n \to \infty$. In particular, this implies 
$\Ldim(\bar{\H}) \geq \vcdim(\bar{\H}) \geq (\log(n))^{1 - o(1)}$, and shows that for infinite $\X$ there exists $\H_{\infty}\subseteq\{0,1,\star\}^\X$ with $\vcdim(\H_{\infty}) = 1$ and $ \td(\H_{\infty})\leq 2$, while $\Ldim(\bar \H) = \vcdim(\bar{\H})=\infty$ for every disambiguation $\bar{\H}$ of $\H_{\infty}$.
\end{theorem}
Below, in \Cref{cor:disambiguation-positive-intro} we show that the bound in \Cref{thm:disambiguation-negative1} is nearly tight.
\Cref{thm:disambiguation-negative1} resolves, in a strong negative sense, an open problem presented by \citet*{Attias19robust}, which sought a disambiguation whose VC dimension is bounded by a (linear) function of $\vcdim(\H)$. 
Further, \Cref{thm:disambiguation-negative1} is our workhorse for proving 
    the impossibility results discussed in the previous sections
    regarding expressivity (\Cref{thm:expressive}) the failure of the ERM principle (\Cref{thm:total-ERM-failure-intro}), the image of any learning algorithm (\Cref{thm:vcdim-image}), sample compression schemes (\Cref{thm:compression}), and private PAC learning (\Cref{thm:threshold}).

    Interestingly, its proof hinges on a recent breakthrough in communication complexity 
    and its implications in graph theory: ~\citet*{Goos15cis,Ben-David17sensitivity,Balodis:21}.
    Despite the advantage that our proof of \Cref{thm:disambiguation-negative1} is short and simple, it unfortunately provides only little insight on the structure of the concluded class $\H$. In part, this is due to the complexity of the relevant result in graph theory, which is obtained by a series of reductions, some of which are unintuitive. It will be interesting to exhibit a natural partial VC class which demonstrates this separation. Towards this end, we propose a geometric candidate in \Cref{sec:geomargin}.

\paragraph{A Sauer-Shelah-Perles Lemma for Partial VC Classes?}
So far we discussed several differences between partial and total VC classes.
    All of these differences boil down to \Cref{thm:disambiguation-negative1}.
    We next investigate which properties of total VC classes are {\it retained} by partial classes.

Arguably the most basic property of VC classes is manifested 
    by the Sauer-Shelah-Perles Lemma (SSP) \citep*{sauer:72}.
    This lemma bounds the cardinality of a class $\H\subseteq\{0,1\}^n$ with $\vcdim(\H) = d$ by 
    \[\lvert\H\rvert \leq {n \choose \leq d} .\]
    {\it Is there an analogue of the SSP Lemma for partial classes?}
    An immediate and direct generalization of it to partial classes would be 
    that $\lvert\H\rvert \leq {n \choose \leq \vcdim(\H)}$ for every partial class $\H\subseteq\{0,1,\star\}^\X$. 
    However it is easy to see that this is false, as witnessed e.g.\ 
    by the class $\H = \{0,\star\}^n$ which satisfies $\vcdim(\H)=0$ and $\lvert \H\rvert = 2^{n}$. A more mature candidate for extending the SSP Lemma to partial classes is via     
    disambiguations: 
\begin{center}
{\it Can every $\H\subseteq\{0,1,\star\}^n$ be disambiguated
    by a total class $\bar\H\subseteq\{0,1\}^n$ such that $\lvert\bar\H\rvert \leq {n \choose \leq \vcdim(\H)}$?}
\end{center}
    Indeed, the above class $\H=\{0,\star\}^n$ is disambiguated by $\bar\H=\{0^n\}$ which  satisfies this inequality.
    Unfortunately, \Cref{thm:disambiguation-negative1} also refutes this version:
    it demonstrates a one-dimensional class such that every disambiguating class has size which is at least a quasipolynomial in $n$.
    On the positive side, it turns out that \Cref{thm:disambiguation-negative1} 
    is the only obstacle for this version in the sense that relaxing the polynomial bound
    to a quasipolynomial one works:

    % Notice that \Cref{thm:disambiguation-negative1}
    % also suggests that there is no useful analog of the celebrated Sauer-Shelah-Perles Lemma for partial classes: indeed, it demonstrates a partial class whose VC dimension 
    % is $1$ of a quasi-polynomial size, which can not be disambiguated by a class of a polynomial size. On the other hand, the Sauer-Shelah-Perles Lemma asserts that (total) classes with VC dimension $1$ have a polynomial (in fact linear) size.
    % Our next result proves a nearly tight converse to \Cref{thm:disambiguation-negative1}, and might be interpreted as a quasi-polynomial Sauer-Shelah-Perles Lemma for VC classes:
\begin{theorem}[Quasipolynomial Sauer-Shelah-Perles Lemma]
\label[theorem]{cor:disambiguation-positive-intro}
Let $\H$ be a partial concept class on a finite $\X$ with $\vcdim(\H) = d$. Then there exists a disambiguation $\bar{\H}$ of $\H$ of size 
\[|\bar{\H}| = |\X|^{O(d\log(|\X|))}.\]
\end{theorem}

Like the SSP Lemma, also \Cref{cor:disambiguation-positive-intro} yields a dichotomy for partial classes: for every partial class $\H\subseteq\{0,1,\star\}^\X$, either there are arbitrarily large finite $\X'\subseteq\X$ of size $n$ 
such that any disambiguation $\bar\H$ of $\H\vert_{\X'}$ has size 
\[\lvert\bar\H\rvert = 2^{n},\]
or there exists a polynomial $\mathsf{poly}$ such that for every finite $\X'\subseteq\X$ of size $n$ there exists a disambiguation $\bar\H$ of $\H\vert_{\X'}$ whose size  
\[\lvert \bar\H\rvert\leq \mathsf{poly}(n^{\log (n)}).\]
Note that in the latter case, the disambiguation $\bar\H$ \underline{depends} on $\X'$.
Further, \Cref{thm:disambiguation-negative1} implies that such dependence is, in general, necessary; that is, there cannot be a single universal disambiguation $\bar\H$ of $\H$  satisfying $\lvert\bar\H\vert_{\X'}\rvert = o(2^{n})$ for all finite $\X'\subseteq\X$,
where $n=\lvert\X'\rvert$.
Indeed, such an $\bar \H$ would have a finite VC dimension, which would contradict \Cref{thm:disambiguation-negative1}.
Nevertheless, the next result (of which \Cref{cor:disambiguation-positive-intro} is a corollary) shows that it is possible to (strongly) disambiguate any partial VC class $\H$ while maintaining a quasipolynomial bound on the growth function for initial
finite subsets $\X'\subseteq\X$:

\begin{theorem}
\label[theorem]{thm:disambiguation-positive-intro}
Let $\X = \nats = \{1,2,\ldots\}$ 
and let $\H$ be a partial concept class on $\X$ with $\vcdim(\H) = d < \infty$. Then there exists a strong disambiguation $\bar{\H}$ of $\H$, so that for every finite $m$, the projection of $\bar{\H}$ on $[m]$ has size at most $(m+1)^{(d+1)\log_2(m) + 2} = m^{O(d\log(m))}$.
\end{theorem}

%\shmarg{SH: rephrased this comment to be clear that it's a new result here, not a result of theirs that we're stating. (but maybe there's another better way to say it)}
This result is new; however, after expressing it to 
others in personal communications, 
it has recently been applied in the work 
of \citet*{attias:21} in order to prove a bound on the fat-shattering dimension of $k$-fold maxima of real valued function classes.

Let us conclude this discussion about disambiguations 
    and the SSP Lemma with a question:
\begin{question}[Polynomial Growth $\implies$ Disambiguation?]
\label[question]{qn:polynomial}
Let $\H\subseteq\{0,1,\star\}^\X$ and assume there exists a polynomial $\mathsf{poly}$
such that for every finite $\X'\subseteq \X$ there exists a disambiguation $\bar\H$
of $\H\vert_{\X'}$ of size $\lvert\bar\H\rvert\leq \mathsf{poly}(n)$, where $n=\lvert\X'\rvert$.
Does there exist a disambiguation $\bar\H$ of $\H$ 
such that $\vcdim(\bar\H)<\infty$?
\end{question}

We next discuss two techniques for disambiguating which will be useful in our proofs.

\paragraph{Disambiguating by Sample-Compression.}

Sample compression schemes naturally imply disambiguations:
    indeed, consider a partial concept class $\H\subseteq\{0,1,\star\}^\X$ over a finite domain of size $\lvert \X\rvert = n$, and assume we are given a sample compression scheme
    for $\H$ of size $k$. Therefore, for any partial concept $h\in \H$
    there exist $x_1,\ldots, x_k\in \supp(h)$ such that $h$ is extended 
    by the total concept 
    \[\bar\h = \rho\Bigl((x_i,h(x_i))_{i=1}^k,B\Bigr),\]
    where $\rho$ is the reconstruction function of the compression scheme, and $B$
    is a bit-string of side information of length at most $k$.
    In particular, by applying $\rho$ on all such sequences of length 
    at most $k$
    and all such bit-strings $B$, we obtain a disambiguation of $\H$ of size $n^{O(k)}$. 
    This is summarized in the following proposition:
%\shmarg{Ron: As far as I can see, this argument gives a bound which is $2^{2k}$ times the one stated.}    
\begin{proposition}
\label[proposition]{prop:compression-disambiguation}
For any finite $\X$ and any partial concept class $\H$, 
if $\H$ has a compression scheme of size $k$, %then 
there exists a disambiguation $\bar{\H}$ of $\H$ 
of size at most $(c|\X|/k)^{k}$ 
for a numerical constant $c$.
\end{proposition}

\paragraph{Disambiguating by Majority-Votes.}

We conclude this section with highlighting one idea which is used 
    in the proofs of \Cref{cor:disambiguation-positive-intro,thm:disambiguation-positive-intro}. Let $\H\subseteq\{0,1,\star\}^n$ be a partial class. 
    Consider an online learning setting 
    in which an adversary picks a target partial concept $h\in \H$,
    and then in each round $i=1,\ldots,n$, the learner first guesses a label $\hat y_i$. Then, if $i\in \supp(h)$ and $\hat y_i \neq h(i)$ then the learner is given the correct value $h(i)$. (Otherwise, if $i\notin\supp(h)$ or $\hat y_i=h(i)$ then the learner gets no feedback.)
    Notice that a learner which makes at most $k$ mistakes, in the worst case over all $h\in \H$, defines a disambiguation $\bar\H$ of $\H$ whose size $\lvert\bar\H\rvert$ is at most ${n \choose\leq k}$.

Our proofs follow by exhibiting a learner which makes 
    at most $O(\vcdim(\H)\log (n))$ mistakes. 
    This is done by considering a kind of majority-vote using 
    the family of sets which are shattered by $\H$.
    We refer the reader to \Cref{sec:proofs-disambiguation}
    for more details.

% \subsubsection{Closure Properties}\label{sec:closure}
% \shay{
% Perhaps remove from the main results section.
% (i) Partial VC classes are closed under finite unions (ii) 
% There exists a partial class of VC dimension 1 whose 3-wise majority vote has infinite VC dimension
% (iii) if the supports have finite VC dimension then have closure under arbitrary boolean operations
% (iv) Connection with the specialist framework?}
\subsection{Online Learning}
\label[section]{sec:online-learning-results}

We conclude \Cref{sec:results} with a characterization of online learnability.
    The following theorem shows that the Littlestone dimension
    retains its role of characterzing online learnability. 
    See Section~\ref{sec:online-learning} for a precise 
    definition of the online learning setting, in 
    both the realizable (mistake-bound) case 
    and the agnostic (regret-bound) case, along with 
    the formal proof, 
    and more-detailed quantitative results. 
    
\begin{theorem}
\label[theorem]{thm:online-learnability-intro}
The following statements are equivalent for a partial concept class $\H\subseteq\{0,1,\star\}^\X$. 
\begin{itemize}
    \item $\LD(\H) < \infty$.
    \item $\H$ is online learnable in the realizable (mistake-bound) setting.
    \item $\H$ is online learnable in the agnostic (regret-bound) setting.
\end{itemize}
\end{theorem}

Like partial VC classes, also partial classes with finite Littlestone dimension
    (= Littlestone classes) exhibit different behaviour from their total counterparts.
    One such example was discussed in \Cref{sec:DP}.
    However, our understanding of partial Littlestone classes is more limited. 
    In particular, we conclude with the following basic question:

\begin{question}
\label[question]{qn:LD-disambiguation}
Let $\H$ be a partial class with $\LD(\H)<\infty$.
Does there exist a disambiguation of $\H$ by a total class $\bar{\H}$
such that $\LD(\bar{\H})<\infty$? Is there one with $\vcdim(\bar{\H})<\infty$?
\end{question}

We remark that if the answer to Open Question~\ref{qn:polynomial} is affirmative, then so is the answer to the second part (about VC dimension) of Open Question~\ref{qn:LD-disambiguation}. This follows because Littlestone classes can be disambiguated using the SOA algorithm (see Appendix~\ref{sec:online-learning}).

\section{Three Examples and Two Open Questions}
\label[section]{sec:examples}

We next present three examples of partial concept classes which capture the well-studied
    learning tasks corresponding to linear classification with margin guarantees, boosting, and general classifiers with margin.
    We also pose two open problems regarding disambiguating these classes.

\subsection{Geometric Margin}\label[section]{sec:geomargin}
We next demonstrate the expressivity of partial concepts by
    presenting the classical results regarding learnability
    of linear classifiers with margin as the PAC learnability of a partial concept class.
    Since this basic result cannot be expressed as the PAC learnability of a natural (total) concept class, its presentation in introductory classes to machine learning usually deviates from the classical PAC learning theory.
    Thus, this demonstrates a possible didactic value of the theory of partial concept classes.
    
Let $V$ be a (possibly infinite dimensional) real Hilbert space,
    and let $R,\gamma >0$ be the margin parameters. 
\begin{definition}[Separability with Margin]
    A sample $(x_1,y_1),\ldots,(x_n,y_n)\in V\times\{0,1\}$
    is {\it $(R,\gamma)$-separable} if:
    \begin{enumerate}
        \item there exists a ball $B\subseteq V$ of radius $R$ such that 
        $x_1,\ldots, x_n\in B$, and
        \item the distance between the convex hull of $\{x_i : y_i=1\}$
    and the convex hull of $\{x_i : y_i=0\}$ is at least $2\gamma$.
    \end{enumerate}
\end{definition}
    In other words, a sample is $(R,\gamma)$-separable, if the $0$-labelled examples
    and $1$-labelled examples can be separated by a linear classifier with margin $\gamma$
    and all examples lie in a ball of radius $R$.

Let $\H_{R,\gamma}$ denote the class
\begin{align*} \H_{R,\gamma} := \Bigl\{h\in\{0,1,\star\}^V : &\bigl(\forall x_1,\ldots, x_n\in \supp(h)\bigr):\\
&(x_1,h(x_1)),\ldots,(x_n,h(x_n)) \text{ is $(R,\gamma)$-separable} \Bigr\}.
\end{align*}

The following proposition provides tight bounds on the VC dimension and the Littlestone
    dimension of $\H_{R,\gamma}$ (in order to focus on the parameters $R, \gamma$ and not on the dimension of $V$, we assume that the latter is large, specifically $\mathrm{dim}(V) \geq R^2/\gamma^2$). It is based on classical results concerning linear
    classifiers with margin, dating back to \citet*{rosenblatt1958perceptron}.
\begin{proposition}\label{prop:margin-vcdim}
For all $\gamma,R>0$:
$\vcdim(\H_{R,\gamma}) =  \Theta\!\left(\frac{R^2}{\gamma^2}\right)$
and $\Ldim(\H_{R,\gamma}) = \Theta\!\left(\frac{R^2}{\gamma^2}\right)$.
\end{proposition}
\begin{proof}
Since $\vcdim\leq \Ldim$, it suffices to show that
\[\vcdim(\H_{R,\gamma}) = \Omega\!\left(\frac{R^2}{\gamma^2}\right)\quad \text{ and } \quad\Ldim(\H_{R,\gamma}) = O\!\left(\frac{R^2}{\gamma^2}\right).\]
The upper bound on $\Ldim(\H_{R,\gamma})$ follows by the classical mistake-bound analysis of the Perceptron algorithm~\citep*{rosenblatt1958perceptron}, which implies that $\H_{R,\gamma}$ is online learnable in the realizable setting with at most $O\!\left(\frac{R^2}{\gamma^2}\right)$ mistakes, and therefore $\Ldim(\H_{R,\gamma})=O\!\left(\frac{R^2}{\gamma^2}\right)$.

To obtain a lower bound on $\vcdim(\H_{R,\gamma})$,
    let $e_1, e_2,\ldots$ be an orthonormal basis for $V$,
    and consider the set 
    \[C = \left\{Re_i : i \leq \frac{R^2}{\gamma^2} \right\}.\]
    Note that $C$ is shattered: indeed, $C$ is contained in the ball of radius $R$ centered at the origin,
    and for every partition of $\{i : i\leq R^2/\gamma^2\}$ into two sets $A,B$,
    let $w$ denote the vector
    \[w = \frac{\gamma}{R}\Bigl(\sum_{i\in A} e_i - \sum_{i\in B}e_i\Bigl).\]
    Note that $\|w\|^2 = \frac{\gamma^2}{R^2}(\lvert A\rvert + \lvert B\rvert) \leq 1$
    and that $w\cdot Re_i =\gamma$ for $i\in A$ and $w\cdot Re_i =-\gamma$ for $i\in B$.
    Thus, $w$ witnesses that the distance between the convex-hull of $\{Re_i : i\in A\}$ and the convex-hull of $\{Re_i : i\in B\}$ is $\geq 2\gamma$.
\end{proof}

We conclude this example with an open question: Can learnability of linear classifiers 
    under margin assumptions be modeled by the PAC learnability of a total concept class?
\begin{question}
Does there exist a disambiguation of $\H_{R,\gamma}$ by a total class $\bar \H\subseteq\{0,1\}^V$
whose VC/Littlestone dimensions are bounded by a function of $R, \gamma$?
\end{question}
It seems plausible that the answer to this question is no: 
    in particular, our attempts to find ``natural'' 
    (geometrically defined) disambiguations
    resulted with classes whose VC dimension 
    depends on the dimension of the underlying Hilbert space. Note that if the answer here is indeed negative, then so is the answer to Open Questions~\ref{qn:polynomial} and \ref{qn:LD-disambiguation}.

\subsection{Boosting}
\label[section]{sec:boosting}

Boosting is a celebrated machine learning approach which is based on the idea of combining weak and moderately inaccurate hypotheses to a strong and accurate one. 
The following example concerns boosting under the assumption that the weak hypotheses belong to a class of bounded capacity. This setting was explored in detail by~\citet*{Alon20Boosting}, and is inspired by the common understanding that weak hypotheses are ``rules-of-thumbs'' from an ``easy-to-learn class''. (Schapire and Freund '12, Shalev-Shwartz and Ben-David '14.) 
Formally, it is assumed the class of weak hypotheses has a bounded VC dimension. 

One of the main goals addressed by \citet*{Alon20Boosting} is to characterize which target concepts can be learned
    by boosting weak hypotheses from a given base-class $\B$.
    As we will now demonstrate, {\it the setting introduced by \citet*{Alon20Boosting} can be naturally
    expressed by partial concept classes.}
    
The starting point of \citet*{Alon20Boosting} is a reformulation of the weak learnability assumption:
	Recall that the $\gamma$-weak learnability assumption asserts that if $c:\X\to \{0,1\}$ is the target concept
	then, if the weak learner is given enough $c$-labeled examples drawn from any input distribution over $\X$, 
	it will return an hypothesis which is $\gamma$-correlated with $c$. %(See the e.g.\ \citet*{Shalev-Shwartz2014}, Definition 10.1.)
	%Since here it is assumed that the weak learner is a strong learner for the base-class $\B$,
	One can rephrase the weak learnability assumption only in terms of~$\B$ using the following notion:\footnote{In fact, $\gamma$-realizability corresponds to the {\it empirical weak learning assumption} by \citet*{schapire:12}[Chapter 2.3.3]. The latter is a weakening of the standard weak PAC learning assumption which suffices to guarantee generalization.}
\begin{definition}[$\gamma$-realizable samples (\citet*{Alon20Boosting})]\label[definition]{def:gammaRealizable}
Let $\B\!\subseteq\!\{0,\!1\}^\X$ be the base-class and let $\gamma\in(0,1]$. 
    A sample~$S=((x_1,y_1),\ldots,(x_n,y_n))$ is \underline{\it $\gamma$-realizable} with respect to $\mathcal{B}$
    if for any probability distribution $D$ over $S$ there exists $b\in\mathcal{B}$ such that 
\[
 \Pr_{(x,y)\sim D}[b(x) \ne y] \leq \frac{1-\gamma}{2}.
\]

\end{definition}
Note that for $\gamma=1$ the notion of $\gamma$-realizability specializes to the classical notion of realizability (i.e., consistency with the class). 
	Also note that as $\gamma\to 0$, the set of $\gamma$-realizable samples becomes larger.
	
Using this notion one can describe the (partial) class of concepts which can be learned 
    by boosting $\gamma$-accurate hypotheses from $\B$. We denote this class by $\H_\gamma$ and it is defined as follows:
%\shmarg{Ron: This definition ignores the difference between $\{\pm 1\}$ and $\{0,1\}$. Shay: right, I copy pasted and missed it. Thanks}
\begin{align*}
  \H_\gamma = \Bigl\{h\in\{0,1,\star\}^{\X} : &\bigl(\forall x_1,\ldots, x_n\in \supp(h)\bigr):\\
  &(x_1,h(x_1)),\ldots,(x_n,h(x_n)) \text{ is $\gamma$-realizable by $\B$} \Bigr\}.
\end{align*}

Although \citet*{Alon20Boosting} lacked the terminology of partial concept classes,
    they explicitly studied bounds on the VC dimension of $\H_\gamma$
    (which they denoted by $\gamma$-VC dimension).
    They provided the following upper bounds:
\begin{theorem}[\citet*{Alon20Boosting}]
\label[theorem]{thm:generalGammaVC}
Let $\mathcal{B}$ be a class with VC dimension $d$, and let $\gamma>0$.
    Then, the following upper bounds on $\vcdim(\H_\gamma)$ hold: 
    \[\vcdim(\H_\gamma)=O\left(\frac{d}{\gamma^{2}}\log(d/\text{\ensuremath{\gamma}})\right) = \tilde O\Bigl(\frac{d}{\gamma^2}\Bigr),\]
    and 
    \[\vcdim(\H_\gamma)= O_d \left(\frac{1}{\gamma^{\frac{2d}{d+1}}}\right),\]
where $O_d(\cdot)$ conceals a multiplicative constant that depends only on $d$.
\end{theorem}
\citet*{Alon20Boosting} further demonstrated base-classes $\B$ which imply
    tightness in some ranges of the parameters $\gamma,d$. 
    However, in general it remains open to establish tight bounds
    on $\vcdim(\H_\gamma)$ in both $\gamma,d$.
    
\paragraph{A Disambiguation.}
In contrast with the previous example of linear classifiers, 
    here we can prove that there exists a disambiguation with a bounded VC dimension:
\begin{theorem}[A Disambiguation]
\label[theorem]{thm:boosting-disambiguation}
Let $\mathcal{B}$ be a class with VC dimension $d$, and let $\gamma>0$.
Then, there exists a disambiguation of $\H_\gamma$ by a total concept class $\bar \H_\gamma$, such that
\[\vcdim\left(\bar\H_\gamma\right) = \tilde O\left(\frac{d\cdot d^\star}{\gamma^2}\right),\]
where $d^\star\leq 2^{d+1}$ is the dual VC dimension of $\B$.
\end{theorem}
\begin{proof}
Let $k= O(\frac{d^\star}{\gamma^2})$. Define $\bar \H_\gamma$ to be the class of all majority-votes of $k$ hypotheses from $\B$.
By standard bounds on the VC dimension of composed classes
we have $\vcdim(\bar\H_\gamma)=\tilde O\left(\frac{d\cdot d^\star}{\gamma^2}\right)$.

It remains to show that $\bar\H_\gamma$ disambiguates $\H_\gamma$.
This follows by a standard combination of a Minimax argument and uniform convergence
(a.k.a $\epsilon$-approximation):
let $(x_1,y_1),\ldots ,(x_n,y_n)$ be a sample realizable by $\H_\gamma$.
Thus, for every distribution $D$ over $\{(x_i,y_i)\}_{i=1}^n$ there exists
a weak-hypothesis $b\in\B$ such that
\[\Pr_{(x,y)\sim D}[b(x) \ne y] \leq \frac{1-\gamma}{2}.\]
By the Minimax Theorem~\citep*{von-neumann:44}, there exists a distribution $P$ over
$\B$ such that
\[\bigl(\forall (x_i,y_i)\bigr): \Pr_{b\sim P}[b(x_i)\neq  y_i] \leq \frac{1-\gamma}{2}.\]
Thus, by the uniform convergence theorem~\citep*{vapnik:68} applied to the dual class $\B^\star$, it follows that with positive probability, the majority vote of a sequence of independently drawn $b_1,\ldots, b_k \sim P$ (where $k = O(\frac{d^{\star}}{\gamma^2})$) satisfies:
\[\bigl(\forall (x_i,y_i)\bigr):\quad\bigl(\mathsf{Majority}(b_1,\ldots,b_k)\bigr)(x_i)=y_i,\]
as required.
\end{proof}

Note that the dual VC dimension $d^\star$ can be exponential in the VC dimension. Thus, the following question remains:
    \begin{question}
    Let $\mathcal{B}$ be a class with VC dimension $d$, and let $\gamma>0$.
Does there exist a disambiguation of $\H_{\gamma}$ by a total class $\bar{\H}\subseteq\{0,1\}^{\X}$
whose VC dimension is bounded by a polynomial in $d,\gamma^{-1}$?
\end{question}

\subsection{General Separators with Margin} 

As a final example, we present an example of a partial 
concept class that \emph{can} be disambiguated without 
significantly 
increasing the VC dimension.  Specifically, consider 
the case $\X = \{ x \in \reals^d : \|x\| \leq 1 \}$ 
for $d \in \nats$, 
and for $\gamma > 0$ let 
$\G_{d,\gamma}$ be the set of \emph{all} partial functions 
$h : \X \to \{0,1,\star\}$ with 
$\min(\{ \| x_0 - x_1 \| : x_0,x_1 \in \X, h(x_0) = 0, h(x_1)=1 \}\cup\{\infty\}) \geq \gamma$: that is, $\G_{d,\gamma}$ is the set of 
all partial functions having a margin $\gamma$ 
separation between all points labeled $0$ and all 
points labeled~$1$.  This class 
effectively arises in many works (e.g., \citealp*{von-luxburg:04,Gottlieb18compression}), where it 
is typically expressed as a 
margin condition on a data set 
(or, equivalently, a Lipschitz 
constant for the smoothest 
real-valued function that fits the 
data).  For distributions $\PXY$ 
producing data sets 
satisfying this separation, 
there are immediate implications 
for prediction error bounds 
for various simple 
neighborhood-based prediction 
algorithms such as the 
\emph{nearest neighbor} 
algorithm (e.g., \citealp*{cover:67,chaudhuri:14,urner:13b}).

For this class $\G_{d,\gamma}$, we first observe 
that its VC dimension is roughly $\gamma^{-d}$.
Specifically, 
let $M_{d}(\gamma)$ be the $\gamma$-packing number of $\X$: 
that is, the largest number $m$ s.t there exist 
$x_1,\ldots,x_m \in \X$ with 
$\min_{i \neq j} \| x_i - x_j \| \geq \gamma$.
Then we have the following proposition.

\begin{proposition}
\label[proposition]{prop:general-gamma-vcdim}
$\vcdim(\G_{d,\gamma}) = M_{d}(\gamma)$.
In particular, there exist numerical constants 
$c,C \in (0,\infty)$ 
such that 
$\left( \frac{c}{\gamma} \right)^d \leq \vcdim(\G_{d,\gamma}) \leq \left(\frac{C}{\gamma}\right)^d$.
\end{proposition}
\begin{proof}
Let $m = M_{d}(\gamma)$.
To show a lower bound on $\vcdim(\G_{d,\gamma})$,
let $x_1,\ldots,x_m$ be any $\gamma$-packing of $\X$. 
Since any classification of these points has its 
closest $1$ and $0$ -labeled points at distance 
$\geq \gamma$, it follows that $x_1,\ldots,x_m$ 
is shattered by $\G_{d,\gamma}$; 
indeed, $\G_{d,\gamma}$ contains $2^m$ 
functions $h$ with 
$\supp(h) = \{x_1,\ldots,x_m\}$ 
which witness the shattering.
Thus, $\vcdim(\G_{d,\gamma}) \geq m$.  
To show an upper bound, 
for $n \in \nats$ and a sequence $x_1,\ldots,x_n$, 
%is shattered by $\G_{d,\gamma}$, then 
for $(i^*,j^*) = \argmin_{(i,j) : i \neq j} \|x_i - x_j\|$, 
if $\|x_{i^*} - x_{j^*}\| < \gamma$, 
then for any $y_1,\ldots,y_n \in \{0,1\}$ 
such that $(x_1,y_1),\ldots,(x_n,y_n)$ is 
realizable w.r.t.\ $\G_{d,\gamma}$, 
it must be that $y_{i^*} = y_{j^*}$, 
and hence $x_1,\ldots,x_n$ is not shattered 
by $\G_{d,\gamma}$.  Thus, \emph{every} shattered set 
is $\gamma$-separated.  Since $m$ is the maximum size of 
a $\gamma$-separated set, it follows that $\vcdim(\G_{d,\gamma}) \leq m$.
The claimed inequalities in terms of $c,C$ then follow 
from the well-known bounds on $M_{d}(\gamma)$ 
(e.g., \citealp*{szarek:98}).
\end{proof}

Next, we argue that, unlike the 
other two examples above, which seem unlikely 
to have disambiguations of similar 
VC dimension, 
the class $\G_{d,\gamma}$ 
\emph{does} have a (strong) disambiguation 
with VC dimension of comparable size.

\begin{proposition}
\label[proposition]{prop:general-margin-disambiguation}
$\G_{d,\gamma}$ has a strong disambiguation $\bar{\G}_{d,\gamma}$
with $\vcdim(\bar{\G}_{d,\gamma}) = M_{d}(\gamma/2) \leq \left( \frac{2C}{\gamma}\right)^{d}$.
\end{proposition}
\begin{proof}
Fix a maximum-size $(\gamma/2)$-packing 
$S=\{x_1,\ldots,x_m\}$
of $\X$, where $m = M_{d}(\gamma/2)$.
Let $\mathcal{V} = \{V_1,\ldots,V_m\}$ 
be the Voronoi partition induced 
by $S$: 
that is, $V_i = \left\{ x \!\in\! \X : i = \argmin_{i'} \| x - x_{i'} \| \right\}$ 
(breaking ties in the $\argmin$ 
to favor smaller $i'$, so that $\mathcal{V}$ is indeed a partition of $\X$).
Since $S$ is of maximum size, 
any $x \in \X$ has 
$\min_{i} \|x - x_i\| < \gamma/2$.
Thus, each $V_i$ has diameter 
strictly less than 
$\gamma$ by the triangle inequality.
For each $\mathbf{y} = (y_1,\ldots,y_m) \in \{0,1\}^m$, 
let $\bar{h}_{\mathbf{y}}(x) = 
\sum_{i=1}^{m} y_i \ind[ x \in V_i ]$.
Finally, define $\bar{\G}_{d,\gamma} = \{ \bar{h}_{\mathbf{y}} : \mathbf{y} \in \{0,1\}^m \}$.

To see that $\vcdim(\bar{\G}_{d,\gamma}) = M_{d}(\gamma/2)$, 
note that the sequence 
$x_1,\ldots,x_m$ 
is shattered by $\bar{\G}_{d,\gamma}$, 
since each $x_i$ is the unique 
closest point to 
itself (as the other $x_{i'}$ 
points are all $\gamma/2$-far);
thus, $\vcdim(\bar{\G}_{d,\gamma}) \geq m$.
Moreover, since $|\bar{\G}_{d,\gamma}| = 2^m$, 
it necessarily has 
$\vcdim(\bar{\G}_{d,\gamma}) \leq m$.
The inequality, 
upper bounding $M_{d}(\gamma/2)$,
follows from the well-known 
bounds on packing numbers 
in bounded subsets of a 
Euclidean space 
(e.g., \citealp*{szarek:98}).

To complete the proof, we argue that $\bar{\G}_{d,\gamma}$ 
strongly disambiguates $\G_{d,\gamma}$. Let $h \in \G_{d,\gamma}$.
Since each $V_i$ has diameter 
strictly less than $\gamma$, $h$ can assume only one value on $V_i \cap \supp(h)$. Choosing that value as $y_i$ (or an arbitrary value if $V_i \cap \supp(h) = \emptyset$), we get a $\mathbf{y} = (y_1,\ldots,y_m)$ such that $\bar{h}_{\mathbf{y}} \in \bar{\G}_{d,\gamma}$ agrees with $h$ on its support.
\end{proof}

% \subsection{Classifiers on Manifolds}
% \label{sec:manifolds}

%Others?  Classifiers on manifolds, dispersion, graph cut size

\section{Connections to Other Notions in the Literature}
\label[section]{sec:related-work}

\subsection{Data-Dependent Generalization Guarantees}\label[section]{sec:luckiness}
In this section we describe how one can view data-dependent generalization bounds
    as {\it Structural Risk Minimization} over partial concept classes. 

Already in 1974, Vapnik and Chervonenkis \citep*{vapnik:74} showed that standard 
    VC-dimension-based bounds can be significantly improved in the case of linear classifiers that correctly classify the data with a large margin. 
    More generally, data-dependent guarantees provide bounds on the {\it generalization error} of a classifier that can be computed using {\it the same data} that was used to train the classifier~\citep*{Shawe-Taylor98data,Herbrich02luckiness}.
	This makes such bounds particularly appealing in the context of model selection.\footnote{
	Namely, given two competing classifiers, prioritize the one for which the data-dependent bound is better.}	Recently, data-dependent bounds have been used to study generalization in deep neural networks; see e.g.~\citep*{Bartlett17margin,Dziugaite17deep,Neyshabur17exploring,Dziugaite20robust,DanKarolina20PAC}.

Data-dependent analysis is often based on assumptions which \underline{cannot} be modeled in the traditional PAC learning setting:
    namely, it cannot be expressed as the PAC learnability of a given concept class. Consider for example the task of learning a high-dimensional linear classifier with $\gamma$-margin on the unit ball;
	the distribution-free sample complexity of this task is proportional to $1/\gamma^2$, as witnessed e.g.\ by the
	classical Perceptron algorithm~\citep*{rosenblatt1958perceptron}. However, note that the hypotheses outputted by the Perceptron ---
	namely the class of linear classifiers --- has PAC sample complexity (or VC dimension) that scales linearly with the Euclidean dimension, 
	and can therefore be arbitrarily larger than $1/\gamma^2$ and even infinite.
% 	is \underline{not} PAC learnable
% 	as its VC dimension is proportional to the dimension of the space (which can be arbitrarily large or even infinite).
	To the best of our knowledge, the same applies to all learning algorithms in this context.
% 	Similarly, all known learning algorithms in this case use hypothesis classes which are not PAC learnable.
	Thus, it seems that {\it learnability of large-margin linear classifiers cannot be expressed as the PAC learnability of a concept class.} In any case, there is certainly no simple and natural VC class of total concepts which disambiguates large-margin linear classifiers.

Consequently, the general framework for data-dependent analysis
	{deviated} from the traditional PAC setting \citep*{Shawe-Taylor98data,Herbrich02luckiness}.
	Technically, this is done by introducing a data-dependent ``luckiness'' function
	which induces a (data-dependent) hierarchy of hypotheses (luckier hypotheses precede less lucky  ones, as we discuss in more detail below).
% 	This enables performing a (data-dependent) {\it Structural Risk Minimization} using that hierarchy.	
	For example, in the case of large margin linear classifiers,
	the luckiness of each linear separator is its margin with respect to the input sample. 

While the luckiness framework has been successfully applied in various     contexts, {\it it does not yield a crisp notion of
    learnability in the spirit of PAC learning.}
	Moreover, the general results in this context require the luckiness function to satisfy rather arcane technical conditions
	and, while these conditions suffice for proving bounds on a case-by-case basis in various situations, 	
	it is not clear whether they are necessary in general.
	
\paragraph{Data-Dependent Generalization Guarantees via Partial Concept Classes.} 	
An attractive feature of partial concept classes is that they allow to express a variety of 
learning guarantees for specific types of data 
as ``standard'' learning guarantees with respect to a partial concept class: for example, the study of learning guarantees for linear classifiers with margin reduces to the PAC learnability of the partial concept class $\H_{R,\gamma}$ defined in \Cref{sec:geomargin}.
Furthermore, this framework leads to a natural approach for 
proving \emph{data-dependent} learning guarantees: 
i.e., bounds on $\er_{\PXY}(\hat{h}_n)$ 
that do not require assumptions on 
$\PXY$, but rather are expressed in terms of 
properties of the data set.
This can be achieved via a standard application of 
the principle of 
\emph{Structural Risk Minimization} (SRM): 
that is, rather than constructing 
a \emph{data-dependent} hierarchy  
of total concept classes as 
considered by \citep*{Shawe-Taylor98data},
we can establish data-dependent error bounds 
based on a \emph{fixed and data-independent} 
sequence of \emph{partial} concept classes, 
so that we can apply \emph{standard} SRM arguments 
as in \citep*{vapnik:74,vapnik:74b,vapnik:98}.

Specifically, consider any sequence 
$\H_1, \H_2, \ldots$ 
of partial concept classes, 
and for each $i$ let $\alg_i$ be a learning 
algorithm designed for the class $\H_i$.
For any data sequence $S = \{(x_i,y_i)\}_{i=1}^{n}$ 
in $\X \times \{0,1\}$, 
define 
$\hat{\er}_{S}(\H_i) := \min_{h \in \H_i} \frac{1}{|S|} \sum_{i} \ind[ h(x_i) \neq y_i ]$.
First we describe a realizable version of SRM. 
For each $i$, suppose there is a bound 
$B_i(n,\delta)$ such that, for any $\PXY$, 
for $S \sim \PXY^n$, 
with probability at least $1-\delta$, if 
$\hat{\er}_{S}(\H_i) = 0$, then 
$\er_{\PXY}(\alg_i(S)) \leq B_i(n,\delta)$.
Then we can easily produce a method with a corresponding 
data-dependent error bound: 
choose $\hat{i}$ of minimal $B_{\hat{i}}(n,\delta/\hat{i}(\hat{i}+1))$
subject to 
$\hat{\er}_{S}(\H_{\hat{i}}) = 0$ (if it exists), 
and output $\hat{h} = \alg_{\hat{i}}(S)$.
The corresponding guarantee is that, 
with probability at least $1-\delta$, 
if $\hat{i}$ exists, then 
\begin{equation*} 
\er_{\PXY}(\hat{h}) \leq B_{\hat{i}}(n,\delta/\hat{i}(\hat{i}+1)).
\end{equation*}
This holds by a simple union bound, so that the 
$B_i(n,\delta/i(i+1))$ guarantees hold simultaneously 
for all $i$ with probability at least 
$1 - \sum_i \delta/i(i+1) = 1-\delta$.
%\shmarg{SH: I'm planning to extract a lemma that says this and put it in Section~\ref{sec:agnostic}.}
In Section~\ref{sec:agnostic} (Lemma~\ref{lem:agnostic-empirical-bound}), 
we give a general algorithm that can always achieve
the type of guarantee for $\alg_i$ required above, 
specifically with  
\begin{equation*} 
B_i(n,\delta) = O\!\left( \frac{\vcdim(\H_i)}{n}\log^2(n) + \frac{1}{n}\log\!\left(\frac{1}{\delta}\right) \right).
\end{equation*}

For instance, for the margin example in 
Section~\ref{sec:geomargin}, 
since $\vcdim(\H_{R,\gamma}) = \Theta\!\left(\frac{R^2}{\gamma^2}\right)$ 
(from Proposition~\ref{prop:margin-vcdim}),
we can recover the data-dependent margin 
bounds of \citep*{Shawe-Taylor98data} 
by taking the classes $\H_i$ in the hierarchy 
as the partial concept classes $\H_{R_i,\gamma_i}$ 
for an appropriate sequence of $(R_i,\gamma_i)$: 
for instance, it suffices to define 
$R_i = j_i$ and $\gamma_i = 1/k_i$, 
where $(j_i,k_i)$ is an enumeration of $\nats^2$
satisfying $i \leq (j_i+1)^2 (k_i+1)^2$. %%% this is possible, right?
Thus, with probability at least $1-\delta$, 
if the data set $S$ has $x$'s contained in a ball 
of radius $\hat{R}$ and $S$ is linearly separable 
with margin $\hat{\gamma}$, then we may choose 
the class $\H_{\lceil \hat{R} \rceil, 1/\lceil 1/\hat{\gamma} \rceil}$ 
to recover the bound 
$\er_{\PXY}(\hat{h}) = O\!\left( \frac{\hat{R}^2}{\hat{\gamma}^2} \frac{1}{n} \log^2(n) + \frac{1}{n}\log\!\left(\frac{1}{\delta}\right) \right)$
from \citep*{Shawe-Taylor98data}.
%\shmarg{SH: it would be nice to be able to say that we even recover the bound for the same algorithm...}

We can similarly derive a bound that does not 
require $\hat{\er}_{S}(\H_{\hat{i}}) = 0$, recovering 
the full spirit of the SRM principle.
Specifically, suppose that for each $\H_i$, 
the learning algorithm $\alg_i$ guarantees that, 
for any $\PXY$, for $S \sim \PXY^n$, 
with probability at least $1-\delta$, 
$\er_{\PXY}(\alg_i(S)) \leq \hat{\er}_{S}(\H_i) + B_i(n,\delta)$.
Then let us choose $\hat{i}$ to minimize 
$\hat{\er}_{S}(\H_{\hat{i}}) + B_{\hat{i}}(n,\delta/\hat{i}(\hat{i}+1))$,
and output $\hat{h} = \alg_{\hat{i}}(S)$.
As above, by the union bound, 
we have that with probability at least $1-\delta$, 
\begin{equation*}
\er_{\PXY}(\hat{h}) 
\leq \hat{\er}_{S}(\H_{\hat{i}}) + B_{\hat{i}}(n,\delta/\hat{i}(\hat{i}+1)).
\end{equation*}
Again, in Section~\ref{sec:agnostic} (Lemma~\ref{lem:agnostic-empirical-bound}), we propose a 
general algorithm $\alg_i$ that can provide guarantees 
as required above with 
\begin{equation*}
B_i(n,\delta) = O\!\left(\sqrt{\frac{\vcdim(\H_i)}{n}\log^2(n)+\frac{1}{n}\log\!\left(\frac{1}{\delta}\right)}\right).
\end{equation*}

We can also extend this to capture both cases, 
by supposing $\alg_i$ has the guarantee that, for any 
$\PXY$, for $S \sim \PXY^n$, with probability at least 
$1-\delta$, $\er_{\PXY}(\alg_i(S)) \leq \hat{\er}_{S}(\H_i) + B_i(\hat{\er}_{S}(\H_i),n,\delta)$.
Choosing $\hat{i}$ to minimize 
$\hat{\er}_{S}(\H_{\hat{i}}) +  B_{\hat{i}}(\hat{\er}_{S}(\H_{\hat{i}}),n,\delta/\hat{i}(\hat{i}+1))$ and outputting $\hat{h} = \alg_{\hat{i}}(S)$,
we get that with probability at least $1-\delta$, 
$\er_{\PXY}(\hat{h}) \leq \hat{\er}_{S}(\H_{\hat{i}}) +  B_{\hat{i}}(\hat{\er}_{S}(\H_{\hat{i}}),n,\delta/\hat{i}(\hat{i}+1))$.
In particular, in Section~\ref{sec:agnostic} (Lemma~\ref{lem:agnostic-empirical-bound}), 
we propose a general algorithm $\alg_i$ that 
provides a guarantee $B_i$ as required above, 
with %\shmarg{SH: I'm planning to add this.}
\begin{equation*}
B_i(\hat{\eps},n,\delta) = O\!\left( \sqrt{\hat{\eps} \left(\frac{\vcdim(\H_i)}{n}\log^2(n) + \frac{1}{n}\log\!\left(\frac{1}{\delta}\right)\right)} + \frac{\vcdim(\H_i)}{n}\log^2(n)+\frac{1}{n}\log\!\left(\frac{1}{\delta}\right) \right). 
\end{equation*}

%\subsubsection{Comparison to SRM with Data-dependent Hierarchies}
\paragraph{Comparison to SRM with Data-dependent Hierarchies.}
The SRM framework by~\citet*{Shawe-Taylor98data} is based on a
data-dependent regularization function which they call {\it luckiness}: 
let $\H$ be a total concept class, and let $m$ denote any input-sample size.
A luckiness function is a mapping $L\colon \X^m\times \H\to \mathbb{R}^+$ which, given an input sample $S=\{(x_i,y_i)\}_{i\leq m}$,
assigns to each hypothesis $h\in \H$ a real number $L(x_1,\ldots,x_m; h)$ which measures its ``luckiness''.
The intuition is that when choosing between two competing concepts with equal empirical error rate on the data, 
we should prefer the one with a larger value of $L(x_1,\ldots,x_m ;\cdot)$. 
For example, in the context of linear classification with margin 
(on a bounded space), 
the luckiness function $L(x_1,\ldots,x_m; h)$ assigns to each linear classifier its margin with respect to $x_1,\ldots, x_m$.

While a complete formal comparison of the two frameworks is beyond the 
scope of this work,
we note that nearly all of the essential features of the 
data-dependent SRM framework of 
\citep*{Shawe-Taylor98data} can be captured and generalized by 
the present framework of SRM with data-\emph{independent} hierarchies of 
\emph{partial} concept classes.

Let us call a luckiness function $L$ \emph{projective} if,
for any $x_1,\ldots,x_n$ and $h \in \H$, 
every $m \in \nats$ and $i_1,\ldots,i_m \in [n]$ 
satisfy $L(x_{i_1},\ldots,x_{i_m},h) \geq L(x_1,\ldots,x_n,h)$.
All of the examples of luckiness functions given by 
\citep*{Shawe-Taylor98data} are projective (including the margin example), 
and it is not hard to see that one can convert any 
luckiness function into a projective one by defining 
$L'(x_1,\ldots,x_n,h) = \min_{m}\min_{i_1,\ldots,i_m}L(x_{i_1},\ldots,x_{i_m},h)$.
Given any projective luckiness function $L$, 
we can construct a hierarchy of partial concept classes 
$\H_1 \subseteq \H_2 \subseteq \cdots$ as follows.
For $r>0$, we say that a partial concept $h\colon \X\to\{0,1,\star\}$ 
is {\it $r$-lucky} if there exists a total concept $h'\in \H$ such that $h(x)=h'(x)$ for all $x\in\mathsf{supp}(h)$ (i.e., $h'$ extends $h$),
and $L(\vec x; h')\geq r$ for every $\vec x\in (\mathsf{supp}(h))^*$.
Let $\tilde{\H}_r$ denote the class of all $r$-lucky partial concepts.\footnote{For measurability, it may be desirable to restrict to only those $h$ with finite support.  This does not affect the validity of the claims.}
Note that $\tilde{\H}_r : r\in \R^+$ is a hierarchy of partial concept classes 
({i.e., $\tilde{\H}_{r}\supseteq \tilde{\H}_{s}$ for $r \leq s$}).
Moreover, for any given $r$ and data sequence 
$S = \{(x_1,y_1),\ldots,(x_m,y_m)\} \in (\X \times \{0,1\})^m$, 
$S$ is realizable w.r.t.\ the data-dependent total concept class 
$\{ h \in \H : L(x_1,\ldots,x_m;h) \geq r \}$ 
\emph{if and only if} 
$S$ is realizable w.r.t.\ the partial concept class $\tilde{\H}_r$.
Thus, the data-independent hierarchy $\tilde{\H}_r$ captures the essential 
information given by the luckiness function.
Moreover, the rather-complex technical requirements on $L$ imposed 
by \citep*{Shawe-Taylor98data} imply, in particular, a bound on the 
VC dimension of the partial concept classes $\tilde{\H}_r$.\footnote{Technically, 
the assumption in~\citet*{Shawe-Taylor98data} bounds 
the {\it effective} VC dimension of $\tilde{\H}_r$:
i.e., the VC dimension w.r.t.\ typical samples; but also all other results discussed here apply under this assumption. }
Thus, 
we can recover the types of data-dependent error bounds provided by 
\citep*{Shawe-Taylor98data} using the above SRM technique with 
partial concept classes $\H_i = \tilde{\H}_{r_i}$, for a 
suitable discretization $r_1 \geq r_2 \geq \cdots$ 
(e.g., chosen so that $\vcdim(\H_i)=i$).
On the other hand, our framework allows us to use SRM with  
\emph{any} sequence of partial concept classes, including those 
not induced by a luckiness function on a class of total concepts.

\subsection{Multiclass Classification}

One basic question that immediately arises when considering 
partial concepts is how this setting differs from the 
$3$-label \emph{multiclass} classification problem \citep*{ben-david:95}. 
In both cases, there is a class $\H$ of functions 
$\X \to \{0,1,\star\}$.
The only distinction is in the definition of PAC learning, 
where the multiclass setting would allow distributions $\PXY$ 
on $\X \times \{0,1,\star\}$, 
whereas the setting of partial concepts restricts to 
distributions supported on $\X \times \{0,1\}$.
That is, a distribution $\PXY$ on $\X \times \{0,1,\star\}$ 
is realizable w.r.t.\ $\H$ in the $3$-label multiclass setting 
if $\inf_{h \in \H} \PXY(\{ (x,y) : h(x) \neq y\} ) = 0$, 
and otherwise the definition of PAC learnability remains the 
same as in Definition~\ref{defn:pac-learnable}.

As it turns out, 
any partial concept class $\H$ that is PAC learnable 
in the $3$-label multiclass setting is also PAC learnable in the 
partial concepts setting.
%, since the latter merely restricts which distributions are admitted.  
However, there are simple examples 
where the reverse implication fails.  A simple example of this 
is the class $\H$ of all functions $\nats \to \{0,1,\star\}$
whose image is $\{0,\star\}$.  Generally, we can relate learnability 
in these two settings by considering the VC dimension 
of the \emph{supports} of the partial concepts, 
as shown in the following simple result.

\begin{proposition}
\label[proposition]{prop:multiclass}
Let $\H$ be a class of functions $\X \to \{0,1,\star\}$.
The following %statements 
are equivalent.
\begin{enumerate}
\item $\H$ is PAC learnable in the $3$-label multiclass settting.
\item $\H$ is PAC learnable in the partial concepts setting \textbf{and} $\vcdim(\{\supp(h) : h \in \H \}) < \infty$.
\end{enumerate}
\end{proposition}
%\new{Change this to: the following statements are equivalent: (i) $\H$ is PAC learnable in the multiclass setting, (ii) $\H$ is learnable as a partial class (i.e.\ $\vc(\H)<\infty$) and $\vc(\{\supp(h) : h\in \H\})<\infty$.}
\begin{proof}
PAC learnability in the $3$-label multiclass setting is known 
to be completely characterized by a family of 
combinatorial complexity measures \citep*{ben-david:95}.
In particular, the \emph{Graph dimension} is defined as 
$d_{G} := \sup_{h_0} \vcdim( \{ x \mapsto \ind[ h(x) = h_0(x) ] : h \in \H \} )$, where $h_0$ ranges over all functions $\X \to \{0,1,\star\}$.
Another complexity measure, 
known as the \emph{Natarajan dimension} \citep*{natarajan:89},
denoted $d_N$, 
is defined as the largest $d$ such that 
there exist 
$(x_1,y_1^{(0)},y_1^{(1)}),\ldots,(x_d,y_d^{(0)},y_d^{(1)}) 
\in \X \times \{0,1,\star\}^2$ 
with $y_i^{(0)} \neq y_i^{(1)}$ for all $i$, 
and with the property that 
$\forall b_1,\ldots,b_d \in \{0,1\}$, 
$\exists h \in \H$ with 
$\forall i \leq d, h(x_i)=y_i^{(b_i)}$.
In the case of $3$-label multiclass classification,
\citet*{ben-david:95} show that 
$d_{N} \leq d_{G} \leq c d_{N}$ 
for a finite numerical constant $c$.
Moreover, \citet*{natarajan:89,ben-david:95} 
have shown that $\H$ is PAC learnable in the 
$3$-label multiclass setting if and only if $d_{N} < \infty$.

In particular, note that $\vcdim(\H)$ is merely the 
quantity resulting from restricting each 
$y_i^{(0)}=0$ and $y_i^{(1)}=1$ in the definition of 
Natarajan dimension, so that we always have 
$d_{N} \geq \vcdim(\H)$.  Thus, any $\H$ that is 
PAC learnable in the $3$-label multiclass setting 
has $\vcdim(\H) < \infty$, so that our Theorem~\ref{thm:pac-learnability} 
implies $\H$ is also PAC learnable in the partial concepts 
setting.
Moreover, note that for any sequence $x_1,\ldots,x_d$ 
shattered by $\{ \supp(h) : h \in \H \}$, 
if we take $h_0 : \X \to \{0,1,\star\}$ 
as any function equal $\star$ on all of 
$x_1,\ldots,x_d$, then this sequence is also shattered by 
$\{ x \mapsto \ind[ h(x) = h_0(x) ] : h \in \H \}$.
Therefore, $d_{G} \geq \vcdim(\{ \supp(h) : h \in \H \})$,
so that if $\H$ is PAC learnable in the $3$-label multiclass 
setting, then $\vcdim(\{ \supp(h) : h \in \H \}) < \infty$.
%so that if $\vcdim(\{ \supp(h) : h \in \H \}) = \infty$, 
%then $\H$ is not PAC learnable in the $3$-label multiclass setting.

In the other direction, let 
$(x_1,y_1^{(0)},y_1^{(1)}),\ldots,(x_d,y_d^{(0)},y_d^{(1)})$ 
be as in the definition of $d_{N}$. Then $\{x_i:\,\star \notin \{y_i^{(0)}, y_i^{(1)}\}\}$ is shattered by the partial concept class $\H$, while $\{x_i:\,\star \in \{y_i^{(0)}, y_i^{(1)}\}\}$ is shattered by $\{\supp(h):\,h \in \H\}$. Therefore, we have 
$d_{N} \leq \vcdim(\H) + \vcdim(\{\supp(h) : h \in \H\})$.
Thus, since any $\H$ that is PAC learnable in the partial concepts setting must have $\vcdim(\H) < \infty$ 
(by our Theorem~\ref{thm:pac-learnability}), 
we find that if $\vcdim(\{\supp(h) : h \in \H\}) < \infty$
as well, then $d_N < \infty$, 
and hence $\H$ is also PAC learnable in the 
$3$-label multiclass setting.
\end{proof}

The comparison to $3$-label multiclass classification yields some 
further interesting observations.
For instance, one can show that 
Proposition~\ref{prop:multiclass} also implies that, 
when $\vc(\{\supp(h) : h \in \H\}) < \infty$, the ERM principle 
\emph{does} hold for learning in the partial concepts setting.\footnote{This follows from the fact that, with 
$\vcdim(\H) < \infty$ and $\vcdim(\{\supp(h) : h \in \H\})<\infty$, we have a Sauer-Shelah-Perles type bound 
on the total number of $\{0,1,\star\}$ patterns possible 
on any data set, from which uniform convergence guarantees follow for the losses.} 
This contrasts with the discussion above where we found that 
ERM learners can fail spectacularly for some partial concept 
classes with $\vc(\H) < \infty$ (cf Proposition~\ref{prop:partial-ERM-failure}).

Moreover, this connection to multiclass classification has 
a further implication for disambiguation.  Specifically, 
we have the following result.

\begin{proposition}
\label[proposition]{prop:disambiguation-support-vc}
Any partial concept class $\H$ can be strongly disambiguated 
to a total concept class $\bar{\H}$ with 
$\vcdim(\bar{\H}) = O( \vcdim(\H) + \vcdim(\{ \supp(h) : h \in \H \}) )$.
\end{proposition}
\begin{proof}
Define $\bar{\H} = \{ x \mapsto \ind[ h(x) = 1 ] : h \in \H \}$.
Continuing the notation introduced in the proof of Proposition~\ref{prop:multiclass}, we have 
$\vcdim(\bar{\H}) \leq d_{G} = O(d_{N})$ 
(where the last equality is from \citealp*{ben-david:95}, in this case 
of $3$-class multiclass classification). 
Then, as established in the proof of Proposition~\ref{prop:multiclass}, 
we have $d_{N} \leq \vcdim(\H) + \vcdim(\{\supp(h) : h \in \H\})$, 
which completes the proof.
\end{proof}

In particular, this means that if $\H$ is a PAC learnable 
partial concept class, and $\vcdim(\{ \supp(h) : h \in \H \}) < \infty$,
then it can be disambiguated to a learnable total concept class.
This contrasts with the general case discussed in the sections above, 
where we found that 
there exist learnable partial concept classes $\H$ 
that cannot be disambiguated to learnable total concept classes 
(see Theorems~\ref{thm:expressive} and \ref{thm:disambiguation-negative1}).

Another property enjoyed by classes $\H$ that are PAC learnable 
in the $3$-label multiclass setting is that they satisfy 
\emph{closure properties}.  Specifically, for any finite $k$ 
and classes $\H_1,\ldots,\H_k$ that are PAC learnable in the 
$3$-label multiclass setting, 
and any function $U : \{0,1,\star\}^k \to \{0,1,\star\}$, 
the class 
$\{ x \mapsto U(h_1(x),\ldots,h_k(x)) : \forall i, h_i \in \H_i \}$
is also learnable in the $3$-label multiclass setting. 
%%% SH: we might want to cite something for the above.
Together with Proposition~\ref{prop:multiclass}, 
we may conclude that, for any partial concept classes 
$\H_1,\ldots,\H_k$ with $\vcdim(\H_i) < \infty$ 
and $\vcdim(\{\supp(h) : h \in \H_i\})<\infty$, 
and for any function $U : \{0,1,\star\}^k \to \{0,1,\star\}$, 
the partial concept class 
$\{ x \mapsto U(h_1(x),\ldots,h_k(x)) : \forall i, h_i \in \H_i \}$ 
is PAC learnable in the partial concepts setting.
Interestingly, this property is \emph{not} true for general 
partial concept classes, with unrestricted supports.
Specifically, we have the following result.

\begin{proposition}
\label[proposition]{prop:closure-failure}
Define the $2$-argument \emph{majority} function $U$: 
if $1 \in \{y,y'\} \subseteq \{1,\star\}$, let $U(y,y') = 1$; 
if $0 \in \{y,y'\} \subseteq \{0,\star\}$, let $U(y,y') = 0$; 
otherwise let $U(y,y') = \star$.
If $|\X|=\infty$, there exist partial concept classes 
$\H_1,\H_2$ with $\vcdim(\H_1)=\vcdim(\H_2)=0$ 
such that $\vcdim( \{ x \mapsto U(h_1(x),h_2(x)) : h_1 \in \H_1, h_2 \in \H_2 \}) = \infty$.
\end{proposition}
\begin{proof}
Take $\H_1$ as the set of all functions with image contained in $\{0,\star\}$,
and $\H_2$ the set of all functions with image contained in $\{1,\star\}$.
For any $d \in \nats$ and any distinct $x_1,\ldots,x_d \in \X$, 
for any $y_1,\ldots,y_d \in \{0,1\}$, 
take $h_1 \in \H_1$ with $\ind[ h_1(x_i) = 0 ] = \ind[ y_i = 0 ]$ for all $i$, 
and $h_2 \in \H_2$ with $\ind[ h_2(x_i) = 1 ] = \ind[ y_i = 1 ]$ 
for all $i$.  In particular, note that $\supp(h_1) \cap \{x_1,\ldots,x_d\}$ 
and $\supp(h_2) \cap \{x_1,\ldots,x_d\}$ are disjoint, 
and their union is $\{x_1,\ldots,x_d\}$.  Moreover, 
$U(h_1(x_i),h_2(x_i)) = y_i$ for all $i$.
Thus, the sequence $x_1,\ldots,x_d$ is shattered by 
$\{ x \mapsto U(h_1(x),h_2(x)) : h_1 \in \H_1, h_2 \in \H_2 \}$.
Since $d$ can be chosen arbitrarily large, this completes the proof.
\end{proof}

In fact, one can even show such a negative result for 
functions $U$ having a \emph{single} argument: that is, 
$U : \{0,1,\star\} \to \{0,1,\star\}$.  For instance, 
taking $U(y) = \ind[ y = 1 ]$, the class 
$\{ x \mapsto U(h(x)) : h \in \H\}$
represents a strong disambiguation of $\H$, so that 
Theorem~\ref{thm:expressive} indicates that, 
even if $\H$ is learnable, it can happen that the class 
$\{ x \mapsto U(h(x)) : h \in \H\}$ is not learnable.

%\subsection{Specialists}
%\shay{See e.g.\ \citet*{Blum:97,Freund:97c}}
%\subsection{Compatibility functions}
%\citep*{balcan:10}
%\sh{SH: maybe we should just make this a remark in the Luckiness section.}

%\newpage

%\section{Open Problems}
%\label{sec:open-problems}

%\sh{SH: We should either clean this up as a place to re-state all the open questions, or else we should cut it.}

%\begin{itemize}
%\item Question (informal): General principles for learning partial concept classes?

%\item Question: Can Littlestone classes be disambiguated?

%\item Question: Differential privacy for partial concept classes with $\LD(\H) < \infty$?

%\item Question: The optimal sample complexity of PAC learning partial concept classes (regarding log factors)

%\item Question: The optimal sample complexity of agnostically PAC learning partial concept classes (regarding two log factors)

%\item Question: The optimal regret bound for agnostic online learning of partial concept classes (regarding a log factor)

%\item Question: If we let $\H'$ be a maximal set such that all 
%finite realizable sequences are the same as in $\H$, 
%then does every $\PXY$ realizable w.r.t.\ $\H$ 
%have $\inf_{h \in \H'} \er_{\PXY}(h) = 0$?

%\item Question: General real-valued disambiguations 
%such that ERM with hinge loss works?  
%(analogous to the geometric margin case, where 
%ERM with linear functions and hinge loss 
%works).  Could lead to a general framework 
%where we can talk about regularization and 
%such things.
%\end{itemize}

\appendix

\section{Formal Definitions of Complexity Measures}
\label[section]{sec:complexities}

Before getting into the detailed 
results and proofs, we 
first elaborate on the definitions 
of the combinatorial complexity 
measures appearing in our results. 
As mentioned 
in Section~\ref{sec:results}, 
when suitably expressed, the 
complexity measures all inherit 
precisely the same definitions 
as for the traditional setting 
of total concept classes.
Nevertheless, for those readers 
not familiar with the original 
definitions for total concept 
classes, we state the definitions 
here in full detail.

\begin{definition}[Vapnik-Chervonenkis Dimension]
\label[definition]{defn:vc-dim}
For a partial concept class $\H$, 
the \emph{VC dimension} of $\H$, 
denoted $\vcdim(\H)$, 
is the maximum number 
$d \in \nats \cup \{0\}$ 
such that $\exists x_1,\ldots,x_{d}$ 
with $\{ (h(x_1),\ldots,h(x_{d})) : h \in \H \} \supseteq \{0,1\}^{d}$. Such a sequence $\{x_1,\ldots,x_{d}\}$ 
is said to be \emph{shattered} by $\H$.  If there is 
no largest such $d$, then define $\vcdim(\H)=\infty$.
\end{definition}

Next we state the definition of 
the Littlestone dimension.  Recall that a 
sequence of examples $(x_1,y_1),\ldots,(x_n,y_n) \in \X \times \{0,1\}$
is said to be \emph{realizable} w.r.t.\ $\H$ 
if $\exists h \in \H$ with $\forall i \leq n$, $h(x_i)=y_i$.

\begin{definition}[Littlestone dimension]
\label[definition]{defn:littlestone-dimension}
For any partial concept class $\H$, 
the \emph{Littlestone dimension} of $\H$, denoted by $\LD(\H)$, 
is the largest integer $d$ such that 
there exists a set 
\linebreak $\{ x_{\mathbf{y}} : \mathbf{y} \in \bigcup_{0 \leq i \leq d-1} \{0,1\}^i \} \subseteq \X$ 
with the property that, 
for every $y_1,\ldots,y_d \in \{0,1\}$, 
the sequence 
$$(x_{()},y_1),(x_{(y_1)},y_2),(x_{(y_1,y_2)},y_3),\ldots,(x_{(y_1,\ldots,y_{d-1})},y_d)$$
is realizable w.r.t.\ $\H$.
In particular, if no $x$ has more than one realizable label in $\{0,1\}$, then $\LD(\H)=0$.
On the other hand, if no such largest $d$ exists, we define $\LD(\H) = \infty$.
\end{definition}

To interpret the definition, it is conventional to think of the 
$x_{\mathbf{y}}$ points as being organized into a binary \emph{tree} 
based on the prefixes of $\mathbf{y}$, where edges corresponding to a left 
branch are labeled $0$ and edges corresponding to a right branch 
are labeled $1$: that is, the bits of $\mathbf{y}$ determine 
whether to branch left or right at each level along the 
path leading to the node $x_{\mathbf{y}}$.  Then $\LD(\H)$ is the 
maximum \emph{depth} $d$ of a complete tree of this type, 
such that 
all descending paths from the root correspond to a sequence realizable 
w.r.t.\ $\H$ when each non-terminal node $x_{\mathbf{y}}$ on the path 
is given the label of the edge followed next in the path.

Next we restate the definition 
of the Threshold dimension:

\begin{definition}[Threshold dimension]
\label[definition]{defn:threshold}
The \emph{Threshold dimension} of a partial concept class $\H$, 
denoted by $\td(\H)$, 
is the maximum integer $d$ for which there exist 
$x_1,\ldots, x_d\in \X$ and $h_1,\ldots h_d\in\H$ such that 
$h_i(x_j)=\ind[ i \leq j ]$.  If no such largest $d$ exists, 
define $\td(\H) = \infty$.
\end{definition}

We conclude by restating the definition of \emph{compression scheme} 
in formal detail.
Specifically, the following definition is 
originally due to \citet*{littlestone:86}.

%\new{Standardize the definition sample compression schemes in this work, 
%and perhaps present it in an earlier part, possibly in the relevant section of the introduction.}

\begin{definition}[Compression Scheme]
\label[definition]{defn:compression-scheme}
A \emph{compression scheme} is a pair $(\kappa,\rho)$, 
consisting of a compression function 
$\kappa : (\X \times \{0,1\})^* \to (\X \times \{0,1\})^* \times \{0,1\}^*$ 
and a reconstruction function 
$\rho : (\X \times \{0,1\})^* \times \{0,1\}^* \to \{0,1\}^{\X}$, 
satisfying the following property.
For any sequence $S \in (\X \times \{0,1\})^*$, 
$\kappa(S)$ evaluates to some 
$(S',B) \in (\X \times \{0,1\})^* \times \{0,1\}^*$ 
where $S'$ is a sequence of elements of $S$ 
(possibly re-ordered, and possibly including copies, 
having length at most $|S|$).%
\footnote{We also constrain $(\kappa,\rho)$ to be such that, 
for any $n \in \nats$, 
$((x_1,y_1),\ldots,(x_{n-1},y_{n-1}),x_n) \mapsto \rho(\kappa((x_1,y_1),\ldots,(x_{n-1},y_{n-1})))(x_n)$ is a measurable function.} 

The \emph{size} of the compression scheme for a given 
sample size $m$ is $\max_{S \in (\X \times \{0,1\})^m} |\kappa(S)|$ 
(i.e., the length of the sequence $S'$ plus the number of bits in $B$), 
and the (unqualified) \emph{size} of the compression scheme is the 
maximum size over all $m$, or infinite if the size can be unbounded.

For any partial concept class $\H$, 
a sample compression scheme \emph{for} $\H$ is a compression scheme 
$(\kappa,\rho)$ with the additional property that, for every finite 
data sequence $S$ realizable w.r.t.\ $\H$, 
$\rho(\kappa(S))$ is correct on $S$: i.e., 
$\hat{\er}_{S}(\rho(\kappa(S)))=0$.
\end{definition}

The intention in the above definition is that 
$\rho(\kappa(\cdot))$ is interpreted as a learning algorithm.
For brevity, we will sometimes leave off the bit sequence $B$, 
simply specifying $\kappa : (\X \times \{0,1\})^* \to (\X \times \{0,1\})^*$
and $\rho : (\X \times \{0,1\})^* \to \{0,1\}^{\X}$, 
reflecting the special case where no extra bits are ever output 
by the compression function.

\section{Proofs of Disambiguation}
\label[section]{sec:proofs-disambiguation}
\begin{proof}[of \Cref{prop:compact}]
The proof is similar to that of Theorem~\ref{thm:boosting-disambiguation}. Let $k=O(d^{\star}/\gamma^2)$. Consider the class $\bar \H_k$ of all $k$-wise majority votes of concepts from $\bar\H$.
We will show that $\bar\H_k$ disambiguates $\H$ (note that by standard bounds on the variability of the VC dimension under composition/aggregation, we have that $\vc(\bar\H_k)=\tilde O(\vc(\bar\H)\cdot k) = \tilde O(\frac{d\cdot d^*}{\gamma^2})$, as required). 

Let $S$ be a sample realizable by $\H$. We claim that for every distribution $\PXY$ over $S$ there exists $\bar h\in \bar\H$
such that $\er_{\PXY}(\bar h) \leq \frac{1-\gamma/2}{2}$. 
This will suffice, because then an application
of the Minimax Theorem yields a distribution $Q$ over $\bar\H$ such that  
$\Pr_{\bar h\sim Q}[\bar h(x)\neq y]\leq \frac{1-\gamma/2}{2}$ for every $(x,y)\in S$.
Then, an application of the VC theorem ($\eps$-approximation/uniform-convergence) to the dual class $\bar \H^{\star}$ implies that
with positive probability, a random i.i.d sample $\bar h_1,\ldots, \bar h_k \sim Q$, where $k=O(d^{\star}/\gamma^2)$, will satisfy that its majority vote realizes $S$ entirely.

Thus, it remains to show that for every distribution $\PXY$ over $S$ there exists $\bar h\in \bar\H$
such that $\er_{\PXY}(\bar h) \leq \frac{1-\gamma/2}{2}$.
Indeed, consider such a distribution $\PXY$, and draw a random sample $S'$ of size $O(d/\gamma^2)$ from $\PXY$.
By assumption, there must exist $\bar h=\bar h(S')\in \bar\H$ whose empirical loss on the sample satisfies $\hat{\er}_{S'}(\bar h)\leq \frac{1-\gamma}{2}$.
By the VC theorem (again, $\eps$-approximation/uniform-convergence), it follows that with positive probability 
(over the generation of $S'$), we have that the population loss satisfies 
\[\er_{\PXY}(\bar h)\leq \hat{\er}_{S'}(\bar h) + \gamma/4\leq \frac{1-\gamma/2}{2},\]
as required.

\end{proof}

\begin{proof}[of Theorem~\ref{thm:disambiguation-negative1}]
Interestingly, our proof exploits a recent line of breakthroughs in complexity-theory and combinatorics. %~\citet*{}.
    For our purpose, it will be convenient to use the following combinatorial formulation of these results,
    which provides a nearly tight bound to a question posed by Alon, Saks, and Seymour (for background on this question,
    see the survey by \citet*{Bousquet:14}).
    Let $G=(V,E)$ be a simple graph. Recall that the chromatic number of $G$, denoted by $\chi(G)$, 
    is the minimum $k$ for which there exists a coloring $c:V\to[k]$ such that every edge $\{u,v\}\in E$
    satisfies $c(u)\neq c(v)$. The {\it biclique partition number} of $G$, denoted $\mathsf{bp}(G)$
    is the minimum number of complete bipartite graphs needed to partition the edge set of $G$.
    The next result follows from a recent line of works by \citet*{Goos15cis,Ben-David17sensitivity,Balodis:21}:
\begin{theorem}\label[theorem]{thm:ASS}
For every $n$, there exists a simple graph $G=(V,E)$ with $\mathsf{bp}(G)=n$ such that
\[\chi(G) \geq n^{(\log(n))^{1-\eps(n)}},\]
where $\eps(n)$ is a sequence satisfying $\eps(n) \to_{n\to\infty}0$.
\end{theorem}

Let $G=(V,E)$ be a graph as promised by \Cref{thm:ASS}, and let $B_i=(L_i,R_i,E_i)$
    be $n$ complete bipartite graphs which witness that $\mathsf{bp}(G)=n$.
    Define a partial concept class $\H_n\subseteq\{0,1,\star\}^n$ as follows:
    for each vertex $v\in V$ there is a partial concept $c_v\in \H$ such that for every $i\in [n]$:
    \[
    c_v(i) = 
    \begin{cases}
    0   &v\in L_i,\\
    1   &v\in R_i,\\
    \star   &\text{otherwise.}
    \end{cases}
    \]
We finish the proof with the following two lemmas:
\begin{lemma}
\label[lemma]{lem:2-patterns}
$\vcdim(\H_n) = 1$ and $\td(\H_n)\leq 2$. 
{In fact, on every pair of coordinates $\{i,j\}$ the class~$\H$ realizes at most $2$ patterns, that is:
\[(\forall i,j\in [n]):\quad \Bigl\lvert \{0,1\}^2 \cap \{\bigl(h(i),h(j)\bigr) : h\in \H\}\Bigr\rvert \leq 2\]}
\end{lemma}
\begin{proof}
$\vcdim(\H_n) > 0$ because every edge $\{u,v\}\in E$ satisfies $\{u,v\}\in E_i$ for some $i\in [n]$
and thus $\{c_v(i), c_u(i)\} = \{0,1\}$ which implies that $\{i\}$ is shattered by $\H_n$
and hence $\vcdim(\H_n)\geq 1$.

Note that $\vcdim(\H_n) < 2$ and $\td(\H_n)\leq 2$ follow from 
\[(\forall i,j\in [n]):\quad \Bigl\lvert \{0,1\}^2 \cap \{\bigl(h(i),h(j)\bigr) : h\in \H\}\Bigr\rvert \leq 2,\]
and thus it suffices to prove the latter.
Let $\{i,j\}\subseteq [n]$ be a pair of distinct coordinates.
Assume towards contradiction that 
\[\Bigl\lvert \{0,1\}^2 \cap \{\bigl(h(i),h(j)\bigr) : h\in \H\}\Bigr\rvert \geq 3.\]
Thus, either both patterns $00$ and $11$ are realized, or both patterns $01$ and $10$ are realized.
%It suffices to prove that the patterns $00$ and $11$ cannot be both realized on $\{i,j\}$. Indeed, 
We first rule out the former: assume towards contradiction that both $00$ and $11$ are realized.
Thus, there exist two partial concepts $c_u,c_v\in \H$ for $u,v\in V$
such that $c_u(i)=c_u(j)=0$ and $c_v(i)=c_v(j)=1$. Thus, by the definitions of $c_u,c_v$ 
it follows that $u\in L_i\cap L_j$ and $v\in R_i\cap R_j$ and therefore the edge $\{u,v\}$ is covered by both $B_i$ and $B_j$,
which contradicts the assumption that the edges of $B_1,\ldots, B_n$ partition the edges of $G$.
Similarly, the realization of both patterns $01$ and $10$ also implies an edge which is covered twice and hence a contradiction.
\end{proof}

\begin{lemma}
Let $\bar \H\subseteq\{0,1\}^n$ be a disambiguation of $\H_n$.
Then $\bar \H$ defines a coloring of $G$ using $|\bar \H|$ colors.
Therefore,
\[\lvert \bar \H\rvert \geq n^{(\log(n))^{1-\eps(n)}},\]
as required.
\end{lemma}
\begin{proof}
Assign to each vertex $v\in V$ a color $\bar c_v\in \bar \H$ such that $\bar c_v$ disambiguates the partial concept $c_v\in \H_n$.
(I.e., $\bar c_v$ extends $c_v$ to a total concept in $\{0,1\}^n$.)
Indeed, this is a proper coloring since for every edge $\{u,v\}\in E$ there exists a complete bipartite $B_i=(L_i,R_i,E_i)$
such that $u\in L_i$ and $v\in R_i$ or vice versa. 
Thus, $c_u(i)\neq c_v(i)$ and both are in $\{0,1\}$, therefore also $\bar c_u(i)\neq \bar c_v(i)$.
Hence, $u,v$ are assigned different colors, as required.
\end{proof}

Thus, we have shown the first part in \Cref{thm:disambiguation-negative1} by demonstrating
    the classes $\H_n$, for $n\in\mathbb{N}$.
    For the second part, we need to show that over an infinite $\X$ (say $\X=\nats$),
    there exists $\H_{\infty}\subseteq\{0,1,\star\}^\X$ such that $\vcdim(\H_{\infty})=1$, $\td(\H_{\infty}) \leq 2$ 
    and every total concept class $\bar\H$ that disambiguates $\H_{\infty}$
    satisfies $\vcdim(\bar\H)=\infty$.
    One way to construct $\H_{\infty}$ is by taking disjoint copies of the classes $\H_n$
    (i.e., each $\H_n$ has its domain $\X_n$, the domains $\X_n$ are mutually disjoint,
    and $\H_{\infty}$ is defined by taking the union $\cup_n \widetilde{\H}_n$, where $\widetilde{\H}_n$ is obtained from $\H_n$ by adding $\star$'s outside its domain).
    It is easy to see that $\vc(\H_{\infty})=1$ and $\td(\H_{\infty})\leq 2$.
    To see that every disambiguating class $\bar\H$ must have an unbounded VC dimension,
    notice that such a class $\bar \H$ simultaneously disambiguates all of the $\H_n$'s.
    Thus, by the (contra-positive of) the Sauer-Shelah-Perles Lemma~\citep*{sauer:72},
    $\vc(\bar\H)$ must be unbounded.

\end{proof}

\begin{proof}[of \Cref{cor:disambiguation-positive-intro}]
Assume without loss of generality that $\X=[n]=\{1,\ldots, n\}$.  

For any $\H'\subseteq \H$ , we define its shattering strength:
\[ s\bigl(\H'\bigr) = \bigl\lvert\bigl\{S\subseteq[n] ~ : ~ S\text{ is shattered by } \H' \bigr\}\bigr\rvert. \]
Note in particular that  $s\bigl(\H'\bigr) \leq {n \choose \leq \vcdim(\H')}\leq {n \choose \leq d}$. Also, for $(x,y) \in [n] \times \{0,1\}$ we denote \[ \H'| (x,y) = \{h \in \H':\, h(x) = y\}.\]

Define the VC-majority function associated with a subclass $\H'$ by letting $M_{\H'}(x)$ for $x \in [n]$ be the value $y \in \{0,1\}$ which maximizes $s(\H'|(x,y))$, with an arbitrary tie-breaking rule. Observe that
\[ s(\H') \ge s(\H'|(x,0)) + s(\H'|(x,1)). \]
Indeed, every $S$ that is shattered by one of the subclasses $\H'|(x,0)$ or $\H'|(x,1)$
is also shattered by $\H'$ and if $S$ is shattered by both $\H'|(x,0)$ and $\H'|(x,1)$
then both $S$ and $S\cup\{x\}$ are shattered by~$\H'$.

Consider the following strong disambiguation algorithm. Given a partial concept $h \in \H$, write the entries of its disambiguation in the natural order. Start with $\H'$ equal the entire class $\H$ and compute the value of its VC-majority function at $x=1$. Write this value as the entry at $x$ and leave the class intact, except if $x \in \supp(h)$ and $h(x) \ne M_{\H}(x)$. In the latter case, write the opposite value, and update the class by adding $(x, h(x))$ as a constraint: i.e., update $\H'$ to $\H' | (x,h(x))$. Proceed to the next value of $x$ and carry out this procedure with the updated subclass, and so on, until reaching $x=n$.

We claim that given any partial concept $h \in \H$, the number $u(h)$ of updates is at most 
\[\log_2(s(\H))\leq \log_2\Bigl({n \choose \leq d}\Bigr) \leq 1+d\log_2 (n).\] 
Indeed, by the above, after each update the strength of the updated subclass is at most half of the strength of the subclass before the update, and at all times the maintained subclass contains the partial concept $h$.
When running this disambiguation algorithm, the output is determined by the location of the updates. 
By the above bound on $u(h)$, the number of ways to place the updates is at most 
\[ {n \choose \leq 1 + d\log_2 (n)} = n^{O(d \log(n))}.\]
\end{proof}

\begin{proof}[of Theorem~\ref{thm:disambiguation-positive-intro}]
{The proof follows a similar idea like the proof of \Cref{cor:disambiguation-positive-intro}, but we use a carefully tailored weighted VC-majority rather than an unweighted one.}

For a finite sequence $(x_1,y_1),\ldots,(x_k,y_k)$ with $x_i \in \nats$, $y_i \in \{0,1\}$ and $x_1 < \ldots < x_k$, denote by $\H|(x_1,y_1),\ldots,(x_k,y_k)$ the subclass of those partial concepts $h \in \H$ such that $h(x_i) = y_i$ for all $i$. For such a constrained subclass, we define its weight:
\[ w(\H|(x_1,y_1),\ldots,(x_k,y_k)) = \sum_S \frac{1}{n(S)^{d+1}}, \]
where the summation is over all nonempty subsets $S$ of $\nats \setminus \{1,\ldots,x_k\}$ that are shattered by this subclass, and $n(S)$ denotes the largest element of $S$. In the special case when $k=0$, i.e., the class is the entire $\H$, the definition is the same, taking $\{1,\ldots,x_k\} = \emptyset$. Observe that in this case (and hence in every case) the sum is finite:
\[ w(\H) \le \sum_n \frac{n^{d-1}}{n^{d+1}} = \sum_n \frac{1}{n^2} = \frac{\pi^2}{6}, \]
where the inequality holds because for a fixed $n$, the number of terms $\frac{1}{n^{d+1}}$ is at most the number of subsets of $[n]$ of size $d$ or less that include the element $n$, and this number is $\sum_{i=0}^{d-1} {n-1 \choose i} \le n^{d-1}$.

Define the VC-weighted majority function associated with such a constrained subclass by letting $M_{\H|(x_1,y_1),\ldots,(x_k,y_k)}(x)$ for $x \in \nats \setminus \{1,\ldots,x_k\}$ be the value $y \in \{0,1\}$ which maximizes $w(\H|(x_1,y_1),\ldots,(x_k,y_k),(x,y))$, with an arbitrary tie-breaking rule. Observe that
\[ w(\H|(x_1,\!y_1),\ldots,(x_k,\!y_k)) \!\ge\! w(\H|(x_1,\!y_1),\ldots,(x_k,\!y_k),\!(x,\!0)) + w(\H|(x_1,\!y_1),\ldots,(x_k,\!y_k),\!(x,\!1)). \]
Indeed, every term $\frac{1}{n(S)^{d+1}}$ that appears in one of the two sums on the right-hand side, also appears in the sum on the left-hand side (for the same $S$). If a term appears in both sums on the right-hand side, then it appears twice in the sum on the left-hand side: once for $S$ and once for $\{x\} \cup S$.

Consider the following strong disambiguation algorithm. Given a partial concept $h \in \H$, write the entries of its disambiguation in the natural order. Start with the entire class $\H$ and compute the value of its VC-weighted majority function at $x=1$. Write this value as the entry at $x$ and leave the class intact, except if $x \in \supp(h)$ and $h(x) \ne M_{\H}(x)$. In the latter case, write the opposite value, and update the class by adding $(x, h(x))$ as a constraint. Proceed to the next value of $x$ and carry out this procedure with the updated subclass, and so on.

We claim that given any partial concept $h \in \H$, the number $u(m)$ of updates up to $x=m$ is at most $(d+1)\log_2(m) + 2$. Indeed, by the above, after each update the weight of the updated subclass is at most half of the weight of the subclass before the update. Consider the subclass before the last update up to $x=m$ (assuming there is at least one -- otherwise there is nothing to prove). For the last update to occur, that subclass must shatter at least the singleton $S = \{x\}$, so its weight then is at least $\frac{1}{m^{d+1}}$. Hence
\[ \frac{1}{m^{d+1}} 2^{u(m)-1} \le w(\H) \le \frac{\pi^2}{6}, \]
which implies that $u(m) \le \log_2(\frac{\pi^2}{6} m^{d+1}) + 1 \le (d+1)\log_2(m) + 2$, as claimed.

When running this disambiguation algorithm, the first $m$ entries of the output are determined by the location of the  updates up to $x=m$. By the above bound on $u(m)$, the number of ways to place the updates is at most $\sum_{i=0}^{\lfloor (d+1)\log_2(m) + 2 \rfloor} {m \choose i} \le (m+1)^{(d+1)\log_2(m) + 2} = m^{O(d \log(m))}$.
\end{proof}

\section{PAC Learnability: Proofs and Sample Complexity Bounds}
\label[section]{sec:proofs-learnability}

This section presents the formal 
proofs associated with PAC learnability, both in the realizable 
case and agnostic case.  
In particular, these results will 
imply Theorem~\ref{thm:pac-learnability} from Section~\ref{sec:pac-intro}, 
but we will also establish 
quantitative versions that provide 
upper and lower bounds on the 
optimal sample complexity in each 
case.

Here, and in later sections, we use the notation 
$\log(x) := \max\{ \ln(x), 1 \}$.

\subsection{Realizable PAC Learning}
\label[section]{sec:proofs-realizable-learning}
The following lemma describes basic properties of the notion of realizability with respect to partial concepts (as used in \Cref{defn:pac-learnable}):
it demonstrates the relationship with the classical definition for classes $\H\subseteq\{0,1\}^\X$ which contain only total concepts, and implies that the definition we use is more general.
\begin{lemma}[Connection with the classical notion of realizability]\label[lemma]{lem:def-realizable}
Let $\H\subseteq\{0,1,\star\}^\X$ and let $\PXY$ be a distribution over $\X\times\{0,1\}$ such that 
\[\inf_{h\in\H}\er_\PXY(h) = 0.\]
Then, for every $n$, $S\sim \PXY^n$ is realizable by $\H$ with probability $1$. Conversely, if $\PXY$ is a distribution with finite or countable support such that for every $n$, $S \sim P^n$ is realizable by $\H$ with probability $1$, then $\inf_{h\in\H}\er_\PXY(h) = 0$.  
Moreover, if $\H$ contains only total concepts and $\vc(\H) < \infty$ then the converse holds regardless of the support of $\PXY$.
\end{lemma}
Thus, our definition generalizes the classical one in the sense that any distribution $\PXY$
which is realizable in the classical sense ($\inf_{h\in\H}\er_\PXY(h) = 0$) is also realizable according to our definition (every sample drawn from it is realizable with probability $1$). Hence, any algorithm which learns $\H$ in the realizable case according to our definition, also learns it according to the traditional sense. On the other hand, our definition of realizable PAC learning in the partial concept class setting may admit realizable distributions for which $\inf_{h\in\H}\er_{\PXY}(h) \ne 0$. For example, if $\X$ is the interval $[0,1]$ and $\H \subseteq \{0,\star\}^\X$ consists of the functions satisfying that $h^{-1}(0)$ has Lebesgue measure $1/2$, then the uniform $\PXY$ over $\X \times \{0\}$ is realizable according to our definition, yet $\inf_{h\in\H}\er_\PXY(h) = 1/2$. 

\smallskip

\begin{proof}
Notice that for every $h\in\H$:
\[\er_{\PXY}(h) = \E_{S\sim \PXY^n} \hat{\er}_S(h) \geq \E_{S\sim \PXY^n}\bigl[ \min_{h\in \H}\hat{\er}_S(h)\bigr],\]
and therefore also 
\[ \inf_{h\in \H}\er_\PXY(h) \geq \E_{S\sim \PXY^n}\bigl[ \min_{h\in \H}\hat{\er}_S(h)\bigr]\geq 0.\]
Thus, if $\inf_{h\in\H}\er_\PXY(h) = 0$ then $\E_{S\sim \PXY^n}\bigl[ \min_{h\in \H}\hat{\er}_S(h)\bigr]= 0$ and hence $S\sim \PXY^n$
is realizable with probability $1$.

For the converse, assume first that $\PXY$ has finite support of size $n$. Taking $S$ to be the support of $\PXY$, which occurs with positive probability for $S \sim \PXY^n$, shows that the entire support is realizable by $\H$, that is, $\min_{h\in\H}\er_\PXY(h) = 0$. If the support of $\PXY$ is countably infinite, we can enumerate it and repeat the above argument for its first $n$ elements, showing the existence of $h\in\H$ realizing those $n$ elements. Taking $n \to \infty$ implies that $\inf_{h\in\H}\er_\PXY(h) = 0$. Finally, assume that $\H\subseteq\{0,1\}^\X$ contains only total concepts
and satisfies $\vcdim(\H)<\infty$, and $\PXY$ satisfies that 
$S\sim \PXY^n$
is realizable with probability $1$. Then:
\begin{align*}
    0 &=\E_{S\sim \PXY^n} \bigl[\min_{h\in \H}\hat{\er}_S(h)\bigr]\\
    &\geq \inf_{h\in \H}\Bigl[\er_\PXY(h)- O\Bigl(\sqrt{\frac{\vc(\H)}{n}}\Bigr)\Bigr] \tag{by uniform convergence}\\
    &=\inf_{h\in \H}\er_\PXY(h) - o(1).
\end{align*}
Thus, by letting $n\to\infty$ we see that $\inf_{h\in \H}\er_\PXY(h)=0$ as claimed.
\end{proof}

The following theorem establishes upper and lower bounds 
on the optimal sample complexity of PAC learning for any 
given partial concept class $\H$.  In particular, this supplies part of 
the proof of 
Theorem~\ref{thm:pac-learnability} 
(i.e., the part concerning the 
realizable case).\footnote{In all of the results on PAC learning, we implicitly suppose $\H$ satisfies appropriate mild conditions necessary to guarantee measurability of the learning algorithms involved in the proofs of upper bounds.  We refer the interested reader to \citet*{van-der-Vaart:96,van-handel:13} for 
thorough discussions of such 
issues, which we will not discuss further in this article.  We also 
note that such restrictions are only 
required if $\X$ is uncountably infinite.}

\begin{theorem}[PAC Sample Complexity]
\label[theorem]{thm:pac-sample-complexity}
For any partial concept class $\H$ with $\vcdim(\H) < \infty$, 
the optimal sample complexity of PAC learning $\H$, 
$\SC(\eps,\delta)$, satisfies 
the following bounds:
\begin{itemize}
    \item $\SC(\eps,\delta) = O\!\left( \frac{\vcdim(\H)}{\eps}\log\!\left(\frac{1}{\delta}\right)\right)$.
    \item $\SC(\eps,\delta) = O\!\left( \frac{\vcdim(\H)}{\eps}\log^2\!\left(\frac{\vcdim(\H)}{\eps}\right) + \frac{1}{\eps}\log\!\left(\frac{1}{\delta}\right)\right)$.
    \item $\SC(\eps,\delta) = \Omega\!\left( \frac{\vcdim(\H)}{\eps} + \frac{1}{\eps}\log\!\left(\frac{1}{\delta}\right) \right)$.
\end{itemize}
Moreover, if $\vcdim(\H) = \infty$, then $\H$ is not PAC learnable.
\end{theorem}

The following lemma was proven by \citet*{haussler:94} 
for total concept classes.  Here we merely note that the 
result trivially also holds for partial concept classes.

%\shmarg{I guess somewhere we should define measurable classes $\H$ and all that.}

\begin{lemma}[One-inclusion Graph Predictor]
\label[lemma]{lem:one-inclusion-graph}
For any partial concept class $\H$ with $\vcdim(\H)<\infty$, 
there is a function $\alg : (\X \times \{0,1\})^* \times \X \to \{0,1\}$ 
such that, for any $n \in \nats$ and any sequence 
$\{(x_1,y_1),\ldots,(x_n,y_n)\} \in (\X \times \{0,1\})^n$
that is realizable w.r.t.\ $\H$, 
\begin{equation}
\label[equation]{eqn:one-inclusion-loo}
	\frac{1}{n!}\sum_{\sigma\in\mathrm{Sym}(n)}
	\ind\!\left[ \alg(x_{\sigma(1)},y_{\sigma(1)},\ldots,
	x_{\sigma(n-1)},y_{\sigma(n-1)},x_{\sigma(n)}) \neq
	y_{\sigma(n)} \right]
	\le \frac{\vcdim(\H)}{n},
\end{equation}
where $\mathrm{Sym}(n)$ denotes the symmetric group (of permutations of $\{1,\ldots,n\}$).
\end{lemma}
\begin{proof}
As \citet*{haussler:94} proved this result for all 
total concept classes $\H$, we need only note that it 
easily extends to partial concept classes, as follows.
For any $n \in \nats$ and 
$S=\{x_1,\ldots,x_n\}$, let $\X_S$ denote the 
set of distinct elements of the sequence $S$, 
and define $\H_{\X_S}$ as the class of 
all total functions $h : \X_S \to \{0,1\}$ such that 
the sequence $\{ (x,h(x)) : x \in \X_S \}$ 
is realizable w.r.t.\ $\H$. 
In the case that $\H_{\X_S} \neq \emptyset$,
let $\alg_{\X_S}$ be the function 
guaranteed by the lemma for the case of instance 
space equal $\X_S$ and for the total concept class 
$\H_{\X_S}$ defined on this space.  Then for any 
$y_1,\ldots,y_n \in \{0,1\}$ such that 
$\{(x_1,y_1),\ldots,(x_n,y_n)\}$ is realizable 
w.r.t.\ $\H$ (and therefore also realizable w.r.t.\ $\H_{\X_S}$), define 
$\alg(x_1,y_1,\ldots,x_{n-1},y_{n-1},x_n) 
= \alg_{\X_S}(x_1,y_1,\ldots,x_{n-1},y_{n-1},x_n)$.
Since any permutation of the sequence $x_1,\ldots,x_n$ 
leaves the spaces $\X_S$ and $\H_{\X_S}$ unchanged, 
and since it is clear from the definition of VC dimension 
that $\vcdim(\H_{\X_S}) \leq \vcdim(\H)$, it follows 
that \eqref{eqn:one-inclusion-loo} holds for the sequence $\{(x_1,y_1),\ldots,(x_n,y_n)\}$.  

Thus, for any $n \in \nats$ and any 
sequence $\{(x_1,y_1),\ldots,(x_n,y_n)\}$ realizable 
w.r.t.\ $\H$, we have defined the value  $\alg(x_1,y_1,\ldots,x_{n-1},y_{n-1},x_n)$ in a 
way that altogether satisfies \eqref{eqn:one-inclusion-loo}.
To complete the definition, we may (arbitrarily) define 
$\alg(x_1,y_1,\ldots,x_{n-1},y_{n-1},x_n) = 0$
for all sequences $(x_1,\ldots,x_{n}) \in \X^n$ and 
$(y_1,\ldots,y_{n-1}) \in \{0,1\}^{n-1}$ 
such that $\{ h \in \H : \forall i \leq n-1, h(x_i)=y_i, \text{ and } h(x_n) \in \{0,1\} \} = \emptyset$.
\end{proof}

Parts of the proof also rely on a well-known 
generalization bound for compression schemes, 
together with a construction of a particular compression 
scheme based on Boosting.
The following lemma is a classic result 
due to \citet*{littlestone:86}.

\begin{lemma}[Consistent Compression Generalization Bound]
\label[lemma]{lem:consistent-compression-bound}
There exists a finite numerical constant $c \geq 1$ such that, 
for any compression scheme $(\kappa,\rho)$, for any 
$n \in \nats$ and $\delta \in (0,1)$, 
for any distribution $\PXY$ on $\X \times \{0,1\}$, 
for $S \sim \PXY^{n}$, 
with probability at least $1-\delta$, 
if $\hat{\er}_{S}(\rho(\kappa(S)))=0$, then 
\begin{equation*}
\er_{\PXY}(\rho(\kappa(S))) 
\leq \frac{c}{n-|\kappa(S)|} \left( |\kappa(S)|\log(n) + \log\!\left(\frac{1}{\delta}\right) \right).
\end{equation*}
\end{lemma}

The next component is based on a well-known 
Boosting algorithm, known as 
$\alpha$-Boost, which yields a compression scheme 
of a quantifiable size 
that is sample-consistent, 
given access to a ``weak'' learning algorithm; 
see the book of \citet*{schapire:12} for a proof.

\begin{lemma}[Boosting]
\label[lemma]{lem:alpha-boost}
For any $k,n \in \nats$ and 
sequence $(x_1,y_1),\ldots,(x_n,y_n) \in \X \times \{0,1\}$, 
suppose $\alg_w : (\X \times \{0,1\})^k \to \{0,1\}^{\X}$ 
is an algorithm that, 
for any distribution $\PXY$ on $\X \times \{0,1\}$ 
with $\PXY(\{(x_1,y_1),\ldots,(x_n,y_n)\})=1$, 
there exists $S_{\PXY} \in \{(x_1,y_1),\ldots,(x_n,y_n)\}^k$ with  
$\er_{\PXY}(\alg_w(S_{\PXY})) \leq 1/3$. 
Then there is a numerical constant $c \geq 1$ such 
that, for $T = \lceil c \log(n) \rceil$, 
there exist sequences $S_1,\ldots,S_T \!\in\! \{(x_1,y_1),\ldots,(x_n,y_n)\}^k$ such that, for $\hat{h}(\cdot) \!:=\! {\rm Majority}(\alg_w(S_1)(\cdot),\ldots,\alg_w(S_T)(\cdot))$,
it holds that $\hat{h}(x_i) = y_i$ for all $i \in \{1,\ldots,n\}$.
\end{lemma}

We are now ready for the proof of Theorem~\ref{thm:pac-sample-complexity}.

\begin{proof}[of Theorem~\ref{thm:pac-sample-complexity}]
The proof that classes with $\vcdim(\H)=\infty$ are not 
PAC learnable, and indeed also the lower bound 
$\SC(\eps,\delta) = \Omega\!\left( \frac{\vcdim(\H)}{\eps}+\frac{1}{\eps}\log(\frac{1}{\delta})\right)$,
follow by standard arguments from 
\citet*{vapnik:74,blumer:89,ehrenfeucht:89}.
Specifically, for any finite $k \leq \vcdim(\H)$, 
letting $\X_k = \{x_1,\ldots,x_k\}$ be a set shattered by $\H$, 
and letting $\H_{k}$ be the class of all total functions 
$\X_k \to \{0,1\}$, any distribution $\PXY$ on 
$\X_k \times \{0,1\}$ realizable w.r.t.\ $\H_k$ 
can be extended to a distribution on $\X \times \{0,1\}$ 
realizable w.r.t.\ $\H$ with 
$\PXY( (\X \setminus \X_k)\times\{0,1\}) = 0$.
Thus, any lower bound on the sample complexity of 
PAC learning the total concept class $\H_k$ is also 
a lower bound on the sample complexity of learning $\H$.
In particular, \citet*{vapnik:74,blumer:89,ehrenfeucht:89} 
show a lower bound proportional to 
$\frac{k}{\eps}+\frac{1}{\eps}\log(\frac{1}{\delta})$ 
for the sample complexity of PAC learning $\H_k$, 
which is therefore also a lower bound on the sample complexity 
of PAC learning $\H$.  Since this holds for all finite 
$k \leq \vcdim(\H)$, it follows that partial concept 
classes with $\vcdim(\H)=\infty$ 
are not PAC learnable, and partial concept classes $\H$ 
with $\vcdim(\H) < \infty$ have optimal sample complexity 
$\SC(\eps,\delta) = \Omega\!\left(\frac{\vcdim(\H)}{\eps}+\frac{1}{\eps}\log(\frac{1}{\delta})\right)$.

To prove the first upper bound on the optimal sample complexity 
when $\vcdim(\H) < \infty$ 
(which also implies the claim of PAC learnability for 
such classes), we begin by studying the function 
$\alg$ from Lemma~\ref{lem:one-inclusion-graph}, 
following the analogous proof for 
total concept classes given by \citet*{haussler:94}.
In particular, for any distribution $\PXY$ realizable w.r.t.\ $\H$, 
and for 
$(X_1,Y_1),\ldots,(X_{n+1},Y_{n+1})$ i.i.d.\ $\PXY$,
letting $\tilde{h}_n(\cdot) := \alg(X_1,Y_1,\ldots,X_n,Y_n,\cdot)$, 
then by exchangeability of these $n+1$ samples 
and linearity of the expectation, 
we have that 
\begin{align*} 
& \E\!\left[ \er_{\PXY}(\tilde{h}_n) \right] 
 = \E\!\left[ \ind\!\left[ \alg(X_1,Y_1,\ldots,X_n,Y_n,X_{n+1}) \neq Y_{n+1} \right] \right]
\\ & = \frac{1}{(n+1)!}\sum_{\sigma\in\mathrm{Sym}(n+1)}
\E\!\left[ \ind\!\left[ \alg(X_{\sigma(1)},Y_{\sigma(1)},\ldots,X_{\sigma(n)},Y_{\sigma(n)},X_{\sigma(n+1)}) \neq Y_{\sigma(n+1)} \right] \right]
\\ & = 
\E\!\left[ \frac{1}{(n+1)!}\sum_{\sigma\in\mathrm{Sym}(n+1)} \!\!\ind\!\left[ \alg(X_{\sigma(1)},Y_{\sigma(1)},\ldots,X_{\sigma(n)},Y_{\sigma(n)},X_{\sigma(n+1)}) \neq Y_{\sigma(n+1)} \right] \right]
\leq \frac{\vcdim(\H)}{n+1},
\end{align*}
where this last inequality follows from the property \eqref{eqn:one-inclusion-loo} for  
$\alg$ in Lemma~\ref{lem:one-inclusion-graph}, 
which holds with probability one for the sequence 
$(X_1,Y_1),\ldots,(X_{n+1},Y_{n+1})$ 
since $\PXY$ is realizable w.r.t.\ $\H$.

To complete the proof, we again follow an argument 
of \citet*{haussler:94} to convert this algorithm, 
guaranteeing  
$\E[\er_{\PXY}(\tilde{h}_n)] \leq \frac{\vcdim(\H)}{n+1}$, 
into an algorithm with a bound on $\er_{\PXY}(\hat{h})$ 
holding with high probability $1-\delta$.
Specifically, let $m = \left\lfloor \frac{4\vcdim(\H)}{\eps} \right\rfloor \left\lceil \log_{2}\!\left( \frac{2}{\delta} \right) \right\rceil + \left\lceil \frac{32}{\eps} \ln\!\left(\frac{2 \lceil \log_{2}(2/\delta) \rceil}{\delta}\right)   
\right\rceil = O\!\left(\frac{\vcdim(\H)}{\eps}\log(\frac{1}{\delta})\right)$, and let 
$(X_1,Y_1),\ldots,(X_m,Y_m)$ be i.i.d.\ $\PXY$.
Let $n = \left\lfloor \frac{4\vcdim(\H)}{\eps} \right\rfloor$,
and let $S_1$ be the first $n$ samples,  
$S_2$ the next $n$ samples, 
and so on up to $S_k$, 
for $k = \left\lceil \log_{2}\!\left( \frac{2}{\delta} \right) \right\rceil$.
Let $T$ be the remaining $t := m-nk = \left\lceil \frac{32}{\eps} \ln\!\left(\frac{2 \lceil \log_{2}(2/\delta) \rceil}{\delta}\right)   
\right\rceil$ samples.
For each $i \in \{1,\ldots,k\}$, 
let $h_i(\cdot) = \alg(S_i,\cdot)$.
Let $\hat{h}_m = \argmin_{h \in \{h_i : i \leq k\}} \sum_{(x,y) \in T} \ind[h(x) \neq y]$.
Define the learning algorithm for $\H$ as 
returning this $\hat{h}_m$, given 
$(X_1,Y_1),\ldots,(X_m,Y_m)$.

To show that this meets the PAC learning requirement, 
note that each $i \leq k$ has $\E[ \er_{\PXY}(h_i) ] \leq \frac{\eps}{4}$.
Thus, by Markov's inequality, with probability at least 
$\frac{1}{2}$, $\er_{\PXY}(h_i) \leq \frac{\eps}{2}$.
Since these $h_i$ are independent, we have that 
with probability at least $1-2^{-k} \geq 1-\frac{\delta}{2}$, 
at least one $h_{i^*}$ has $\er_{\PXY}(h_{i^*}) \leq \frac{\eps}{2}$.
Also, by a Chernoff bound, for each $i \leq k$, 
on the event $\er_{\PXY}(h_i) \leq \frac{\eps}{2}$,  
\begin{equation*}
\P\!\left( \frac{1}{t} \sum_{(x,y) \in T} \ind[h_i(x) \neq y] > \frac{3}{4}\eps \middle| h_i \right) \leq e^{-t \eps / 24},
\end{equation*}
while on the event $\er_{\PXY}(h_i) > \eps$,
\begin{equation*}
\P\!\left( \frac{1}{t} \sum_{(x,y) \in T} \ind[h_i(x) \neq y] \leq  \frac{3}{4}\eps \middle| h_i \right) \leq e^{-t \eps / 32}.
\end{equation*}
Thus, by the law of total probability (over these two events), 
and a union bound (over all $i \leq k$), 
with probability at least 
$1 - k e^{-t \eps / 32} \geq 1-\frac{\delta}{2}$, 
if any $i$ has $\er_{\PXY}(h_i) \leq \frac{\eps}{2}$, 
then the returned classifier $\hat{h}_m$ has 
$\er_{\PXY}(\hat{h}_m) \leq \eps$.
By a union bound over the above two events, each of 
probability at least $1 - \frac{\delta}{2}$, 
we have that with probability at least $1-\delta$, 
$\er_{\PXY}(\hat{h}_m) \leq \eps$.  This completes the proof of the first upper bound.

To prove the second upper bound, 
again suppose $\vcdim(\H) < \infty$, 
and again let $\alg$ be the function 
from Lemma~\ref{lem:one-inclusion-graph}. 
Let $\PXY$ be realizable w.r.t.\ $\H$, 
let $n \in \nats$, 
and $S = \{(X_i,Y_i)\}_{i \in [n]} \sim \PXY^n$.
Since $\PXY$ is realizable w.r.t.\ $\H$, 
with probability one, 
any distribution $P_0$ supported on 
$\{(X_1,Y_1),\ldots,(X_n,Y_n)\}$ 
is also realizable w.r.t.\ $\H$.
Therefore, 
as established above, for any such distribution $P_0$, 
for $k = 3 \vcdim(\H)$ and $S_{P_0} \sim P_0^k$, 
$\E\!\left[ \er_{P_0}(\alg(S_{P_0},\cdot)) \right] 
\leq 1/3$.  In particular, this implies that, given 
$S$ and $P_0$, there exists a deterministic choice of 
$S_{P_0} \in \{ (X_1,Y_1),\ldots,(X_n,Y_n) \}^k$ 
with $\er_{P_0}(\alg(S_{P_0})) \leq 1/3$.
Thus, $\alg$ satisfies the requirement for $\alg_w$ 
in Lemma~\ref{lem:alpha-boost}, so that 
Lemma~\ref{lem:alpha-boost} implies that 
for a value $T = \lceil c_1 \log(n) \rceil$ (for numerical constant $c_1 \geq 1$), 
there exist $S_1,\ldots,S_T \in \{(X_1,Y_1),\ldots,(X_n,Y_n)\}^k$ 
such that, for $\hat{h}_n(\cdot) := {\rm {Majority}}(\alg(S_1,\cdot),\ldots,\alg(S_T,\cdot))$, 
it holds that $\hat{\er}_{S}(\hat{h}_n) = 0$.
Moreover, note that $\hat{h}_n$ can be expressed as a 
compression scheme, with 
compression function $\kappa$ such that 
$\kappa(S) = (S_1,\ldots,S_T)$ 
and reconstruction function $\rho$ such that 
$\rho(S_1,\ldots,S_T) = \hat{h}_n$. 
Therefore, Lemma~\ref{lem:consistent-compression-bound} 
implies that, with probability at least $1-\delta$, 
\begin{equation*}
\er_{\PXY}(\hat{h}_n) 
\leq \frac{c_2}{n-kT}\left( kT\log(n)+\log\!\left(\frac{1}{\delta}\right) \right)
\end{equation*}
for a numerical constant $c_2 \geq 1$.
For any given $\eps \in (0,1)$, 
the right hand side above 
can be made less than $\eps$ for an 
appropriate choice of 
\begin{equation*}
%n \geq \frac{4c_2}{\eps}\left( 6 c_1 \vcdim(\H) \log^2\!\left( \frac{12 c_1 c_2 \vcdim(\H)}{\eps} \right)+\log\!\left(\frac{1}{\delta}\right) \right),
n = O\!\left( \frac{1}{\eps} \left( \vcdim(\H) \log^2\!\left(\frac{\vcdim(\H)}{\eps}\right) + \log\!\left(\frac{1}{\delta}\right) \right) \right),
\end{equation*}
so that we have that with probability at 
least $1-\delta$, 
$\er_{\PXY}(\hat{h}_n) \leq \eps$. 
This completes the proof.
\end{proof}

We conclude this section by noting the gap between the 
upper and lower bounds in Theorem~\ref{thm:pac-sample-complexity}.
In the case of total concept classes $\H$, 
\citet*{hanneke:16a} showed that the optimal 
sample complexity of PAC learning is exactly 
$\Theta\!\left( \frac{\vcdim(\H)}{\eps} + \frac{1}{\eps}\log\!\left(\frac{1}{\delta}\right)\right)$.
Whether this remains true for the more-general 
setting of partial concept classes is an interesting 
open question:

\begin{question}
Does the optimal sample complexity $\SC(\eps,\delta)$ 
of PAC learning any partial concept class $\H$ 
always satisfy $\SC(\eps,\delta) = \Theta\!\left(\frac{\vcdim(\H)}{\eps} + \frac{1}{\eps}\log\!\left(\frac{1}{\delta}\right) \right)$? 
\end{question}

\subsection{Agnostic PAC Learning}
\label[section]{sec:agnostic}

This section extends the learnability results to the 
agnostic setting, thus, together with the result above, 
fulfilling the complete claim of Theorem~\ref{thm:pac-learnability}.
We first provide a precise definition of agnostic learning, 
as formulating a definition appropriate for 
partial concept classes requires some care.

Recall that we think of partial concept classes as a general way
    for expressing assumptions on the data.
    Thus, we would like to define agnostic learning of a partial concept class $\H$ in a way that reflects the ``distance from realizability'' of a typical sample drawn from the source distribution $\PXY$. This gives rise to the following quantities: for any $n \in \nats$ and data sequence $S \in (\X \times \{0,1\})^n$, define the \emph{empirical error rate} of any partial concept $h$ as $\hat{\er}_{S}(h) = \frac{1}{n}\sum_{i=1}^{n} \ind[h(x_i) \neq y_i]$.
    For a distribution $\PXY$ on $\X \times \{0,1\}$, 
    define the approximation error of $\H$
    with respect to samples of size $n$ as 
    \[\eps^*(n)=\E_{S \sim \PXY^n}\!\left[ \min_{h \in \H} \hat{\er}_{S}(h) \right].\]
    We will later see that $\eps^\star(n)$ is non-decreasing in $n$ and hence $\lim_{n\to\infty}\eps^\star(n)$
    exists and equals $\sup_n \eps^\star(n)$ (see \Cref{lem:def-agnostic}). We define the \emph{approximation error} of $\H$ as
    \begin{equation*} 
    \er_{\PXY}(\H) := \lim_{n\to\infty} \E_{S \sim \PXY^n}\!\left[ \min_{h \in \H} \hat{\er}_{S}(h) \right].
    \end{equation*}
This measures how well a given partial concept class 
    can \emph{fit} data sets that can be sampled from $\PXY$.
    In particular, note that $\er_{\PXY}(\H) = 0$ 
    if and only if $\PXY$ is \emph{realizable} w.r.t.\ $\H$.
    In the agnostic PAC setting, 
    we will be interested in achieving prediction error 
    not-much-worse than $\er_{\PXY}(\H)$, as stated in the following definition.

\begin{definition}[Agnostic PAC Learnability]
\label[definition]{defn:agnostic-pac-learnability}
We say a partial concept class $\H \subseteq \{0,1,\star\}^{\X}$ is \emph{agnostically PAC learnable} 
if $\forall \eps,\delta \in (0,1)$, 
$\exists \SC(\eps,\delta) \in \nats$ and a 
learning algorithm $\alg$ such that, 
for all distributions $\PXY$ on $\X \times \{0,1\}$, 
for $S \sim \PXY^{\SC(\eps,\delta)}$, 
with probability at least $1-\delta$, 
$\er_{\PXY}(\alg(S)) \leq \er_{\PXY}(\H)+\eps$.
The quantity $\SC(\eps,\delta)$ is known as the 
\emph{sample complexity} of $\alg$ for agnostic 
PAC learning, and the 
\emph{optimal} sample complexity of agnostic PAC learning 
for $\H$ is defined as 
the minimum achievable value of $\SC(\eps,\delta)$ 
for each given $\eps,\delta$.
\end{definition}

The following lemma shows that $\er_{\PXY}(\H)$
    is indeed well-defined for every $\H$.
    Also, it shows that for total classes,
    our definition of learnability is not easier to satisfy than the classical one.\footnote{Note that one could naturally adapt the classical definition of agnostic PAC learning to partial concept classes: namely, aiming for 
    $\er_{\PXY}(\alg(S)) \leq \inf_{h \in \H} \er_{\PXY}(h) + \eps$, with probability at least $1-\delta$.  
    However, unlike the criterion in Definition~\ref{defn:agnostic-pac-learnability}, 
    this alternative criterion would \emph{not} 
    recover the implication that, for any given distribution 
    $\PXY$ realizable w.r.t.\ $\H$, agnostic PAC learning 
    under $\PXY$ implies PAC learning under $\PXY$. 
    For instance, for $\X=[0,1]$ 
    and $\H$ all partial functions with image 
    $\{0,\star\}$ and having finite support, 
    the distribution $\PXY$ uniform on $\X \times \{0\}$ 
    is realizable w.r.t.\ $\H$, but $\inf_{h \in \H} \er_{\PXY}(h)=1$, so that even the algorithm 
    that returns $x \mapsto 1$ satisfies the alternative 
    agnostic PAC criterion.  This is another reason to use 
    the definition in terms of $\er_{\PXY}(\H)$ as stated  
    in Definition~\ref{defn:agnostic-pac-learnability}. 
    Note however that all of our results on agnostic learning 
    apply for the alternative definition.} (I.e., any learning algorithm which agnostically learns a total class $\H$ according to \Cref{defn:agnostic-pac-learnability} also agnostically learns $\H$ in the classical PAC sense).
\begin{lemma}\label[lemma]{lem:def-agnostic}
Let $\H\subseteq\{0,1,\star\}^\X$ and let $\PXY$ be a distribution over $\X\times\{0,1\}$. Define \[\eps^\star(n)=\E_{S \sim \PXY^n}\!\left[ \min_{h \in \H} \hat{\er}_{S}(h) \right].\] 
Then:
\begin{enumerate}
    \item For every $n \geq 2$,
$\eps^\star(n)\geq \eps^\star(n-1)$,
and in particular $\lim_{n\to\infty} \eps^\star(n)$ exists.
\item %If $\H\subseteq\{0,1,\star\}^\X$
%contains only total concepts then
Also, 
\[\er_{\PXY}(\H) = \lim_{n \to \infty} \eps^\star(n) \leq \inf_{h\in \H}\er_\PXY(h),\]
and if in addition $\H$ contains only total concepts and $\vc(\H)<\infty$ 
then the above inequality is satisfied with an equality.
\end{enumerate}
\end{lemma}

\begin{proof}

We begin with the first item.
Let $n \ge 2$, and let $S$ be a sequence $(X_1,Y_1),\ldots,(X_n,Y_n)$ drawn i.i.d. from $\PXY$. For $i \in [n]$, denote by $S_{-i}$ the subsequence obtained by removing $(X_i,Y_i)$. For any $h \in \H$ we get from the definition of the empirical error rate and elementary double counting that \[ \hat{\er}_S(h) = \frac{1}{n} \sum_{i=1}^n \hat{\er}_{S_{-i}}(h). \] This implies that \[ \min_{h \in \H} \hat{\er}_S(h) \ge \frac{1}{n} \sum_{i=1}^n \min_{h \in \H} \hat{\er}_{S_{-i}}(h). \] Taking $\E_{S \sim \PXY^n}$ on both sides of this inequality, and noting that for each $i$, $S_{-i} \sim \PXY^{n-1}$, we get \[ \E_{S \sim \PXY^n} \!\left[ \min_{h \in \H} \hat{\er}_S(h) \right] \ge \frac{1}{n} \sum_{i=1}^n \E_{S_{-i} \sim \PXY^{n-1}} \!\left[ \min_{h \in \H} \hat{\er}_{S_{-i}}(h) \right], \] so \[ \eps^\star(n) \ge \frac{1}{n} \sum_{i=1}^n \eps^\star(n-1) = \eps^\star(n-1), \] as required.

Let us now prove the second item.
Notice that for every $h\in\H$ and $n \in \nats$:
\[\er_{\PXY}(h) = \E_{S\sim \PXY^n} \hat{\er}_S(h)\geq \eps^\star(n),\]
and therefore also 
\[\inf_{h\in \H}\er_\PXY(h) \geq \eps^\star(n).\]
Letting $n \to \infty$ yields the stated inequality. In addition if $\H\subseteq\{0,1\}^\X$ contains only total concepts
and satisfies $\vcdim(\H)<\infty$ then, by uniform-convergence, for $S \sim \PXY^n$, with probability at least $1 - \delta$ we have:
\begin{align*}
    \min_{h\in \H}\hat{\er}_S(h) \geq \inf_{h\in \H}\er_\PXY(h) - O\Bigl(\sqrt{\frac{\vc(\H) + \log(\frac{1}{\delta})}{n}}\Bigr).
\end{align*}
Taking $\delta = \frac{1}{n}$ and letting $n \to \infty$ we have the converse inequality
$\lim_{n\to\infty}\eps^\star(n) \geq \inf_{h\in \H}\er_\PXY(h)$,
and hence an equality.
\end{proof}

In particular, in the case of total concept classes $\H$, 
this implies that any learning algorithm that is an 
agnostic PAC learner by our definition is also an 
agnostic PAC learner in the traditional definition, 
and vice versa.

{\vskip 2mm}The portion of Theorem~\ref{thm:pac-learnability} concerning 
agnostic PAC learnability is summarized in the following 
specialized statement.

\begin{theorem}[Agnostic PAC Learnability]
\label[theorem]{thm:agnostic-pac-learnability}
The following statements are equivalent for any partial concept class $\H\subseteq\{0,1,\star\}^\X$. 
\begin{itemize}
    \item $\vcdim(\H) < \infty$.
    \item $\H$ is agnostically PAC learnable.
\end{itemize}
\end{theorem}

Thus, the conditions for agnostic PAC learnability of $\H$ 
are the same as for (realizable) PAC learnability, 
and recover the known conditions for agnostic learnability 
of total concept classes $\H$ (both realizable and agnostic).

{\vskip 2mm}As we did for the realizable case, we will prove a more-detailed 
result on agnostic PAC learning, which also establishes 
upper and lower bounds on the optimal sample complexity.
In particular, Theorem~\ref{thm:agnostic-pac-learnability} 
follows as an immediate implication.

\begin{theorem}[Agnostic PAC Sample Complexity]
\label[theorem]{thm:agnostic-pac-sample-complexity}
For any partial concept class $\H$ with $\vcdim(\H) < \infty$, 
the optimal sample complexity of 
agnostically PAC learning $\H$, $\SC(\eps,\delta)$, 
satisfies the following bounds:
\begin{itemize}
    \item $\SC(\eps,\delta) = O\!\left( \frac{\vcdim(\H)}{\eps^2}\log^2\!\left(\frac{\vcdim(\H)}{\eps}\right) + \frac{1}{\eps^2}\log\!\left(\frac{1}{\delta}\right)\right)$.
%\item I think we can get $\SC(\eps,\delta) = O\!\left( \frac{\vcdim(\H) + \log(\frac{1}{\delta})}{\eps^2} \log^2\!\left(\frac{\vcdim(\H) + \log(\frac{1}{\delta})}{\eps}\right)\right)$.
    \item $\SC(\eps,\delta) = \Omega\!\left( \frac{\vcdim(\H)}{\eps^2} + \frac{1}{\eps^2}\log\!\left(\frac{1}{\delta}\right) \right)$.
\end{itemize}
Moreover, if $\vcdim(\H) = \infty$, then $\H$ is not 
agnostically PAC learnable.
\end{theorem}

While the lower bound will follow from standard approaches, 
to prove the upper bound we will use a technique introduced 
by \citet*{david:16}, which reduces agnostic learning to 
realizable learning.  This technique makes use of two 
main components: generalization bounds for sample compression 
schemes, and a construction of a compression scheme based on Boosting (to which the generalization bounds are then applied).
Both of these components are well known. 
For the Boosting component, we rely on Lemma~\ref{lem:alpha-boost}.  We restate 
the relevant result for compression schemes 
here for completeness.
Specifically, 
the following lemma is a variation on a classic result 
due to \citet*{graepel:05}.\footnote{The original result 
of \citet*{graepel:05} does not include the $\hat{\er}_{S}(\rho(\kappa(S)))$ 
factor inside the square root.  However, this variant follows 
by the same argument, simply substituting the empirical Bernstein 
bound rather than Hoeffding's inequality. 
See \citep*{maurer:09} for a similar result.} 
%%% SH: shall we include a proof?

\begin{lemma}[Agnostic Compression Generalization Bound]
\label[lemma]{lem:compression-bound}
There exists a finite numerical constant $c > 0$ such that, 
for any compression scheme $(\kappa,\rho)$, for any 
$n \in \nats$ and $\delta \in (0,1)$, 
for any distribution $\PXY$ on $\X \times \{0,1\}$, 
for $S \sim \PXY^{n}$, 
letting 
$B(S,\delta) := \frac{1}{n} \left( |\kappa(S)|\log(n) + \log\!\left(\frac{1}{\delta}\right) \right)$, 
with probability at least $1-\delta$,
\begin{equation*}
\left| \er_{\PXY}(\rho(\kappa(S))) - \hat{\er}_{S}(\rho(\kappa(S))) \right| 
\leq c \sqrt{\hat{\er}_{S}(\rho(\kappa(S))) B(S,\delta)} + c B(S,\delta).
\end{equation*}
\end{lemma}
%%% This result is well known in the literature (see e.g., hanneke:20c); for completeness we include a short proof.

We apply Lemma~\ref{lem:compression-bound} 
to a boosting-based compression scheme to obtain the 
following intermediate result, representing the key 
component in the upper bound claimed in Theorem~\ref{thm:agnostic-pac-sample-complexity}.
This also supplies the algorithm supporting the 
claims regarding structural risk minimization 
in Section~\ref{sec:luckiness}.

\begin{lemma}
\label{lem:agnostic-empirical-bound}
For any partial concept class $\H$ with $\vcdim(\H)<\infty$, 
there is a learning algorithm $\alg$ such that, 
for any distribution $\PXY$ on $\X \times \{0,1\}$, 
any $m \in \nats$, and $\delta \in (0,1)$, for $S \sim \PXY^m$, 
letting $\hat{\er}_{S}(\H) := \min_{h \in \H} \hat{\er}_{S}(h)$, 
with probability at least $1-\delta$, 
the output $\hat{h} := \alg(S)$ satisfies 
\begin{equation*}
\er_{\PXY}(\hat{h}) \!\leq\! \hat{\er}_{S}(\H) 
+ c \sqrt{\hat{\er}_{S}(\H) \frac{1}{m}\!\left( \vcdim(\H) \log^2(m) \!+\! \log\!\left(\frac{1}{\delta}\right) \right)}
+ \frac{c}{m}\!\left( \vcdim(\H) \log^2(m) \!+\! \log\!\left(\frac{1}{\delta}\right) \right)
\end{equation*}
for a finite numerical constant $c$.
\end{lemma}
\begin{proof}
We follow a reduction-to-realizable technique 
of \citet*{david:16}.
Specifically, let $\alg_{w}$ be a PAC learning algorithm 
for $\H$ (for the realizable case) achieving 
the optimal sample complexity 
$\SC_{\RE}(\eps',\delta')$ for 
PAC learning $\H$ (in the realizable case), 
where $\eps'=\delta'=1/3$. 
Note that, without loss of generality, we may 
suppose $\alg_w$ outputs total functions 
(e.g., replacing all $\star$ with $0$ cannot 
increase $\er_{\PXY}(h)$ for any $h$).
We will first explain that this may serve 
as a weak learning algorithm $\alg_w$ 
in Lemma~\ref{lem:alpha-boost}.
Specifically, let $k = \SC_{\RE}(1/3,1/3)$.
Given any $n \in \nats$ and 
any sequence 
$(x_1,y_1),\ldots,(x_n,y_n) \in \X \times \{0,1\}$
realizable w.r.t.\ $\H$, 
any distribution $\PXY$ on $\X \times \{0,1\}$ 
with $\PXY(\{(x_1,y_1),\ldots,(x_n,y_n)\})=1$ 
is realizable w.r.t.\ $\H$, 
and therefore guarantees that, 
for $S \sim \PXY^{k}$, with probability at least $2/3$, 
$\er_{\PXY}(\alg_{w}(S)) \leq 1/3$. 
In particular, this implies there exists at least 
one $S_{\PXY} \in \{(x_1,y_1),\ldots,(x_n,y_n)\}^k$ 
with $\er_{\PXY}(\alg_{w}(S)) \leq 1/3$.

Now, let $\PXY$ be any distribution on $\X \times \{0,1\}$, let 
$m \in \nats$, and let 
$S \!=\! \{(X_1,Y_1),\ldots,(X_m,Y_m)\}$ have 
distribution $\PXY^m$.
Let $R$ denote the longest subsequence of 
$(X_1,Y_1),\ldots,(X_m,Y_m)$ that is realizable 
w.r.t.\ $\H$ (breaking ties based on a fixed measurable total ordering of such sequences). 
If $|R|=0$, then (arbitrarily) 
define $\hat{h}$ as the all-$0$ function 
$\hat{h}(x)=0$.
Otherwise, if $|R|>0$, by the above property of $\alg_w$, together with 
Lemma~\ref{lem:alpha-boost}, for $T = \lceil c' \log(|R|) \rceil$ (for a numerical constant $c'$), 
there exist $S_1,\ldots,S_T \in R^k$ such that, 
letting 
$\hat{h}(\cdot) := {\rm Majority}(\alg_w(S_1)(\cdot),\ldots,\alg_w(S_T)(\cdot))$,
we have $\hat{\er}_{R}(\hat{h}) = 0$.

In particular, this implies 
$\hat{\er}_{S}(\hat{h}) \leq \frac{m - |R|}{m} = \hat{\er}_{S}(\H)$.
Moreover, $\hat{h}$ is the output of the compression 
scheme that selects $\kappa(S) = (S_1,\ldots,S_T)$ 
and $\rho(\kappa(S)) = \hat{h}$ 
(or in the case $|R|=0$, $\kappa(S)=\{\}$ and $\rho(\kappa(S))=\hat{h}$).
Therefore, Lemma~\ref{lem:compression-bound} implies 
that, %for $m \geq 2k \lceil c' \log(m) \rceil$, 
with probability at least $1-\delta$, 
\begin{align*}
&\er_{\PXY}(\hat{h}) 
 \leq \hat{\er}_{S}(\hat{h}) 
+ c'' \sqrt{\hat{\er}_{S}(\hat{h})\frac{1}{m}\left( k T \log(m) + \log\!\left(\frac{1}{\delta}\right) \right)}
+ c'' \frac{1}{m}\left( k T \log(m) + \log\!\left(\frac{1}{\delta}\right)\right)
\\ & \leq \hat{\er}_{S}(\H) 
+ c''' \sqrt{\hat{\er}_{S}(\H) \frac{1}{m}\!\left( \vcdim(\H) \log^2(m) \!+\! \log\!\left(\frac{1}{\delta}\right) \right)}
+ c''' \frac{1}{m}\!\left( \vcdim(\H) \log^2(m) \!+\! \log\!\left(\frac{1}{\delta}\right)\right)
\end{align*}
for appropriate numerical constants $c'',c'''$, 
where the last inequality is due to 
Theorem~\ref{thm:pac-sample-complexity}.
%Also note that the inequality trivially holds if  
%$m < 2k \lceil c' \log(m) \rceil$ (as it is greater than $1$).
%This completes the proof.
\end{proof}

We are now ready for the proof of Theorem~\ref{thm:agnostic-pac-sample-complexity}.

\begin{proof}[of Theorem~\ref{thm:agnostic-pac-sample-complexity}]
As was true in the proof of Theorem~\ref{thm:pac-sample-complexity}, 
the proof that classes with $\vcdim(\H)=\infty$ are not 
agnostically PAC learnable, and the lower bound 
$\SC(\eps,\delta) = \Omega\!\left( \frac{\vcdim(\H)}{\eps^2}+\frac{1}{\eps^2}\log(\frac{1}{\delta})\right)$,
follow by standard arguments from 
\citet*{vapnik:74,anthony:99}.
Specifically, for any finite $k \leq \vcdim(\H)$, 
letting $\X_k = \{x_1,\ldots,x_k\}$ be a set shattered by $\H$, 
and letting $\H_{k}$ be the class of total functions 
$h : \X_k \to \{0,1\}$, all distributions $\PXY^{(k)}$ on 
$\X_k \times \{0,1\}$ 
can be extended to a distribution $\PXY$ on $\X \times \{0,1\}$ 
with 
$\PXY( (\X \setminus \X_k)\times\{0,1\}) = 0$.
Moreover, letting $(X_1,Y_1),(X_2,Y_2),\ldots$ be i.i.d.\ $\PXY$, 
we always have 
\begin{align*} 
\er_{\PXY}(\H) & = \sup_n \E_{S \sim \PXY^n}\!\left[ \min_{h \in \H} \hat{\er}_{S}(h) \right] 
\\ & \leq \sup_n \inf_{h \in \H} \E_{S \sim \PXY^n}\!\left[ \hat{\er}_{S}(h) \right] 
= \inf_{h \in \H} \er_{\PXY}(h) 
= \min_{h \in \H_k} \er_{\PXY^{(k)}}(h).
\end{align*}
Additionally, note that
\begin{align*}
\er_{\PXY}(\H) & %\geq \lim_{n\to\infty} \E_{S \sim \PXY^n}\!\left[ \min_{h \in \H} \hat{\er}_{S_n}(h) \right]
%\\ & 
\geq \lim_{n \to \infty} \E\!\left[ \sum_{i=1}^{k} \min_{y \in \{0,1\}} \frac{1}{n} \sum_{t=1}^{n} \ind[ X_t = x_i ] \ind[ Y_t \neq y ]  \right]
\\ & = \sum_{i=1}^{k} \E\!\left[  \min_{y \in \{0,1\}} \lim_{n \to \infty} \frac{1}{n} \sum_{t=1}^{n} \ind[ X_t = x_i ] \ind[ Y_t \neq y ]  \right]
\\ & = \sum_{i=1}^{k} \min_{y \in \{0,1\}} \PXY( \{(x_i,1-y)\} ) 
= \min_{h \in \H_k} \er_{\PXY^{(k)}}(h),
\end{align*}
where we have used the Dominated Convergence Theorem, 
continuity of the $\min$, and the Strong Law of Large Numbers.
Thus, 
$\er_{\PXY}(\H) = \min_{h \in \H_k} \er_{\PXY^{(k)}}(h)$.
This means that $\SC(\eps,\delta)$ can be no smaller than 
the sample complexity $M_k(\eps,\delta)$ of agnostically learning 
the total concept class $\H_k$ on $\X_k$, in the traditional 
(total concepts) sense: that is, there exists an algorithm that, for all distributions 
$\PXY^{(k)}$ on $\X_k \times \{0,1\}$, from $M_k(\eps,\delta)$ 
i.i.d.\ samples from $\PXY^{(k)}$, outputs an $\hat{h}$ with 
$\er_{\PXY^{(k)}}(\hat{h}) - \min_{h \in \H_k} \er_{\PXY^{(k)}}(h) \leq \eps$, with probability at least $1-\delta$.
Since $\X_k$ is shattered, the standard lower bounds 
based on VC dimension imply a lower bound 
$M_k(\eps,\delta) = \Omega\!\left( \frac{k}{\eps^2} + \frac{1}{\eps^2} \log\!\left(\frac{1}{\delta}\right) \right)$ 
\citep*{vapnik:74,anthony:99,kontorovich:19}.
In particular, 
the sample complexity 
lower bound $\SC(\eps,\delta) = \Omega\!\left( \frac{\vcdim(\H)}{\eps^2} + \frac{1}{\eps^2} \log\!\left( \frac{1}{\delta}\right) \right)$ for the case of $\vcdim(\H) < \infty$ follows immediately, 
as does the fact that having $\vcdim(\H)=\infty$ 
implies that $\H$ is not agnostically PAC learnable.

To prove the upper bound on the optimal sample complexity 
when $\vcdim(\H) < \infty$ 
(which also implies the claim of agnostic 
PAC learnability for such classes), 
consider the algorithm $\alg$ from 
Lemma~\ref{lem:agnostic-empirical-bound}.
From that lemma, we have that for any distribution $\PXY$, 
sample size $m \in \nats$, and $\delta \in (0,1)$, 
for $S = \{(X_1,Y_1),\ldots,(X_m,Y_m)\} \sim \PXY^m$, 
with probability at least $1-\delta/2$, 
the returned classifier $\hat{h} = \alg(S)$ satisfies 
\begin{align*}
\er_{\PXY}(\hat{h}) 
& \!\leq\! \hat{\er}_{S}(\H) 
\!+\! c \sqrt{\hat{\er}_{S}(\H) \frac{1}{m}\!\left( \vcdim(\H) \log^2(m) \!+\! \log\!\left(\frac{2}{\delta}\right) \right)}
\!+\! \frac{c}{m}\!\left( \vcdim(\H) \log^2(m) \!+\! \log\!\left(\frac{2}{\delta}\right) \right)
\\ & \leq \hat{\er}_{S}(\H) + c' \sqrt{\frac{1}{m}\left( \vcdim(\H) \log^2(m) + \log\!\left(\frac{1}{\delta}\right) \right)}
\end{align*}
for appropriate numerical constants $c,c'$.

\bigskip

Taking 
%$m = \left\lceil c'' \frac{\vcdim(\H) + \log(\frac{1}{\delta})}{\eps^2}\log^2\!\left(\frac{\vcdim(\H) + \log(\frac{1}{\delta})}{\eps}\right) \right\rceil$ 
$m = \left\lceil c'' \frac{1}{\eps^2} \left( \vcdim(\H)\log^2\!\left(\frac{\vcdim(\H)}{\eps}\right) + \log\!\left(\frac{1}{\delta}\right) \right) \right\rceil$ 
for a suitable numerical constant $c''$ we get from the above that with probability at least $1 - \delta/2$,
\begin{align*}
\er_{\PXY}(\hat{h}) \leq \hat{\er}_{S}(\H) + \frac{\eps}{2}.    
\end{align*}
It remains to replace the empirical value $\hat{\er}_{S}(\H) := \min_{h \in \H} \hat {\er}_{S}(h)$ with our benchmark $\er_{\PXY}(\H)$. 
We do this by observing that the random variable 
$\min_{h \in \H} \hat {\er}_{S}(h)$ is concentrated around its mean. Indeed, this random variable is a function of the i.i.d. sequence $(X_1,Y_1),\ldots,(X_m,Y_m)$, and changing the value of any $(X_i,Y_i)$ can change $\min_{h \in \H} \hat {\er}_{S}(h)$ by at most $\frac{1}{m}$. By McDiarmid's inequality
\begin{align*}
\P\!\left(\min_{h \in \H} \hat{\er}_{S}(h) > \E_{S' \sim P^m}\left[\min_{h \in \H} \hat {\er}_{S'}(h) \right] + \frac{\eps}{2} \right) \leq e^{-\frac{\eps^2 m}{2}},
\end{align*}
and for our value of $m$ this probability is less than $\delta/2$. By the union bound, with probability at least $1 - \delta$,
\begin{align*}
\er_{\PXY}(\hat{h}) \leq \min_{h \in \H} \hat{\er}_{S}(h) + \frac{\eps}{2} \leq \E_{S' \sim P^m} \left[ \min_{h \in \H} \hat {\er}_{S'}(h) \right] + \eps \leq \er_{\PXY}(\H) + \eps.    
\end{align*}
So taking the algorithm $\alg$ from Lemma~\ref{lem:agnostic-empirical-bound} 
%$\alg(S) = \hat{h}$ 
meets the requirement.
\end{proof}

We note that, in the special case of total concept classes, the optimal sample complexity 
of agnostic PAC learning is known to be exactly 
$\SC(\eps,\delta) = \Theta\!\left( \frac{\vcdim(\H)}{\eps^2} + \frac{1}{\eps^2}\log\!\left(\frac{1}{\delta}\right)\right)$ 
(\citealp*{talagrand:94}; see \citealp*{van-der-Vaart:96}, Theorems 2.14.1 and 2.6.7).
However, the proof of this refined upper bound relies on 
a technique known as \emph{chaining}, and moreover relies on uniform 
convergence, which can fail for partial concept classes.
Thus, the following question remains open:

\begin{question}
Does the optimal sample complexity, $\SC(\eps,\delta)$, of agnostically PAC learning any partial concept class $\H$ 
satisfy $\SC(\eps,\delta) = \Theta\!\left( \frac{\vcdim(\H)}{\eps^2} + \frac{1}{\eps^2}\log\!\left(\frac{1}{\delta}\right) \right)$ ?
\end{question}

%Question: can we get the log factor in the lower bound?  (even just for one example)

\section{Proofs of Results on Traditional Learning Principles}
\label[section]{sec:erm-failure}

%\begin{theorem}[Failure of Empirical Risk Minimization]
%\label{thm:total-ERM-failure-intro}
%There exists a partial concept class $\H$ with $\vcdim(\H)=1$ 
%such that, for any total concept class $\bar{\H}$, there exists an ERM algorithm for $\bar{\H}$ that 
%is not a PAC learning algorithm for $\H$. 
%Further, this applies even if we allow the learner to pick $\bar{\H}$ based on the size of the input sample.
%\end{theorem}

\begin{proof}[of Theorem~\ref{thm:total-ERM-failure-intro}]
Let $\X = \nats$ and 
let $\H_{\infty}$ be the partial concept 
class from the proof of 
Theorem~\ref{thm:disambiguation-negative1}, having $\vcdim(\H_{\infty})=1$ 
while any disambiguation $\bar{\H}_{\infty}$ of $\H_{\infty}$ 
must have 
$\vcdim(\bar{\H}_{\infty})=\infty$.
Also let $h_0$ be a concept with 
$h_0(x) = 0$ everywhere, 
and define $\H = \H_{\infty} \cup \{h_0\}$.
In particular, it follows from Lemma~\ref{lem:2-patterns} 
%by the same reasoning as in the 
%proof of Theorem~\ref{thm:disambiguation-negative1}, 
that any pair of points $(x_1,x_2) \in \X^2$ have at most 
two patterns of labels in $\{0,1\}^2$ realizable w.r.t.\ $\H_{\infty}$.
%for which $(x_1,0),(x_2,1)$ and $(x_1,1),(x_2,0)$ 
%are both realizable w.r.t.\ $\H_{\infty}$.
Therefore, adding $h_0$ to the class does not increase 
its VC dimension, since it can at most increase the 
number of patterns to three, but not to four.
Hence, $\vcdim(\H) = 1$.
%since one of these $01$ or $10$ patterns 
%will still be missing from any two given points: 
%that is, $\vcdim(\H) = 1$.

Now consider any total concept class $\bar{\H}$.  
First, if $\bar{\H}$ is not a 
disambiguation of $\H$, 
then there exists a finite 
sequence $S$ of points $(x,y) \in \X \times \{0,1\}$ that is 
realizable w.r.t.\ $\H$ 
but not realizable w.r.t.\ $\bar{\H}$.
Without loss of generality, 
suppose all elements of $S$ are 
distinct.
Then letting $\PXY$ be the uniform distribution on $S$, 
$\PXY$ is realizable w.r.t.\ $\H$.
However, for any learning algorithm 
producing hypotheses $\hat{h}$ in 
$\bar{\H}$, regardless of how many 
i.i.d.\ samples it is provided 
with, it will always have 
$\er_{\PXY}(\hat{h}) \geq 1/|S|$, 
so that its sample complexity 
for any $\eps < 1/|S|$ is 
infinite: that is, it does not 
satisfy the PAC learning requirement.

On the other hand, consider the case 
where $\bar{\H}$ is a disambiguation 
of $\H$. Then $\bar{\H}$ is also a disambiguation of $\H_{\infty}$,
and therefore, by the property of 
$\H_{\infty}$ from Theorem~\ref{thm:disambiguation-negative1}, it must be that $\vcdim(\bar{\H})=\infty$.
Now let $U_1,U_2,\ldots$ be a sequence of disjoint subsets of $\X$ 
with $|U_i|=i$, such that each 
$U_i$ is shattered by $\bar{\H}$; 
for instance, such a sequence can 
be constructed by first considering 
shattered sets of sizes $4^i$ 
for each $i$, each of which has 
at least $(1/2) 4^i$ elements not 
appearing in any of the smaller sets.
Then consider an ERM algorithm $\alg$ 
for $\bar{\H}$ that, given any data 
sequence $S$ whose elements are all 
contained in $U_i \times \{0\}$ 
for one of the $i \in \nats$, 
$\alg(S)$ returns a function $h$ 
with $h(x) = 1$ for every $x \in U_i$ 
that does not appear in $S$.
$\alg(S)$ may be defined as any ERM 
in the case $S$ is not contained in 
any of the $U_i \times \{0\}$ sets.

Now, given any sample size $m$, 
take $\PXY$ uniform on 
$U_{c m} \times \{0\}$ for an 
integer $c \geq 2$.
Note that this $\PXY$ is realizable 
w.r.t.\ $\H$ since $h_0 \in \H$.
However, for $S \sim \PXY^m$, 
we will have $S$ contained in 
$U_{c m} \times \{0\}$, but at least 
$(c-1)m$ of the $c m$ elements of this set 
will not be present in $S$, 
and therefore $\alg(S)(x)$ will be $1$ 
for at least $(c-1)m$ elements of $U_{c m}$.
Thus, $\er_{\PXY}(\alg(S)) \geq 1-\frac{1}{c}$. 
By choosing $c$ large, this 
can be made arbitrarily close to $1$.

In particular, this implies 
there is no finite sample 
size $m$ at which 
$\P_{S \sim \PXY^m}( \er_{\PXY}(\alg(S)) < 1-\frac{1}{c} ) > 0$ 
holds for all $\PXY$ realizable 
w.r.t.\ $\H$, 
so that $\alg$ is not a PAC learning 
algorithm for $\H$.
\end{proof}

\begin{proof}[of Theorem~\ref{thm:vcdim-image}]
Let $\H_{\infty}$ be the class from 
Theorem~\ref{thm:disambiguation-negative1}.
Consider a finite realizable 
data sequence $S$, and assume without loss of generality that its elements are all distinct. Letting $\PXY$ be the uniform distribution on $S$, and aiming at a prediction error at most $\eps < 1/|S|$, the algorithm running on a large enough sample must output with positive probability a function that realizes $S$.
Since we can choose $S$ as any finite data sequence realizable w.r.t.\ $\H$, the image of the learning algorithm disambiguates $\H_{\infty}$, and 
hence by Theorem~\ref{thm:disambiguation-negative1} 
it must have infinite VC dimension.
\end{proof}

%\begin{theorem}[Sample Compression for Partial Concept Classes]\label{thm:compression}\ \\
%\vspace{-5mm}
%\begin{enumerate}
%    \item Let $\H$ be a partial concept class. Then, there exists a sample compression scheme for~$\H$
%of size $\tilde O(\vcdim(\H)\log (m))$, where $m$ is the size of the input sample.
%    \item There exists a partial concept class $\H$ of $\vcdim(\H)=1$ such that any sample compression scheme for $\H$
%must have size $\Omega(\log(m))$, where $m$ is the size of the input sample.
%In particular, the bounded-size sample compression conjecture is \textbf{false} for partial concept classes.
%\end{enumerate}
%\end{theorem}

\begin{proof}[of Theorem~\ref{thm:compression}]
For the first claim, 
note that Theorem~\ref{thm:pac-sample-complexity} implies that, 
for any realizable data set, 
and any distribution supported on 
those points, there exists a 
sequence of  
$O(\vcdim(\H))$ data points from the $m$ points 
that can be fed into the algorithm 
from Theorem~\ref{thm:pac-sample-complexity} 
to get a hypothesis $\hat{h}$ 
with prediction error at most 
$1/3$ under that distribution.
In conjunction with a standard 
boosting algorithm 
(e.g., $\alpha$-boost; see 
Lemma~\ref{lem:alpha-boost} of Section~\ref{sec:agnostic}), 
we arrive at a sequence 
$\hat{h}_1,\ldots,\hat{h}_{T}$, 
for $T = O(\log(m))$, 
where each $\hat{h}_i$ is based on 
applying the algorithm from 
Theorem~\ref{thm:pac-sample-complexity} 
to some subset of $O(\vcdim(\H))$ 
data points, and where 
${\rm Majority}(\hat{h}_1,\ldots,\hat{h}_T)$ is correct on the entire 
set of $m$ data points. 
Thus, we can specify this classifier 
using $k = O(\vcdim(\H) \log(m))$ 
data points.
Together with $O(k \log(k))$ bits to 
encode an order of the points 
so as to recover precisely which 
points correspond to which $\hat{h}_i$, 
this yields the claimed compression scheme.

For the second claim, consider the partial concept class $\H_{\infty}$ constructed in the proof of Theorem~\ref{thm:disambiguation-negative1}. Recall that $\vcdim(\H_{\infty}) = 1$, and $\H_{\infty}$ is formed by taking disjoint copies of $\H_n$ so that $|\bar \H_n| \geq n^{(\log(n))^{1-o(1)}}$ for any disambiguation $\bar \H_n$ of $\H_n$.

Now suppose there is a 
compression scheme of some size $s_m$ 
depending on the sample size $m$.  
Then note that this also supplies a compression 
scheme for $\H_m$ for data sets of size $m$.
But then Proposition~\ref{prop:compression-disambiguation} 
implies there exists a disambiguation 
$\bar{\H}_m$ of $\H_m$ 
of size at most 
$(c m / s_m)^{s_m}$ for a numerical constant $c$.
On the other hand, 
Theorem~\ref{thm:disambiguation-negative1} 
provides that 
$\bar{\H}_m$ must have size at least 
$m^{(\log(m))^{1-o(1)}}$.
Therefore, 
\[
\left(\frac{c m}{s_m}\right)^{s_m}
\geq m^{(\log(m))^{1-o(1)}},
\]
which implies 
$s_m \geq c' (\log(m))^{1-o(1)}$ 
for a numerical constant $c'$.
\end{proof}

%\begin{theorem}\label{thm:threshold}
%There exists a partial concept class $\H$ with $\td(\H)\leq 2$ but $\LD(\H)=\infty$.
%\end{theorem}

To prove Theorem~\ref{thm:threshold}, 
we will rely on the following 
lemma.

\begin{lemma}
\label[lemma]{lem:LD-compression}
Any partial concept 
class $\H$ with $\LD(\H) < \infty$ 
admits a compression scheme 
of size $\LD(\H)$.
\end{lemma}
\begin{proof}
Consider the optimal 
online learning 
algorithm $\alg_{{\rm SOA}}$ 
of \citet*{littlestone:88}, defined 
formally (and for partial concept classes) 
in the proof of Theorem~\ref{thm:optimal-mistake-bound} 
in Section~\ref{sec:proofs-mistake-bounds}.
For any data sequence 
$(x_1,y_1),\ldots,(x_n,y_n) \in \X \times \{0,1\}$ 
realizable w.r.t.\ $\H$, we initialize 
$K = \{\}$. We then find a point $(x_i,y_i)$
in the sequence with $\alg_{{\rm SOA}}(K)(x_i) \neq y_i$, if one exists, and we 
add $(x_i,y_i)$ to the set $K$. 
We repeat this until there are no remaining 
points $(x_i,y_i)$ with 
$\alg_{{\rm SOA}}(K)(x_i) \neq y_i$.
This specifies a compression function 
$\kappa((x_1,y_1),\ldots,(x_n,y_n)) = K$, 
and the reconstruction function is simply 
$\alg_{{\rm SOA}}(K)$ (noting that this 
is invariant to the order of $K$). 
By Theorem~\ref{thm:optimal-mistake-bound}, 
we have $|K| \leq \LD(\H)$, so that 
this specifies a compression scheme 
of size $\LD(\H)$.
\end{proof}

\begin{proof}[of Theorem~\ref{thm:threshold}]
Consider again the partial concept 
class $\H_{\infty}$ defined in the proof of Theorem~\ref{thm:disambiguation-negative1}, 
and recall that $\td(\H_{\infty}) \leq 2$.
On the other hand, as established in 
Theorem~\ref{thm:compression}, 
$\H_{\infty}$ does not admit a bounded-size 
compression scheme.  Together 
with Lemma~\ref{lem:LD-compression}, it must be that 
$\LD(\H_{\infty}) = \infty$.
\end{proof}

\section{Online Learning: Detailed Results and Specific Bounds}
\label[section]{sec:online-learning}

The \emph{online learning} model for total concept classes is a classic 
framework, introduced by \citet*{littlestone:88}, in which there is a data sequence $(X_1,Y_1),(X_2,Y_2),\ldots$, 
which is considered \emph{arbitrary} (possibly adversarially-chosen), 
and we are interested in the number of \emph{mistakes} made by 
a learning algorithm $\alg$: that is, the number of times $t$ with 
$\alg((X_1,Y_1),\ldots,(X_{t-1},Y_{t-1}))(X_t) \neq Y_t$.

\subsection{Online Learning in the Realizable Case}
\label[section]{sec:online-realizable}

As with the PAC model, the online model has two main variants,
depending on whether we suppose the sequence $(X_i,Y_i)$ is 
realizable w.r.t.\ $\H$ (known as the \emph{realizable} case, or the \emph{mistake bound} model), 
or is arbitrary (known as \emph{agnostic} online learning).
We start by presenting the realizable case.
Specifically, we have the following definition.

\begin{definition}[Realizable Online Learnability]
\label[definition]{defn:online-learnability}
A partial concept class $\H$ is \emph{online learnable} if there 
exists a bound $\MB(\H) < \infty$ such that, 
for every $T \in \nats$, 
there exists a learning algorithm $\alg$ 
that, for every sequence $(X_1,Y_1),\ldots,(X_T,Y_T) \in \X \times \{0,1\}$ 
realizable w.r.t.\ $\H$, 
$\sum_{t=1}^{T} \ind[ \alg((X_1,Y_1),\ldots,(X_{t-1},Y_{t-1}))(X_t) \neq Y_t ] \leq \MB(\H)$.
\end{definition}

\citet*{littlestone:88} proved that, in the realizable case 
w.r.t.\ any \emph{total} concept class $\H$, 
there exists a learning algorithm making a \emph{bounded} number 
of mistakes if and only if the  combinatorial parameter $\LD(\H)$ 
is finite; this parameter has since become known as the 
\emph{Littlestone dimension}, and is a key quantity of interest 
in many learning models.  Here we use the extended definition of 
Littlestone dimension stated in 
Definition~\ref{defn:littlestone-dimension} for \emph{partial} concept classes.

We have the following result, extending Littlestone's classic 
theorem to hold for partial concept classes.  In particular, this supplies part of the claim in Theorem~\ref{thm:online-learnability-intro} 
from Section~\ref{sec:online-learning-results}.

\begin{theorem}[Realizable Online Learnability]
\label[theorem]{thm:online-learnability}
The following statements are equivalent for a partial concept class $\H\subseteq\{0,1,\star\}^\X$. 
\begin{itemize}
    \item $\LD(\H) < \infty$.
    \item $\H$ is online learnable.
\end{itemize}
\end{theorem}

%The proof is presented in Section~\ref{sec:online-proofs}.

\subsection{Proof of Theorem~\ref{thm:online-learnability} and Quantitative Mistake Bounds}
\label[section]{sec:proofs-mistake-bounds}

Theorem~\ref{thm:online-learnability} will follow 
as an immediate implication of the following 
more detailed result, which also supplies a concrete 
expression for the optimal mistake bound $\MB(\H)$.

\begin{theorem}[Optimal Mistake Bound]
\label[theorem]{thm:optimal-mistake-bound}
For any partial concept class $\H$, 
there exists an online learning algorithm 
making at most $\LD(\H)$ mistakes on any 
realizable data sequence.  
Moreover, for any finite $d \leq \LD(\H)$ 
and any (possibly randomized) learning algorithm $\alg$, 
there exists a realizable data sequence of 
length $d$ on which $\alg$ makes an expected 
number of mistakes at least $d/2$.
\end{theorem}
\begin{proof}
The proof of this result is essentially identical to the well-known 
proof for the special case of total concept classes.  We include the 
full details nonetheless, for completeness.

We begin with the upper bound.  For any sequence $S \in (\X \times \{0,1\})^*$, 
define $\H_{S} = \{ h \in \H : \hat{\er}_{S}(h) = 0 \}$.
Then, following \citet*{littlestone:88}, we first note that 
the Littlestone dimension can be interpreted inductively, and in particular,
for any non-empty $\H' \subseteq \H$,  
for every $x \in \X$, there exists $y \in \{0,1\}$ such that
$\LD(\H'_{(x,y)}) < \LD(\H')$, where we interpret $\LD(\{\}) = -1$.
Based on this, if $\LD(\H) < \infty$, we define the 
\emph{Standard Optimal Algorithm} (SOA) $\alg_{{\rm SOA}}$ as follows.
For any $S \in (\X \times \{0,1\})^*$ and $x \in \X$, 
let $\alg_{{\rm SOA}}(S)(x) = \argmax_{y \in \{0,1\}} \LD(\H_{S \cup \{(x,y)\}})$, 
breaking ties to favor $y=0$ (or by any other rule).
In particular, note that if $S$ is realizable w.r.t.\ $\H$, 
then at most one $y \in \{0,1\}$ can have $\LD(\H_{S \cup \{(x,y)\}}) = \LD(\H_{S})$ 
(it is also possible that neither value $y$ has this property).
Thus, for every sequence $(x_1,y_1),\ldots,(x_{T},y_{T})$ realizable w.r.t.\ $\H$, 
at each $t$ for which $\alg_{{\rm SOA}}((x_1,y_1),\ldots,(x_{t-1},y_{t-1}))(x_t) \neq y_t$, 
it must be that $\LD(\H_{\{ (x_1,y_1),\ldots,(x_t,y_t) \}}) \leq \LD(\H_{\{ (x_1,y_1),\ldots,(x_{t-1},y_{t-1}) \}}) - 1$.
Thus, there can be at most $\LD(\H)$ times $t$ at which this occurs: 
that is, $\alg_{{\rm SOA}}$ makes at most $\LD(\H)$ mistakes on any realizable sequence.

To prove the lower bound, we consider the set 
$\{ x_{\mathbf{y}} : \mathbf{y} \in \bigcup_{0 \leq i \leq d-1} \{0,1\}^i \} \subseteq \X$ 
from the definition of $\LD(\H)$.
We construct a data sequence via the probabilistic method.
Choose $\mathbf{y} = (y_1,\ldots,y_d) \in \{0,1\}^d$ 
uniformly at random, and define the data sequence 
as $X_1 = x_{()}$, 
$X_2 = x_{(y_1)}$, 
$X_3 = x_{(y_1,y_2)}$, 
$\ldots$, 
$X_d = x_{(y_1,\ldots,y_{d-1})}$, 
and for each $i \in \{1,\ldots,d\}$ let $Y_i = y_i$.
In particular, note that each $Y_t$ is independent of $(X_1,Y_1),\ldots,(X_{t-1},Y_{t-1}),X_t$.
Thus, for any learning algorithm $\alg$, we have 
$\P( \alg((X_1,Y_1),\ldots,(X_{t-1},Y_{t-1}))(X_t) \neq Y_t ) \geq \frac{1}{2}$ 
(it would be equal $1/2$ if the algorithm were restricted to outputting $0$ or $1$, not $\star$).
Therefore 
\[\E\!\left[ \sum_{t=1}^{d} \ind[ \alg((X_1,Y_1),\ldots,(X_{t-1},Y_{t-1}))(X_t) \neq Y_t ] \right] \geq \frac{d}{2}\] 
by linearity of the expectation.
In particular, by the law of total expectation, this implies there exists 
a deterministic sequence $(X_1,Y_1),\ldots,(X_d,Y_d)$ with this property.
\end{proof}

\subsection{Online Learning in the Agnostic Case}
\label[section]{sec:online-agnostic}

Similarly to the PAC model, the online learning model also has an 
\emph{agnostic} variant \citep*{ben2009agnostic}, which makes \emph{no} assumptions 
about the data sequence, and rather than bounding the number 
of mistakes, it bounds the \emph{excess} of the number of 
mistakes  made by the algorithm compared to the number made 
by the best hypothesis in the class $\H$: known as the 
\emph{regret}. Specifically, we have the following definition.

\begin{definition}[Agnostic Online Learnability]
\label[definition]{defn:agnostic-online-learnability}
A partial concept class $\H$ is 
\emph{agnostically online learnable} if there 
exists a sequence $\Regret(\H,T) = o(T)$ such 
that, for every $T \in \nats$, 
there exists a (possibly randomized) learning algorithm $\alg$ 
that, for every sequence $(X_1,Y_1),\ldots,(X_T,Y_T) \in \X \times \{0,1\}$, 
\[
\E \sum_{t=1}^{T} \ind[ \alg((X_1,Y_1),\ldots,(X_{t-1},Y_{t-1}))(X_t) \neq Y_t ] \leq \Regret(\H,T) + \min_{h \in \H} \sum_{t=1}^{T} \ind[ h(X_t) \neq Y_t ].
\]
\end{definition}

For the special case of $\H$ a \emph{total} concept class, 
\citet*{ben2009agnostic} proved that $\H$ is agnostically 
online learnable if and only if $\LD(\H) < \infty$.  
Here we extend this result to partial concept classes. 
In particular, this supplies the second part of the claim in Theorem~\ref{thm:online-learnability-intro} 
from Section~\ref{sec:online-learning-results}, 
so that proving this result will complete the proof 
of Theorem~\ref{thm:online-learnability-intro} as well.

\begin{theorem}[Agnostic Online Learnability]
\label[theorem]{thm:agnostic-online-learnability}
The following statements are equivalent for a partial concept class $\H\subseteq\{0,1,\star\}^\X$. 
\begin{itemize}
    \item $\Ldim(\H) < \infty$.
    \item $\H$ is agnostically online learnable.
\end{itemize}
\end{theorem}

%The proof is presented in Section~\ref{sec:online-proofs}.

\subsection{Proof of Theorem~\ref{thm:agnostic-online-learnability} and Quantitative Regret Bounds for the Agnostic Case}
\label[section]{sec:proofs-regret-bounds}

As above, Theorem~\ref{thm:agnostic-online-learnability} 
will follow as an immediate implication of the following 
more detailed result, which also supplies a concrete 
bound on the form of the optimal regret $\Reg(\H,T)$.

\begin{theorem}[Optimal Regret Bound]
\label[theorem]{thm:optimal-regret-bound}
For any partial concept class $\H$ with $\LD(\H) > 0$, 
there exists an online learning algorithm 
achieving an expected regret guarantee 
\[
\Reg(\H,T) = O\!\left(\sqrt{\LD(\H) T \ln(T/\LD(\H))} \right).
\]
Moreover, for any finite $d \leq \LD(\H)$, any $T \in \nats$ such that $T \geq d$, 
and any learning algorithm $\alg$, 
there exists a data sequence of length $T$ 
on which $\alg$ has expected regret 
$\Omega\!\left( \sqrt{ d T } \right)$.
\end{theorem}

The proof makes use of a classic result for learning from expert advice 
\citep*{vovk:90,vovk:92,littlestone:94,cesa-bianchi:97,kivinen:99,singer:99}; 
see Theorem 2.2 of \citet*{cesa-bianchi:06}.  
To simplify notation, we write $x_{1:t} := (x_1,\ldots,x_t)$.

\begin{lemma}[Experts; \citealp*{cesa-bianchi:06}, Theorem 2.2]
\label[lemma]{lem:experts}
For any $N,T \in \nats$ and $f_1,\ldots,f_N$ functions 
$\X^* \to [0,1]$, letting $\eta = \sqrt{(8/T)\ln(N)}$, 
for any $(x_1,y_1),\ldots,(x_T,y_T) \in \X \times [0,1]$, 
letting $w_{0,i} = 1$ and $w_{t,i} = e^{-\eta \sum_{s \leq t} |f_i(x_{1:s})-y_s|}$ for each $t \leq T$, $i \leq N$, 
letting $\bar{f}_t(x_{1:t},y_{1:(t-1)}) = \sum_i w_{t-1,i} f_{i}(x_{1:t}) / \sum_{i'} w_{t-1,i'}$, it holds that 
\begin{equation*}
\sum_{t=1}^{T} \left| \bar{f}_{t}(x_{1:t},y_{1:(t-1)}) - y_t \right| 
- \min_{1 \leq i \leq N} \sum_{t=1}^{T} |f_i(x_{1:t})-y_t| 
\leq \sqrt{(T/2)\ln(N)}.
\end{equation*}
\end{lemma}

We are now ready for the proof of Theorem~\ref{thm:optimal-regret-bound}.

\begin{proof}[of Theorem~\ref{thm:optimal-regret-bound}]
As was the case of Theorem~\ref{thm:optimal-mistake-bound},
the proof of this result is essentially based on existing 
proofs for the special case of total concept classes, though 
in this case a few important changes are required.
We include the full details for completeness.

The upper bound is based on the work of \citet*{ben2009agnostic}.
Consider any data sequence 
$(X_1,Y_1),\ldots,(X_T,Y_T)$, 
and let $h^* = \argmin_{h \in \H} \sum_{t=1}^{T} \ind[h(X_t) \neq Y_t]$ 
(breaking ties arbitrarily).
Let $t_1,\ldots,t_{q}$ denote the subsequence of $1,\ldots,T$ 
such that $h^*(X_{t_i}) \neq \star$.
In particular, $(X_{t_1},h^*(X_{t_1})),\ldots,(X_{t_q},h^*(X_{t_q}))$ 
is realizable w.r.t.\ $\H$.
Let $\alg_{{\rm SOA}}$ be as in the proof of Theorem~\ref{thm:optimal-mistake-bound}, 
and recall from that proof that 
$\alg_{{\rm SOA}}$ makes at most $\LD(\H)$ mistakes on any realizable data sequence.
Now construct a subsequence $J^*$ of $\{t_1,\ldots,t_q\}$ as follows.
Initialize $J^* = ()$.
For each $i \in \{1,\ldots,q\}$ in increasing order, 
if $\alg_{{\rm SOA}}( \{ (X_{t},h^*(X_t)) \}_{t \in J^*} )(X_{t_i}) \neq h^*(X_{t_i})$ 
then append $t_i$ to the sequence $J^*$ and continue.
In particular, note that the sequence $\{(X_{t},h^*(X_{t}))\}_{t \in J^*}$ 
is realizable w.r.t.\ $\H$, so that $\alg_{{\rm SOA}}$ makes at most $\LD(\H)$ mistakes.
On the other hand, enumerating $J^* = \{j_1,\ldots,j_{|J^*|}\}$, 
for each $k \leq |J^*|$ we have 
$\alg_{{\rm SOA}}( \{ (X_{j_i},h^*(X_{j_i})) \}_{i < k} )(X_{j_k}) \neq h^*(X_{j_k})$.
Thus, it must be that $|J^*| \leq \LD(\H)$.
Moreover, for each $t \in \{t_1,\ldots,t_q\}$ with $t \notin J^*$, 
we have 
$\alg_{{\rm SOA}}( \{ (X_{j_i},h^*(X_{j_i})) \}_{j_i < t} )(X_{t}) = h^*(X_{t})$.

Now, for each subsequence $J$ of $\{1,\ldots,T\}$ with $|J| \leq \LD(\H)$, 
for each $j \in J$, inductively define $\hat{Y}^{J}_{j} = 1-\alg_{{\rm SOA}}( \{ (X_{j'},\hat{Y}^{J}_{j'}) \}_{j' \in J : j' < j})(X_{j})$.
Then define an algorithm $\alg^{J}$ that, 
for each $t \in \{1,\ldots,T\}$ with $t \notin J$, has 
$\alg^{J}( X_1,\ldots,X_{t} ) = \alg_{{\rm SOA}}( \{ (X_{j},\hat{Y}^{J}_{j}) \}_{j \in J : j < t} )(X_{t})$,
and for each $t \in J$, 
$\alg^{J}( X_1,\ldots,X_{t} ) = 1-\alg_{{\rm SOA}}( \{ (X_{j},\hat{Y}^{J}_{j}) \}_{j \in J : j < t} )(X_{t})$.
In particular, note that for all $i \in \{1,\ldots,q\}$, 
$\alg^{J^*}(X_1,\ldots,X_{t_i}) = h^*(X_{t_i})$.
Therefore, since $h^*(X_t) \neq Y_t$ at every $t \notin \{t_1,\ldots,t_q\}$, 
we have 
$\sum_{t=1}^{T} \ind\!\left[ \alg^{J^*}(X_1,\ldots,X_{t}) \neq Y_t \right] 
\leq \sum_{t=1}^{T} \ind[ h^*(X_t) \neq Y_t ]$.

Now to describe the online learning algorithm achieving the regret guarantee 
from the theorem, we will apply the classic exponential weights algorithm 
with these $\alg^{J}$ predictors as the ``experts''.  
Specifically, applying Lemma~\ref{lem:experts} with the above value of $T$, 
with $N = \sum_{i=0}^{\LD(\H)} \binom{T}{i}$, and with functions $f_1,\ldots,f_N$ 
given by an enumeration of the algorithms 
$\alg^{J}$, $J \subseteq \{1,\ldots,T\}$, $|J| \leq \LD(\H)$, 
we conclude that, for $\bar{f}_{t}$ as defined in Lemma~\ref{lem:experts}, 
\begin{equation*}
\sum_{t=1}^{T} \left| \bar{f}_{t}(X_{1:t},Y_{1:(t-1)}) - Y_t \right| 
- \min_{1 \leq i \leq N} \sum_{t=1}^{T} |f_i(X_{1:t})-Y_t| 
\leq \sqrt{(T/2)\ln(N)}.
\end{equation*}
Let us then define a randomized predictor $\alg$ such that, 
for any $t \in \{1,\ldots,T\}$,  
\linebreak $\alg((X_1,Y_1),\ldots,(X_{t-1},Y_{t-1}))(X_t)$ 
evaluates to $1$ 
with probability $\bar{f}_{t}(X_{1:t},Y_{1:(t-1)})$, and otherwise 
evaluates to $0$ 
(where this random evaluation occurs independently for each $t$).
We then have 
\begin{align*}
& \E\!\left[ \sum_{t=1}^{T} \ind\!\left[ \alg((X_1,Y_1),\ldots,(X_{t-1},Y_{t-1}))(X_t) \neq Y_t \right]  \right]
= \sum_{t=1}^{T} \left| \bar{f}_{t}(X_{1:t},Y_{1:(t-1)}) - Y_t \right| 
\\ & \leq \min_{1 \leq i \leq N} \sum_{t=1}^{T} |f_i(X_{1:t})-Y_t| 
+ \sqrt{(T/2)\ln(N)} 
\\ & \leq \sum_{t=1}^{T} \ind\!\left[ \alg^{J^*}(X_1,\ldots,X_t) \neq Y_t \right] 
+ \sqrt{(T/2)\ln\!\left( \sum_{i \leq \LD(\H)} \binom{T}{i} \right)}
\\ & \leq \sum_{t=1}^{T} \ind\!\left[ h^*(X_t) \neq Y_t \right] 
+ O\left(\sqrt{\LD(\H) T \ln\!\left( \frac{T}{\LD(\H)} \right)}\right).
\end{align*}
This completes the proof of the upper bound.

The lower bound proof is essentially identical to the existing proof 
for total concepts from \citet*{ben2009agnostic}, but we include the details for completeness.
Given $0 < d \leq \LD(\H)$ and $T \geq d$, let $k = \lfloor T/d \rfloor$.
Define the sequence $Y_1,\ldots,Y_T$ as independent ${\rm {Bernoulli}}(1/2)$ random variables.
Consider the set 
$\{ x_{\mathbf{y}} : \mathbf{y} \in \bigcup_{0 \leq i \leq d-1} \{0,1\}^i \}$ 
from the definition of $\LD(\H)$ (Definition~\ref{defn:littlestone-dimension}).
Let $T_0 = 0$, 
and for each $i \in \{1,\ldots,d-1\}$, let $T_i = k i$, 
and let $T_d = T$.
Then, for each $i \in \{1,\ldots,d\}$, 
let $y_i = {\rm Majority}(Y_{T_{i-1}+1},\ldots,Y_{T_i})$ 
and for each $t \in \{ T_{i-1}+1,\ldots, T_i \}$,
define $X_t = x_{\{y_{i'}\}_{i' < i}}$.

Since the $Y_t$ values are independent, for any learning algorithm we certainly have 
\begin{equation*}
\E\!\left[ \sum_{t=1}^{T} \ind\!\left[ \alg((X_1,Y_1),\ldots,(X_{t-1},Y_{t-1}))(X_t) \neq Y_t \right] \right] \geq \frac{T}{2}
\end{equation*}
(with equality if $\alg$ outputs $0$ or $1$).
It remains only to upper bound the value 
\linebreak $\E\!\left[ \min_{h \in \H} \sum_{t=1}^{T} \ind[ h(X_t) \!\neq\! Y_t ] \right]$.
In particular, consider a partial concept $\bar{h} \!\in\! \H$ 
with $\bar{h}(x_{\{y_{i'}\}_{i' < i}}) \!=\! y_i$ for each $i \in \{1,\ldots,d\}$, 
which exists by definition of $x_{\mathbf{y}}$ from Definition~\ref{defn:littlestone-dimension}.
Then, for each $i \in \{1,\ldots,d\}$, 
\begin{align*} 
 \sum_{t = T_{i-1}+1}^{T_i} \!\!\!\left( 2 \ind\!\left[ \bar{h}(X_t) = Y_t \right] - 1 \right) 
& = \sum_{t = T_{i-1}+1}^{T_i} \!\!\!\left( 2 \ind\!\left[ Y_t = y_i \right] - 1  \right)  
 = \left| \sum_{t = T_{i-1}+1}^{T_i} \!\!( 2 Y_t - 1 ) \right|,
\end{align*}
and Khinchine's inequality (see Lemma A.9 of \citealp*{cesa-bianchi:06}) 
implies
\begin{equation*}
\E\!\left[ \left| \sum_{t = T_{i-1}+1}^{T_i} ( 2 Y_t - 1 ) \right| \right] 
\geq \sqrt{(T_i - T_{i-1})/2}.
\end{equation*}
Thus, since 
\begin{equation*}
\E\!\left[ \sum_{t = T_{i-1}+1}^{T_i} \left( 2 \ind\!\left[ \bar{h}(X_t) = Y_t \right] - 1 \right) \right] 
= (T_i-T_{i-1}) - 2 \E\!\left[ \sum_{t = T_{i-1}+1}^{T_i} \ind\!\left[ \bar{h}(X_t) \neq Y_t \right] \right],
\end{equation*}
we conclude that 
\begin{equation*}
\E\!\left[ \sum_{t = T_{i-1}+1}^{T_i} \ind\!\left[ \bar{h}(X_t) \neq Y_t \right] \right]
\leq \frac{T_i - T_{i-1}}{2} - \sqrt{\frac{T_i - T_{i-1}}{8}}.
\end{equation*}
Therefore, 
\begin{equation*}
\E\!\left[ \sum_{t = 1}^{T} \ind\!\left[ \bar{h}(X_t) \neq Y_t \right] \right]
\leq \sum_{i=1}^{d} \frac{T_i - T_{i-1}}{2} - \sqrt{\frac{T_i - T_{i-1}}{8}}
\leq \frac{T}{2} - \sqrt{\frac{d^2 k}{8}}
\leq \frac{T}{2} - \frac{1}{4} \sqrt{d T}.
\end{equation*}
Altogether, 
\begin{equation*}
\E\!\left[ \sum_{t=1}^{T} \ind\!\left[ \alg((X_1,Y_1),\ldots,(X_{t-1},Y_{t-1}))(X_t) \neq Y_t \right] - \min_{h \in \H} \sum_{t=1}^T \ind[ h(X_t) \neq Y_t ] \right] 
\geq (1/4) \sqrt{d T}.
\end{equation*}
In particular, by the law of total expectation, this also implies 
there exists a ($\alg$-dependent) deterministic choice of the sequence 
$(X_1,Y_1),\ldots,(X_T,Y_T)$ satisfying this.
\end{proof}

We note that the upper bound for total concept classes 
has been refined by \citet*{alon:21} to match the 
lower bound up to numerical constants: 
that is $\Reg(\H,T) = \Theta\!\left( \sqrt{\LD(\H) T} \right)$.
However, that proof uses techniques for which it is unclear 
whether they can be extended to partial concept classes.
Thus, there remains an open question:

\begin{question}
Is the optimal 
regret for partial concept classes always 
$\Theta\!\left( \sqrt{\LD(\H) T}\right)$?
\end{question}

\bibliography{ideas,learning}

\end{document}